\documentclass[english,,man,floatsintext]{apa6}
\usepackage{lmodern}
\usepackage{amssymb,amsmath}
\usepackage{ifxetex,ifluatex}
\usepackage{fixltx2e} 
\ifnum 0\ifxetex 1\fi\ifluatex 1\fi=0 
  \usepackage[T1]{fontenc}
  \usepackage[utf8]{inputenc}
\else 
  \ifxetex
    \usepackage{mathspec}
  \else
    \usepackage{fontspec}
  \fi
  \defaultfontfeatures{Ligatures=TeX,Scale=MatchLowercase}
\fi
\IfFileExists{upquote.sty}{\usepackage{upquote}}{}
\IfFileExists{microtype.sty}{%
\usepackage{microtype}
\UseMicrotypeSet[protrusion]{basicmath} 
}{}
\usepackage{hyperref}
\hypersetup{unicode=true,
            pdftitle={The Language of Generalization},
            pdfauthor={Michael Henry Tessler~\& Noah D. Goodman},
            pdfkeywords={genericity, generalization, generics, pragmatics, semantics, Bayesian
modeling},
            pdfborder={0 0 0},
            breaklinks=true}
\ifnum 0\ifxetex 1\fi\ifluatex 1\fi=0 
  \usepackage[shorthands=off,main=english]{babel}
\else
  \usepackage{polyglossia}
  \setmainlanguage[]{english}
\fi
\usepackage{graphicx,grffile}
\makeatletter
\def\maxwidth{\ifdim\Gin@nat@width>\linewidth\linewidth\else\Gin@nat@width\fi}
\def\maxheight{\ifdim\Gin@nat@height>\textheight\textheight\else\Gin@nat@height\fi}
\makeatother
\setkeys{Gin}{width=\maxwidth,height=\maxheight,keepaspectratio}
\IfFileExists{parskip.sty}{%
\usepackage{parskip}
}{
\setlength{\parindent}{0pt}
\setlength{\parskip}{6pt plus 2pt minus 1pt}
}
\ifx\paragraph\undefined\else
\let\oldparagraph\paragraph
\renewcommand{\paragraph}[1]{\oldparagraph{#1}\mbox{}}
\fi
\ifx\subparagraph\undefined\else
\let\oldsubparagraph\subparagraph
\renewcommand{\subparagraph}[1]{\oldsubparagraph{#1}\mbox{}}
\fi

\let\rmarkdownfootnote\footnote%
\def\footnote{\protect\rmarkdownfootnote}

  \title{The Language of Generalization}
    \author{Michael Henry Tessler\textsuperscript{1}~\& Noah D.
Goodman\textsuperscript{1}}
    \date{}
  
\shorttitle{The language of generalization}
\affiliation{
\vspace{0.5cm}
\textsuperscript{1} Department of Psychology, Stanford University}
\usepackage{csquotes}
\usepackage{upgreek}
\usepackage{longtable}
\usepackage{lscape}
\usepackage{multirow}
\usepackage{tabularx}
\usepackage[flushleft]{threeparttable}
\usepackage{threeparttablex}

\makeatletter
\newcommand\LastLTentrywidth{1em}
\newlength\longtablewidth
\newcommand{\getlongtablewidth}{\begingroup \ifcsname LT@\roman{LT@tables}\endcsname \global\longtablewidth=0pt \renewcommand{\LT@entry}[2]{\global\advance\longtablewidth by ##2\relax\gdef\LastLTentrywidth{##2}}\@nameuse{LT@\roman{LT@tables}} \fi \endgroup}
\usepackage{tabularx}
\usepackage{multicol}
\usepackage{wrapfig}
\usepackage{caption}
\usepackage{booktabs}
\usepackage{amsmath}
\usepackage{graphicx}
\usepackage{subcaption}
\usepackage{longtable}
\usepackage{array}
\usepackage{multirow}
\usepackage{xcolor}

\authornote{This paper is based on a Ph.D.~dissertation
submitted to Stanford University. The research was supported by National
Science Foundation Graduate Research Fellowship DGE-114747 awarded to
MHT and DARPA PPAML grant FA8750-14-2-0009 and an Alfred P. Sloan
Research Fellowship to NDG. A portion of this paper (Case Study 2)
appeared in the proceedings of the 38th Annual Meeting of the Cognitive
Science Society, Philadelphia, 2016. We are deeply grateful to Ellen
Markman, Gregory Scontras, Thomas Icard, Michael C. Frank, Michael
Franke, and Sophie Bridgers for extensive conversations about and
support for this work. In addition, the work was greatly improved
through countless discussions with members of the Computation and
Cognition Lab and the Language and Cognition Lab at Stanford, as well as
the participants of \emph{A Generic Workshop} at the Center for the
Study of Language and Information (CSLI) at Stanford in May 2017.
Finally, we thank two anonymous reviewers and Joshua Hartshorne for
helping us clarify our main arguments and improving the quality of the
manuscript. All models, analyses, data, and links to experiments used in
this article can be found at
\url{https://github.com/mhtess/genlang-paper}

Correspondence concerning this article should be addressed to Michael
Henry Tessler, 450 Serra Mall, Bldg. 420, Rm. 316, Stanford, CA 94305.
E-mail:
\href{mailto:tessler@mit.edu}{\nolinkurl{tessler@mit.edu}}}

\abstract{
Language provides simple ways of communicating generalizable knowledge
to each other (e.g., ``Birds fly'', ``John hikes'', ``Fire makes
smoke''). Though found in every language and emerging early in
development, the language of generalization is philosphically puzzling
and has resisted precise formalization. Here, we propose the first
formal account of generalizations conveyed with language that makes
quantitative predictions about human understanding. We test our model in
three diverse domains: generalizations about categories (generic
language), events (habitual language), and causes (causal language). The
model explains the gradience in human endorsement through the interplay
between a simple truth-conditional semantic theory and diverse beliefs
about properties, formalized in a probabilistic model of language
understanding. This work opens the door to understanding precisely how
abstract knowledge is learned from language.

}

\usepackage{amsthm}

\theoremstyle{definition}

\theoremstyle{definition}

\theoremstyle{definition}

\theoremstyle{remark}

\begin{document}
\maketitle

\newcommand{\denote}[1]{\mbox{ $[\![ #1 ]\!]$}}
\newcommand*\diff{\mathop{}\!\mathrm{d}}
\definecolor{Red}{RGB}{255,0,0}
\definecolor{Green}{RGB}{10,200,100}
\definecolor{Blue}{RGB}{10,100,200}

\newcommand{\mht}[1]{{\textcolor{Blue}{[mht: #1]}}}
\newcommand{\ndg}[1]{{\textcolor{Green}{[ndg: #1]}}}
\newcommand{\red}[1]{{\textcolor{Red}{#1}}}

\hypertarget{introduction}{%
\section{Introduction}\label{introduction}}

Knowledge that extends beyond the present context is crucial to thrive
in our open-ended, dynamic world. Yet such knowledge can be difficult to
extract from the environment: The relevant observations may be costly
(e.g., learning that a plant is poisonous) or rare (e.g., understanding
that lightning strikes tall objects). Fortunately, we are not limited to
acquiring generalizations on our own; language allow us to communicate
generalizations to each other. By sharing generalizable knowledge, we
flourish collectively without individually needing to taste
potentially-poisonous plants or personally witness lightning strikes.
Being able to flexibly communicate generalizations from one generation
to the next supports the faithful transmission of knowledge necessary
for culture to cumulatively evolve (Henrich, 2015; Tomasello, 1999).

The \emph{language of generalization} covers a diverse swath of natural
language expressions. Generic language conveys generalizations about
categories (e.g., ``Dogs have four legs''; Carlson, 1977; Cohen, 1999;
Leslie, 2007; Nickel, 2008) and is the most well-studied case of
generalizations in language.\footnote{Some writers refer to the all of
  the \emph{language of generalization} as \enquote{generic language} or
  generics. In both the empirical and theoretical literatures, however,
  analysis and experiments often focus only on generalizations about
  categories. We use the more narrow-scoped terms (e.g., generics,
  habituals, and causals) to highlight the diversity of semantic types
  being predicated.} In contrast to statements about concrete
individuals (e.g., \enquote{Rufus has four legs}), generic statements
refer to inherently unobservable categories (e.g., the category of
\textsc{dog}) and convey information that extends beyond the present
context, a fact which children as young as 2 appreciate (Cimpian \&
Markman, 2008). Simple events (e.g., \enquote{John ran yesterday}) can
be generalized into habitual sentences (e.g., \enquote{John runs}), and
even events of complex inferential types such as actual causal events
(e.g., \enquote{The fire caused the smoke}) can be described in
generalization (e.g., \enquote{Fire causes smoke}).

Understanding the language of generalization is a project with
far-reaching implications. The language of generalization is ubiquitous
in everyday conversation, is found in every language (Behrens, 2005;
Carlson \& Pelletier, 1995), and conveys rich meanings, impacting
motivation (Cimpian, Arce, Dweck, \& Markman, 2007), transmitting
stereotyped beliefs about social categories (Rhodes, Leslie, \& Tworek,
2012), and making meaning from experience (Orvell, Kross, \& Gelman,
2017). It is highly-prevalent in child-directed speech (Gelman, Goetz,
Sarnecka, \& Flukes, 2008) and its ability to refer to abstractions
beyond the present context suggests its centrality to the growth of
conceptual knowledge (Gelman, 2004).

Despite its ability to convey abstract knowledge, its ubiquity in
discourse, and its relative morphosyntactic simplicity, the language of
generalization displays subtle context-sensitivities that make it
difficult to formalize. \enquote{Robins lay eggs} sounds true and
\enquote{Robins are female} does not. Yet in each case, only 50\% of the
category has the property (i.e., only the females lay eggs).
\enquote{Mosquitos carry malaria} sounds true despite malaria being
present in only a tiny fraction of mosquitos. Even more perplexing:
\enquote{Supreme Court Justices have even social security numbers} is
thought to be intuitively a false generalization even if it were the
case that on the current bench, nine out of nine justices had even
social securtity numbers (i.e., when the sentence ``All Supreme Court
Justices have even social security numbers'' is true; Cohen, 1999).
Similar context-sensitivity can be observed with habituals:
\enquote{Mary climbs mountains} could imply a few mountain climbs a year
for Mary, but \enquote{John runs} would be infelicitious if John went
for a run three times last year.

These observations have led some to conclude that the literal meaning of
a generic statement (and by analogy, other generalizations in language)
involves more than just the number of members of the category who have
the property, otherwise known as the \emph{prevalence} of the feature in
the category. Instead, theorists have argued that generics are not
treatable by the standard tools of truth-functional semantics (Montague,
1973), but rather should be thought of semantically as a direct,
linguistic manifestation of abstract relations between kinds and
properties (Leslie, 2008; Prasada, 2000; Prasada, Hennefield, \& Otap,
2012). For example, the statements \enquote{Bishops move diagonally} or
\enquote{The Speaker of the House succeeds the Vice President} are true
not because of a tendency on behalf of instances of a category to
actually uphold the property, but rather the existence of a conceptual
relationship (e.g., what it means, in the game of chess, to be a
bishop). True generics can be supported by different underlying types of
category--property relations (e.g., principled vs.~statistical
connections; Prasada \& Dillingham, 2006), and thus support
qualitatively different inferences (e.g., ``Being striped is one aspect
of being a tiger'' is generally endorsed, while ``Carrying malaria is
one aspect of being a mosquito'' is not, even while ``Tigers are
striped'' and ``Mosquitos carry malaria'' are both intuitively true;
Prasada, Khemlani, Leslie, \& Glucksberg, 2013). This conceptual view of
generics has been influential in psychology, because it predicts
qualitative differences between different kinds of generics.

Insofar as there is a single class of linguistic expressions that convey
generalization, however, there should be something common to them all: a
literal meaning that unifies generic, habitual, and causal language. In
this paper, we propose such a semantic core based on prevalence and
formalized using the tools of truth-functional semantics (Cohen, 1999;
Montague, 1973). A semantics based on prevalence will not be enough to
capture subtle sensitivities to context that the language of
generalization exhibit. We propose that the meaning of generalizations
is underspecified or \emph{vague}, and that listeners derive a more
precise interpretation in context using probabilistic world knowledge.

The fact that generics are vague does not preclude them from being
treated with formal models. We draw upon the tools of Bayesian models of
cognition (Tenenbaum, Kemp, Griffiths, \& Goodman, 2011) to formalize
the vagueness and context-dependence of generic language (Frank \&
Goodman, 2012; Goodman \& Lassiter, 2015). The Bayesian model separates
the semantics of an utterance conveying a generalization from the world
knowledge a listener would use to interpret the utterance, a key
theoretical advancement beyond previous accounts. This formal model is
the first of its kind to make quantitative predictions about human
understanding of the l anguage of generalization.

The paper is organized as follows. In the next section, we describe our
computational framework for interpreting generalizations and our precise
model for endorsing such statements (i.e., a model of truth judgments).
To illustrate how the model works, we then work through a number of
standard examples from the linguistics literature that any model of
generics should be able to accomodate. The third section discusses the
relationship of our model to previous accounts of generics from the
linguistics and philosophical literatures. These theoretical sections
are followed by three empirical case studies: generalizations about
categories (\emph{generic language}), events (\emph{habitual language}),
and causes (\emph{causal language}). In the case study of generics, we
use measure relevant background knowledge and prevalence to predict
endorsements of familiar generic statements (e.g., \enquote{Robins lay
eggs}). In the study of habituals, we measure background knowledge but
manipulate the prevalence or frequency of the event to predict
endorsements of habitual statements about novel agents (e.g.,
\enquote{John runs} given that he's run a certain number of times in the
past). In this case study, we also further examine the nature of the
relevant probabilities for endorsing generalizations, asking whether it
matters how often John has run in the past (past frequency) or how often
a speaker expects him to run in the future (predictive probability).
Finally, in our last case study, we manipulate background knowledge to
show its causal influence on endorsing generalizations about novel
causal events (e.g., \enquote{Herb X makes wugs sleepy}). We compare our
model to two previously articulated quantitative models of generics as
well as a lesioned-version of our model. In all cases, we find a very
strong agreement of our model's predictions to human elicited
endorsements, where the simpler models fall short. We conclude our paper
by clarifying some of the theoretical claims of the model and discuss
open-questions for this approach.

\hypertarget{computational-framework}{%
\section{Computational Framework}\label{computational-framework}}

Generalizations are used to make predictions about events or properties
of instances that an agent has yet to experience (Hume, 1888). People
readily predict that the next dog they encounter will have four legs,
drinking another cup of coffee will cause jitters, and a new day will
find the people that we know doing what they habitually do. In each
case, we assign a specific exemplar \(x\) to a category \(k\), and make
a prediction that it will have feature \(f\). This prediction can be
described by a conditional probability: \(P(x \in f \mid x \in k)\), the
probability that \(x\) will have \(f\) given that it is in \(k\)
(formally, an instance of \(k\) will be in the set of things that have
\(f\)), which we will refer to as the \emph{prevalence} and for
convenience write as \(p = P(f \mid k)\). The \emph{prevalence} \(p\) is
a \emph{prevalence in the mind}, a latent belief that a future instance
of a category would have a particular property; others might call this
projectibility (Goodman, 1955). The targets of our predictions can vary
widely: They may be objects (a dog), events (an instance of
coffee-drinking), or more ad-hoc types (a person on a particular day).
The properties also may vary (e.g., having fur, causing jitters, going
to the gym). Yet the mathematical description of the inductive belief is
always given by a probability \(P(f \mid k)\).

Probability is a useful representation for human generalization from
observations (Shepard, 1987). If you observe several \emph{wugs} (a
novel category) that have two legs, you might infer that all wugs have
two legs. But not all properties have such strong projectibility: Seeing
a wug with broken wings tells you comparatively less about other wugs
having broken wings (Nisbett, Krantz, Jepson, \& Kunda, 1983). Abstract,
potentially domain-specific, beliefs about the projectibility of
different properties can be represented by a probability distribution
over the prevalence \(P(p)\) (Kemp, Perfors, \& Tenenbaum, 2007). By
assuming some generative process that could produce one's observations
\(o\)---a likelihood function \(P(o \mid p)\)---Bayes' Theorem provides
the mathematically correct way to update one's prior beliefs from
observations: \(P(p \mid o) \propto P(o \mid p) \cdot P(p)\) (Tenenbaum
\& Griffiths, 2001).

Observational data is not always available, however. Instead, we must
listen to others to learn about properties that are costly to observe
(e.g., \emph{staring at the sun makes you go blind}), events that are
statistically unlikely (e.g., \emph{lightning strikes tall objects}), or
any aspect of the world that we have yet to experience. Fortunately,
language provides simple ways of communicating generalizations.

\hypertarget{communicating-generalizations}{%
\subsection{Communicating
generalizations}\label{communicating-generalizations}}

The language of generalization is easy to spot when a property, that
could apply to an individual, is predicated of a category (e.g.,
\enquote{Dogs have four legs}: \emph{has four legs} could apply to an
individual as in \enquote{Rufus has four legs}). In the semantics
literature on generics, bare plural sentences of this kind are sometimes
described as \emph{characterizing sentences} in contrast with
\emph{kind-denoting} sentences where the property can only meaningfully
apply to the category as a whole (e.g., \enquote{Dinosaurs are extinct};
\emph{extinct} cannot apply to an individual dinosaur). Generalization
can also manifest when describing instances of an individual
(\emph{habitual language}; e.g., ``John smokes''; Carlson, 1977, 2005);
in this case, the particular instance being generalized is an instance
of an individual (e.g., John at a particular moment in time), which also
permits predication (e.g., \enquote{John smoked yesterday after
dinner}). Verbs like \emph{causes} or \emph{makes} also seem to convey
generalization, in this case, about an instance of an actual causal
event (\emph{causal language}; e.g., \enquote{Fire causes smoke}).
Psychologists, linguists, and philosophers have long studied the
language of generalization, as it appears very simple (e.g.,
syntactically) yet its meaning is difficult to formalize.\footnote{We
  further distinguish the problem of formalizing a \emph{meaning} for
  the language of generalization from the problem of \emph{identifying}
  generalizations. Syntax alone is neither necessary nor sufficient for
  a listener to know that the sentence conveys a generalization (e.g.,
  indefinite singulars can encode generalizations: \enquote{A dog has
  four legs} and bare plurals may not: \enquote{Dogs are on my front
  lawn}). Our analysis thus begins once a sentence has been
  disambiguated as conveying a generalization. }

The basic intuition behind our account is that before a listener hears a
novel generalization such as \enquote{Alligators grow to be 10-feet
long}, they do not know how widely distributed the property to be in the
category, including whether or not it is present at all. The utterance
provides a vague sense of how strongly the generalization applies (e.g.,
how many alligators grow to be 10-feet long), which the listener derives
from their knowledge of how the property (\emph{growing to be 10-feet
long}) is distributed among other categories (e.g., other animals). The
decision of whether or not to endorse the generalization is that of a
speaker reasoning about how well the utterance would align their
interlocutor's beliefs about the prevalence of the feature in the
category with those of their own. We formalize this intuition in a
truth-conditional semantics incorporated into a Bayesian model of belief
updating.

\hypertarget{interpretation-model}{%
\subsubsection{Interpretation model}\label{interpretation-model}}

Our model of interpreting the language of generalizations has three
conceptual components: Probability, vagueness, and context. If
generalization from observations can be described by a probability
\(p\), it is natural to posit that same construct will be at the heart
of a semantic theory of the language of generalization (Ingredient 1:
Probability). In semantics, belief updating generally passes through
Boolean \emph{truth values} (Montague, 1973). The simplest way to derive
a Boolean from a scalar quantity like probability is via a
\emph{threshold semantics}: The utterance is true if the relevant scalar
value is above a threshold. For example, the literal meaning of the
sentence \enquote{Some dogs have four legs} is that there is a non-zero
chance that a given dog will have four legs:
\(\mbox{ $[\![ some ]\!]$}(p) := p > 0\). \enquote{Most dogs have four
legs} can also be described as a threshold on prevalence (e.g., the
chance that a dog will have four legs is greater than 50\%):
\(\mbox{ $[\![ most ]\!]$}(p) := p > 0.5\).\footnote{These definitions
  concern the standard semantic truth conditions of quantifiers, not
  their pragmatic interpretations (e.g., that \enquote{some} often
  implies \enquote{not all}).} Thus, the simplest semantics for a
generalization would also be a threshold on the prevalence:
\(\mbox{ $[\![ gen ]\!]$}(p, \theta) := p >\theta\).

The extreme flexibility of generalizations (e.g., \enquote{Mosquitos
carry malaria}; \enquote{Birds lay eggs} vs. \enquote{Birds are female})
suggests that no fixed threshold would suffice. Rather than throw out
the threshold-semantics, we posit that the threshold is
\emph{underspecified} in the literal meaning and is
contextually-determined in a way analogous to how gradable adjectives
like \emph{tall} have contextually-determined thresholds (e.g., what
counts as tall for a person is different than what counts as tall for a
building; Kennedy, 2007; Lassiter \& Goodman, 2013). We formalize this
underspecification of meaning (Ingredient 2: Vagueness) by putting a
probability distribution over \(\theta\): \(P(\theta)\) (Lassiter \&
Goodman, 2013, 2015; cf., Qing \& Franke, 2014).

Finally, the meanings of linguistic expressions manifest in their
capacity to convey information from speaker to hearer (Clark, 1996;
Grice, 1975; Levinson, 1995). Thus, the final ingredient to our model is
\emph{context} (Ingredient 3), which must be minimally formalized as a
listener's prior knowledge about the property \(P(p)\) and which we
describe in the \emph{Worked Examples} section below.

Putting these three ingredients (probability, vagueness, and context)
together, we get the following model of interpreting generalizations in
language:

\begin{eqnarray}
L(p, \theta \mid u) &\propto& {\delta_{\mbox{ $[\![ u ]\!]$}(p, \theta)}  \cdot P(\theta) \cdot P(p)} \label{eq:L0}
\end{eqnarray}

We denote this probabilistic interpretation model by \(L\) to indicate
that it is modeling a listener updating their beliefs according to the
truth-functional meaning of an utterance \(u\). Formally, the
truth-functional meaning is represented by the Kronecker delta function
\(\delta_{\mbox{ $[\![ u ]\!]$}(p, \theta)}\) that returns \(1\) when
the utterance is true (i.e., when \(p > \theta\)) and \(0\) otherwise.

\begin{eqnarray}
\delta_{\mbox{ $[\![ u_{gen} ]\!]$}(p, \theta)} &\propto  & \begin{cases}
1 & \text{if } p > \theta \\
0 & \text{otherwise}
\end{cases}\label{eq:delta}
\end{eqnarray}

Following Lassiter and Goodman (2015), we formalize the vagueness in the
meaning of generalizations \(P(\theta)\) as a uniform distribution over
the support of \(P(p)\).

\begin{figure}

{\centering \includegraphics[width=1\linewidth]{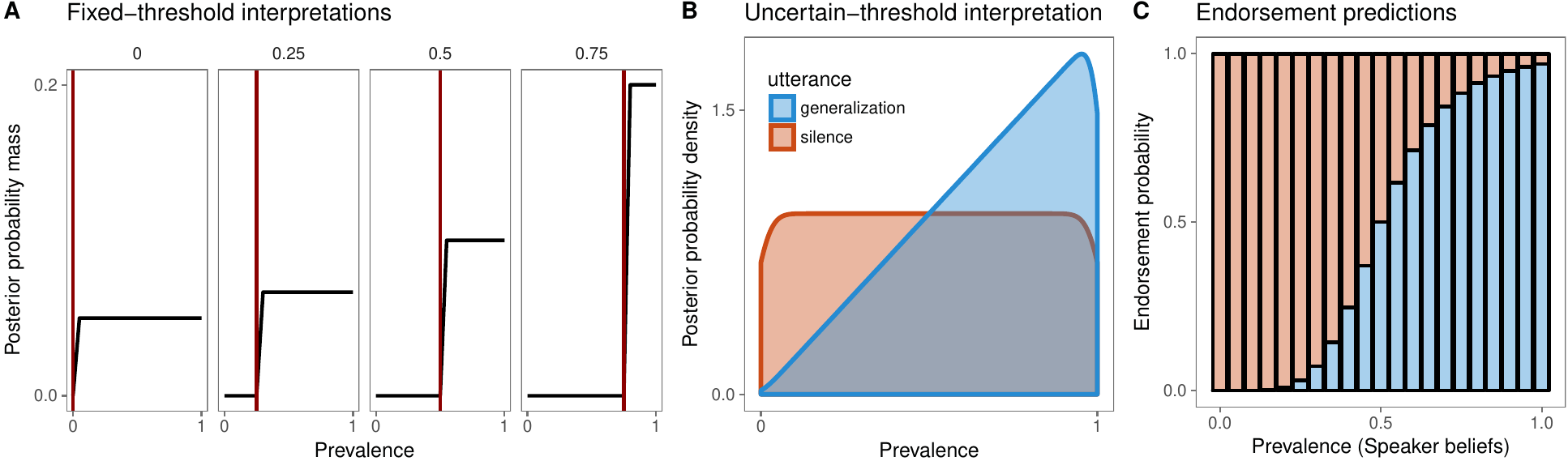} 

}

\caption{Computational model behavior assuming a uniform prior over prevalence. A: Interpretation model posteriors assuming different fixed thresholds (facets). High thresholds rule out more world-states (prevalence levels), leaving fewer world-states among which to distribute the full probability mass. B: Generic interpretation model averages over all thresholds to return a posterior distribution that favors higher prevalence levels in a graded manner. C: Endorsement model predicts higher rates of endorsemens as prevalence levels increase.}\label{fig:uniformPriorSimulation}
\end{figure}

With these ingredients, the literal interpretation model (Eq.
\ref{eq:L0}) computes a posterior interpretation distribution on
prevalence by considering different possible thresholds \(\theta\).
Figure~\ref{fig:uniformPriorSimulation} shows the model behavior
assuming a uniform prevalence prior \(P(p)\), representing a purely
abstract property for which an interpreter has no substantive prior
knowledge. If the threshold were to be very high (e.g.,
\(\theta = 0.75\)), only the highest prevalence levels would be
consistent with the utterance. As the threshold decreases, more and more
prevalence levels become compatible with the threshold
(Figure~\ref{fig:uniformPriorSimulation}A). The interpretation model
weights the different thresholds by the probability that each is true,
resulting in a posterior interpretation distribution that favors higher
prevalence levels (Figure~\ref{fig:uniformPriorSimulation}B). This
interpretation distribution captures the intuition that generalizations
imply high prevalence, though with substantial uncertainty about the
precise quantity.

The model in Eq \ref{eq:L0} describes a general mechanism of
belief-updating via an uncertain threshold semantics. The semantic scale
is determined by the prevalence prior \(P(p)\). For generalizations
about categories (i.e., generic statements), \(P(p)\) ranges over
\([0, 1]\); \(p\) denotes the chance that an instance of a category has
a Boolean property. To apply this same model to generalizations about
events (i.e., habituals like \enquote{John runs}), we use a different,
but related, scale: the frequency within a given time window (e.g.,
number of times John went for a run in the past week).\footnote{This
  measure is closely related to the instantaneous probability density of
  the event, as the time window gets infinitesimally small.} For events
then, \(p\) is a rate and \(P(p)\) ranges over \([0, \infty)\).

\hypertarget{endorsement-model}{%
\subsubsection{Endorsement model}\label{endorsement-model}}

In this paper, we are interested in explaining truth judgments (e.g.,
\enquote{Robins lay eggs} is intuitively true; \enquote{Robins are
female} is not). We model the endorsement task (e.g., \emph{true} vs.
\emph{false}; \emph{agree} vs. \emph{disagree}) as a \emph{speaker
decision} about whether or not to produce the generalization to a naive
listener; thus, the endorsement model has two alternative utterances:
produce the generalization vs.~a null alternative (Degen \& Goodman,
2014; Franke, 2014). For simplicity, we take the null alternative to be
an informationless \enquote{silent} utterance which is always true:
\(\mbox{ $[\![ null ]\!]$}(p, \theta) = \text{True}\).\footnote{The null
  alternative can be realized in at least two other ways: the negation
  of the generalization (e.g., \enquote{It is not the case that Robins
  lay eggs}) or the negative generalization (i.e., \enquote{Robins do
  not lay eggs}). All results reported are similar for these two
  alternatives, and we use the alternative of the \enquote{silent}
  utterance for simplicity.} This kind of speaker model defined in terms
of a listener model formalizes the basic aspects of communicative
reasoning and is the simplest instantiation of a Rational Speech Act
model (Frank \& Goodman, 2012; Goodman \& Frank, 2016).

The endorsement model (called \(S\) for speaker) decides whether or not
to produce the generalization by reasoning about how the listener model
(\(L\); Eq. \ref{eq:L0}) would interpret it: \begin{equation}
S(u \mid p) \propto (\int_{\theta} L(p, \theta \mid u)  \mathop{}\!\mathrm{d}\theta)^\lambda
\label{eq:S1}
\end{equation}

The goal of the endorsement model is to align the listener's \emph{a
priori} beliefs about prevalence (given by the prevalence prior) with
the speaker's beliefs about the prevalence of the feature in the
referent category \(p\), the \emph{referent prevalence} (e.g., that 50\%
of robins lay eggs).\footnote{A more general version of this model can
  relax the assumption that the endorser / speaker has access to a
  specific prevalence \(p\) that it wants to communicate. Rather, the
  endorser may have probabilistic beliefs about the prevalence of the
  property for the referent category \(k\), which would be represented
  by a distribution over \(p\): \(P(p \mid k)\). In this situation, we
  would define the endorsement model decision to be with respect to the
  expected value of the informativity, which integrates over the
  endorsement model's belief distribution:
  \(S(u \mid k) \propto \exp{(\lambda \cdot {\mathbb E}_{p\sim P_{k}} \ln{ \int_{\theta} L(p, \theta \mid u)} \mathop{}\!\mathrm{d}\theta )}\).
  For the empirical case studies described below, these two versions of
  the model make almost identical predictions.} The endorsement model
makes an approximately rational (with degree of rationality
\(\lambda\)), information-theoretic decision based on its beliefs about
which utterance would best achieve the goal of conveying the referent
prevalence \(p\). The semantic threshold \(\theta\) is necessary for
establishing the truth conditions and deriving an interpretation (used
in Eq. \ref{eq:L0}) but is otherwise a nuisance parameter, which the
endorsement model integrates out (note that there is no \(\theta\) in
the left-hand side of Eq. \ref{eq:S1}). Given that the speaker only has
two options (i.e., produce the generalization utterance or stay silent),
the endorsement decision comes down to whether or not the referent
prevalence is more likely under the prevalence prior distribution
\(P(p)\) (i.e., the listener's posterior upon hearing silence) or the
listener generic interpretation distribution \(L(p \mid u)\) given by
Eq. \ref{eq:L0} (Figure~\ref{fig:uniformPriorSimulation}C). In the next
section, we work through different examples from the linguistics
literature, describing the prevalence prior and other relevant model
components for each.

\begin{figure}

{\centering \includegraphics[width=1\linewidth]{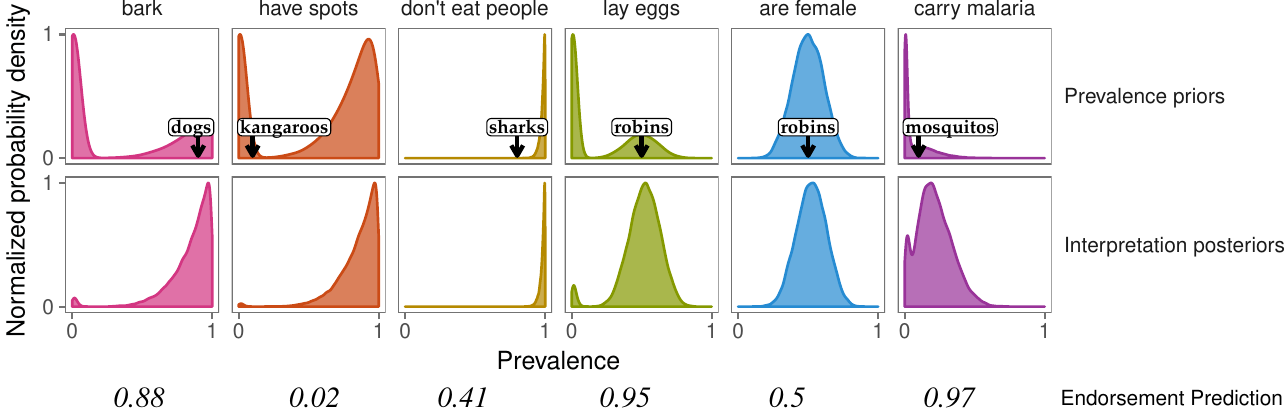} 

}

\caption{Model simulations assuming different prevalence priors. Top: Prevalence priors for six example features. Shapes of the priors were chosen to intuitively correspond to the properties labeling the distributions. Arrows show referent-prevalence for a target category. Bottom: Interpretation model posterior distributions over prevalence upon hearing a generalization about a novel category. Numbers at bottom correspond to endorsement model predictions for endorsing the generalization for the referent-category whose prevalence is shown in the top facets (e.g., "Mosquitos carry malaria").}\label{fig:simulations}
\end{figure}

\hypertarget{worked-examples}{%
\subsection{Worked examples}\label{worked-examples}}

We explore the predictions of the endorsement model in the context of a
few worked examples. These examples are taken from the linguistics
literature on generics and are statements that have been historically
challenging for prevalence-based approaches (see Table 1). We
implemented these and all subsequent Bayesian models in the
probabilistic programming language WebPPL (Goodman \& Stuhlmüller,
2014). All models, analyses, data and links to experiments used in this
paper can be found at \url{https://github.com/mhtess/genlang-paper}.

Endorsement predictions depend upon the uncertain threshold model's
interpretations, which are highly sensitive to the interpretation
model's background knowledge. Background knowledge may be richly
structured, reflecting intuitive theories about the underlying causes of
different kinds of properties (cf., Leslie, 2007). Our model posits,
however, that the only impact of structured knowledge on truth judgments
is in their implications for beliefs about prevalence, formalized in
terms of a prevalence prior \(P(p)\). The prevalence prior reflects a
listener's beliefs about the prevalence of the feature in an unknown or
unfamiliar category and be conceptualized in a theory-neutral way as a
distribution over the prevalence among alternative categories
(Figure~\ref{fig:simulations}).\footnote{Alternatively, the prevalence
  prior can be conceptualized as a marginal distribution on prevalence
  derived from an intuitive theory formalized in a probabilistic
  language of thought (e.g., Goodman et al., 2015). } For example, the
\emph{lay eggs} prevalence prior should be bimodal, with substantial
probability mass near 0-prevalence (since many animal categories do not
have egg-layers) and a secondary component peaked around 50\% (because
among the animals with egg-layers, only the female members of the
category have the property). Conversely, the distribution over the
prevalence of \emph{being female} is unimodal and centered at
50\%-prevalence, because almost all animals have female members in that
proportion. Figure~\ref{fig:simulations} (row 1) shows these and other
intuitive prevalence priors for different properties (we introduce
methods for empirically eliciting these priors in the experimental
section). We now turn our attention to the endorsement model's
predictions.

\begin{table}[ht]
\centering
\begingroup\fontsize{9pt}{10pt}\selectfont
\begin{tabular}{p{2in}|  |p{1.25in}|p{0.75in}|p{2in} |}
  \hline
{\bfseries Example} & {\bfseries Intuitive referent prevalence} & {\bfseries Intuitive truth value} & {\bfseries Issue} \\ 
  \hline
1. Dogs bark & 95\% & True & Not all dogs bark \\ 
   \hline
2. Kangaroos have spots & 5\% & False & Some kangaroos could have spots \\ 
   \hline
3. Robins lay eggs & 50\% & True & Only female robins (50\%) lay eggs \\ 
   \hline
4. Robins are female & 50\% & False & Same number that lay eggs (50\%) \\ 
   \hline
5. Mosquitos carry malaria & 5\% & True & Very few mosquitos actually carry malaria \\ 
   \hline
6. Sharks don't eat people & 95\% & False & Most sharks don't eat people \\ 
   \hline
7. Mary handles the mail from Antarcita & 0 times & True & Consider there has never been any mail from Antarcita \\ 
   \hline
8. Supreme Court Justices have even social security numbers & 100\% & False & Consider 100\% have even social security numbers \\ 
   \hline
9. Elephants live in Africa and Asia & 50\%/50\% & True & Impossible for most to live in Africa and most to live in Asia, and no individual elephant lives in both Africa and Asia \\ 
   \hline
\end{tabular}
\endgroup
\caption{Example sentences that a theory of the language of generalizations should correctly predict. See details of each example in corresponding section.} 
\end{table}

\textbf{1. Dogs bark.}

Often the first observation with generics is that they appear to behave
like universal quantifiers (e.g., \enquote{All}) that permit exceptions:
Not all dogs bark, but still \enquote{Dogs bark} is true. The uncertain
threshold account immediately accomodates this example because there is
no single, fixed threshold beyond which a generic statement becomes
true. Rather, listeners have uncertainty about the threshold which leads
to graded interpretations (e.g., \enquote{Dogs bark} means almost all
dogs bark; Figure~\ref{fig:simulations}; column 1, row 2). The
endorsement model then predicts \enquote{Dogs bark} is a rather good
generic sentence (Figure~\ref{fig:simulations}, bottom row) even though
there are exceptions to the universal generalization (referent
prevalence of barking among dogs roughly 95\%, shown with arrow in
Figure~\ref{fig:simulations} row 2).

\textbf{2. Kangaroos have spots.}

A second observation regarding generics is that their truth conditions
cannot be so lenient so as to always convey existential quantification
(e.g., \enquote{Some}). Indeed, it's intuitively plausible that some
kangaroos do have spots but a rational language user might feel awkward
to assert the generic \enquote{Kangaroos have spots}. The model exhibits
this same restraint, predicting a very low endorsement probability for
this generic sentence (Figure~\ref{fig:simulations}; column 2). The
reason is that, given the interpretation model's background knowledge
about the property, the statement \enquote{Kangaroos have spots} will be
interpreted like \enquote{Dogs bark}: it would mean almost all kangaroos
have spots. The endorsement model, which believes that very few
kangaroos have spots, then would rather not endorse the statement
because it would lead to a too-strong interpretation and mislead the
listener.

\textbf{3. and 4. Robins lay eggs vs.~Robins are female}

One of the most difficult examples for a theory of generics based on
prevalence to handle is the intuitive difference in truth value between
the statements \enquote{Robins lay eggs} and \enquote{Robins are
female}. The former is intuitively true, even though only female robins
could lay eggs (and hence, the prevalence is roughly 50\%). However, the
same implicit restriction to only females does not seem occur for the
latter statement, as \enquote{Robins are female} strikes most as strange
or false. Why is \enquote{Robins lay eggs} a reasonable utterance while
\enquote{Robins are female} is not?

The prior distributions over the prevalence of both features are shown
in Figure~\ref{fig:simulations} (columns 5 and 6). As described above,
the priors are different: Many animals have zero egg-layers (0\%
prevalence), while the vast majority of animal categories have female
members in exactly the same proportion (50\%). Given this background
knowledge, the generic interpretation model returns roughly the same
interpretation distribution for each hypothetical utterance: In each
case, the model believes roughly 50\% have the property. However, only
in the case of \enquote{Robins lay eggs} does the endorsement model
actually assert the generic; it does so because the listener
interpretation would be more aligned with the referent prevalence in
comparison to the interpretation of the null utterance, which is the
prevalence prior. In contrast, the hypothetical generic interpretation
of \enquote{Robins are female} is not different from the prevalence
prior and hence the generic conveys no new information; here, the
generic would not be misleading but uninformative, and hence the model
predicts an endorsement probability of 0.5. Indeed, previous studies on
generic endorsements have found \enquote{Robins are female} to not be
rated as completely false but rather receive an intermediate endorsement
level (i.e., neither true nor false; Prasada et al., 2013).

\textbf{5. Mosquitos carry malaria.}

The statement \enquote{Mosquitos carry malaria} is intuitively true in
spite of the fact that the vast majority of mosquitos are actually
malaria-free. The prevalence prior for \emph{carries malaria} is highly
skewed towards low prevalence levels: many animals do not have
malaria-carriers among them and even for those that do, the prevalence
is expected to be quite low. Then, carrying malaria is significantly
more true of mosquitos than other animals, an intuition that is often
arrived at with this example. This kind of behavior is related to the
construct of \emph{cue validity}---the probability of the feature given
the category e.g.,
\(P({is a mosquito} \mid \text{carries malaria})\)---which we return to
in the next section. The uncertain-threshold model displays a critical
behavior: It can endorse generics when the referent prevalence is very
low.

\textbf{6. Sharks don't eat people.}

When the prevalence of the feature is very high, it is not necessary for
the model to endorse the generic. \enquote{Sharks don't eat people} is
predicted to be a somewhat strange utterance, despite the fact that the
vast majority of sharks do not eat people.\footnote{We create this
  example as the converse of the classic example \enquote{Sharks eat
  people}, predicted to be \emph{true} despite low prevalence (Leslie,
  2007). We reverse this example to show how the model deals with a
  high-prevalence feature for which the generic is predicted to be false
  or infelicitous.\\} Because for almost all animal categories, the
prevalence of \emph{not eating people} is almost 100\% (very few things
eat people), interpreting the statement \enquote{Sharks don't eat
people} would lead one to believe that no sharks eat people (i.e., 100\%
don't eat people), which is too strong. It is interesting that because
of the high referent prevalence, the model is less certain of its
decision, predicting an endorsement probability around 0.4 (i.e.,
somewhat false). We will test this quantitative prediction in Experiment
1 on generic language.

\textbf{7. Mary handles the mail from Antarctica (\emph{yet has never
had the opportunity}).}

Imagine there is a job in the local bureaucrat's office to handle the
mail from Antarctica and this job is assigned to Mary; to date, however,
nobody has ever sent mail to the office from Antarctica (Cohen, 1999).
In other words, the statement \enquote{Mary has handled mail from
Antarctica} is false. \enquote{Mary handles the mail from Antarctica},
however, is still thought to be intuitively true despite zero actual
instances of the event. This highlights an important ambiguity in the
theoretical commitments of the uncertain threshold model: The
endorsement model aims to communicate some referent prevalence \(p\),
but does the prevalence represent the actual, objective frequency in the
world (e.g., the number of times in the past that Mary has handled mail
from Antarctica) or a subjective, predictive degree of belief in the
head (e.g., our prediction that were the appropriate situation to arise,
Mary would be handling the mail from Antarctica)?

We posit that Mary the Antarctic mail handler is an extreme case of
generalizations expressing \emph{predictive} degrees of belief. We
expect this context sets up the expectation that any future mail coming
from Antarctica will be handled by Mary. Thus the predictive probability
that Mary will handle Antarctic mail, should there be some, is high.
Predictive probabilities often track past frequency or actual prevalence
in the world, but people's internal models of how the world works can
lead the two to diverge. In this case our understanding of Mary's job
leads to a strong predictive probability in the absence of past
frequency evidence. We explore this question experimentally in
Exerpiment 2 on habituals.

\textbf{8. Supreme Court Justices have even social security numbers.}

Imagine that all current Supreme Court Justice has a social security
number which was an even number. \enquote{Supreme Court Justices have
even social security numbers} is still considered false, even though the
property holds for exactly 100\% of the category (Cohen, 1999). We
predict the rejection of even social security numbered Justices is the
result of people's intuitive theories guiding their subjective
predictive probabilities, which feed into the endorsement model. That
is, we predict observers strongly believe there is no causal relation
between the evenness of one's social security number and selection for
the Supreme Court, and thus would assign a roughly 50\% subjective
probability to the next justice having an even social security number.
Then, \enquote{Supreme Court Justices have even social security numbers}
would be rated by our model as similar to \enquote{Birds are female},
because all professions have roughly the same probability of having
employees with even social security numbers. This example is thus the
conceptual opposite of Mary the Antarctic mail handler, where our
internal model of the world led us to a strong degree of belief in the
property holding in the future; with the Supreme Court Justices, our
internal models lead to a relatively weak degree of belief in the
property holding in the future.

\textbf{9. Elephants live in Africa and Asia.}

Understanding how a semantic representation composes is another
important test for a theory of the language of generalization. Nickel
(2008) suggests that \enquote{Elephants live in Africa and Asia} is
troubling for prevalence-based accounts (in particular,
majority-quantificational accouts where the generic means \emph{more
than half}; Cohen, 1999) because the statement should be semantically
equivalent to \enquote{Elephants live in Africa and elephants live in
Asia}. It cannot be the case that most (more than half of) elephants
live in Africa and most (more than half of) elephants live in Asia,
unless we are to posit international elephants (i.e., individual
elephants who live part-time in Africa and part-time in Asia), which are
intuitively implausible.

\begin{figure}

{\centering \includegraphics[width=0.8\linewidth]{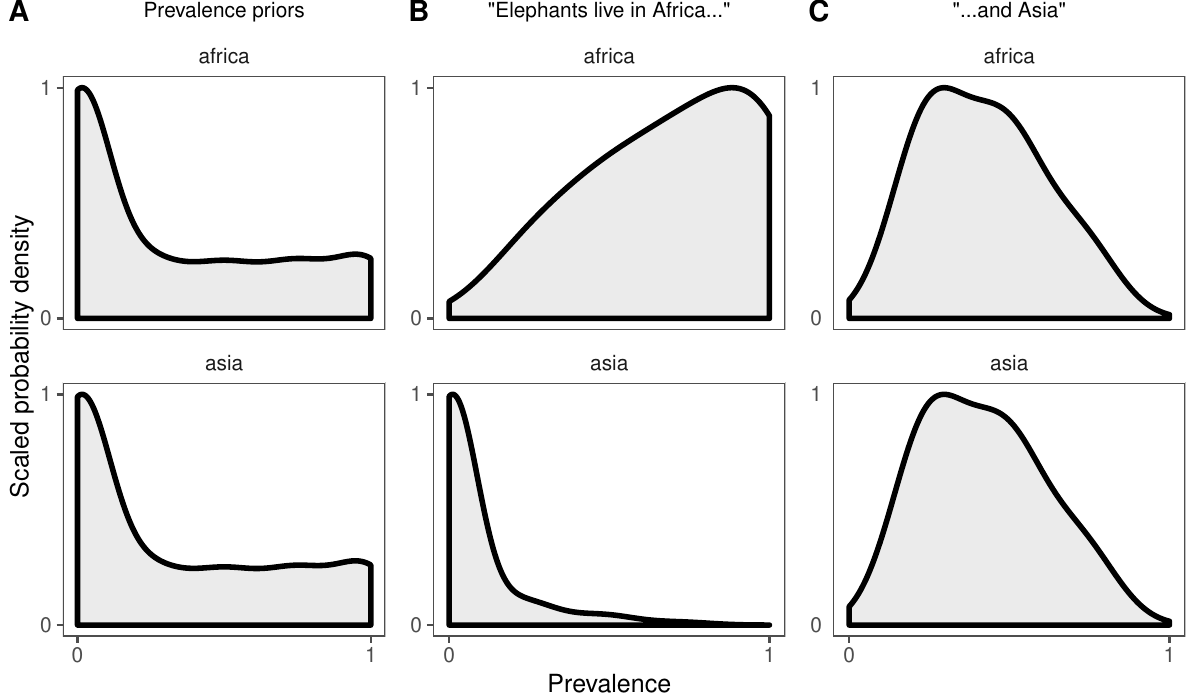} 

}

\caption{Model interpretation of a conjunctive generic ("Elephants live in Africa and Asia"). A: Intuitive priors for marginal distributions of prevalence for living in Africa and living in Africa. B: Partial interpretation upon hearing utternace ("Elephants live in Africa..."). C: Full interpretation upon hearing the end of the sentence ("Elephants live in Africa and Asia"). Uncertainty about the threshold for generic sentences allows the listener to non-monotonoically update its beliefs about the prevalence of the feature in categories.}\label{fig:elephants}
\end{figure}

Our theory does not provide a fixed semantics for generics, but an
uncertain one which can be updated as more information comes in. In
fact, with prior knowledge suggesting against the existence of
international elephants (Figure~\ref{fig:elephants}A), our model
interprets \enquote{Elephants live in Africa and Asia} as meaning that
some elephants live in Africa and that different ones live in Asia
(Figure~\ref{fig:elephants}C). Of theoretical interest, an incremental
parsing of the sentence (i.e., upon hearing only that \enquote{Elephants
live in Africa}) leads our model to believe that most, possibly all,
elephants live in Africa (Figure~\ref{fig:elephants}B). When the
sentence is completed (\enquote{\ldots{}and Asia}), the model
non-monotonically updates its beliefs to something weaker: some
elephants live in Africa and others in Asia
(Figure~\ref{fig:elephants}C). The flexibility of our model accommodates
new evidence that might otherwise contradict previous utterances because
it maintains uncertainty about the precise meaning of the utterance.
From a compositional perspective, this is highly desirable behavior but
we leave the testing of quantitative predictions of this kind for future
work.

\hypertarget{relationship-to-other-semantic-theories}{%
\section{Relationship to other semantic
theories}\label{relationship-to-other-semantic-theories}}

Our formal theory builds on and relates to a number of extant theories
of generics in both the formal semantics and psychological literatures.
The dominant approaches from formal semantics try to describe unified,
objective criteria by which to establish if a generic and other
generalizations are true or false (i.e., generic \emph{truth
conditions}). These views are typically \emph{statistical} in nature in
that they appeal to quantification or the statistics of the world (e.g.,
how many of the kind have the property) to define those unified,
objective criteria. Many of these theories rely upon mechanisms outside
of the truth conditions to make sense of the extreme flexibility
exhibited by generics; unfortunately, these extra-semantic mechanisms
are not described in sufficient detail to generate precise, quantitative
predictions. Other theorists, primarily in philosophy and psychology,
have taken the extreme flexibility of generics to argue against a
quantitative theory based on statistics, instead suggesting that
abstract, mental representations are directly tied to semantics of
generics (e.g., \emph{there is something about being a K which causes it
to F}). Statistical and conceptual theories express the major
contrasting views of generic language (Carlson, 1995).\footnote{We use
  the terms statistical and conceptual to refer to what Carlson (1995)
  referred to as \enquote{inductive} and \enquote{rules and regulations}
  views, respectively.}

\hypertarget{conceptual-accounts}{%
\subsection{Conceptual accounts}\label{conceptual-accounts}}

Conceptual accounts try to identify the core meaning of a generalization
directly with aspects of conceptual structure. The most influential
account in psychology comes from Leslie (2008), which uses the same
starting point as our analysis: Generics express generalization. Leslie
(2008)'s analysis draws upon insights from the psychological literature
on infant generalization to argue that generics tap into an innate,
default mechanism that signals the child (or adult) to generalize the
property to the kind.

The \emph{default generalization} mechanism has three components: the
ability to (i) identify whether or not a feature is a
\emph{characteristic} property of a kind (e.g., one characteristic
property for an animal would pertain to the mode of locomotion for the
animal), (ii) identify whether or not a feature is \emph{striking}
(e.g., can kill you; \enquote{Mosquitos carry the West Nile Virus}), and
(iii) segment \emph{counterinstances} (instances of the kind which do
not have the feature) into positive and negative counterinstances (i.e.,
instances of the kind which have some relevant alternative feature and
instances of the kind which simply lack the feature,
respectively).\footnote{This positive vs.~negative counterinstance
  distinction is a rather technical consideration (though not without
  psychological foundation, see Leslie, 2008, p. 36) used to account for
  the difference between examples like \enquote{Robins lay eggs} vs.
  \enquote{Robins are female}. For \enquote{Robins are female}, male
  robins are positive counterinstances because they have an alternative
  property (being male). For \enquote{Robins lay eggs}, male robins are
  negative counterinstances because they simply lack the property (i.e.,
  male robins do not reproduce by some other method). We will not
  discuss this consideration any further other than to note that that
  the difference between \enquote{Robins lay eggs} and \enquote{Robins
  are female} can be explained by a different mechanism (e.g.,
  informativity with respect to a prevalence prior). The notion that
  alternative features may come into play in generic interpretation may
  also be addressed with prevalence priors constructed with respect to
  the feature (described in the General Discussion).} On this account,
generics are true if some instances of the kind have the feature when
all counterinstances are negative counterinstances and the property is
either characteristic or striking. If the property is neither
characteristic nor striking, then almost all instances of the kind must
have the feature for the generic to be true (e.g., \enquote{Barns are
red}). Many of the key factors in this conceptual approach are
compatible with our view of prevalence as a predictive probability: What
is required is that the distinctions made by conceptual accounts
influence predictive probability, which mediates linguistic effects.
Indeed, one version of this mechanism is already suggested by Leslie
(2008), citing findings by Rothbart, Fulero, Jensen, Howard, and Birrell
(1978), that a speaker's perception of the prevalence of the feature can
be altered by virtue of its dangerousness or distinctiveness.

\hypertarget{default-generalizations}{%
\subsubsection{Default generalizations}\label{default-generalizations}}

The conceptual view of generics argues that it is not desirable to
define generics in the same terms as quantifiers (e.g., a
truth-functional threshold on prevalence), because generics are more
basic or more fundamental than quantifiers (Gelman, 2009; Leslie, 2008).
For example, according to certain measurements, young children have a
nearly adult-like understanding of generics far earlier than they do
with quantified language (Gelman, Leslie, Was, \& Koch, 2015).
Similarly, adults who are told novel information about categories using
quantified language (e.g., \enquote{Most spiders shed their skin}) and
later asked to recall that same information will tend to recall
quantified information as generics (e.g., \enquote{Spiders shed their
skin}), but not visa versa (Leslie \& Gelman, 2012). As a result,
generics are thought of as conveying default generalizations, whereas
quantified language expresses more sophisticated and specific
generalizations. Rather than be contradictory, our model can be seen as
formalizing a default generalization as Bayesian belief updating via a
threshold function whose threshold value is contextually-informed by
knowledge about properties. Alternative formulations of such a default
generalization mechanism could be proposed and quantitatively tested
against our account.

\hypertarget{striking-generics}{%
\subsubsection{Striking generics}\label{striking-generics}}

Leslie (2008) posits a special mechanism to treat generics about
striking properties (e.g., \enquote{Rottweilers maul children}), which
seem to be acceptable even when the prevalence of the property is quite
low (e.g., very few rottweilers maul children). Indeed, Cimpian,
Brandone, and Gelman (2010) found participants willing to endorse
generics about striking properties (e.g., \enquote{Lorches have
dangerous feathers}) more so than that of neutral properties (e.g.,
\enquote{Lorches have purple feathers}) at low levels of prevalence
(e.g., when only 30\% of the category had the property). Leslie (2008)'s
argument for accepting striking generics is straightforward: These are
relevant properties to know about if you want to survive. Our theory
posits two mechanisms by which striking properties could influence
generic endorsement: the prevalence prior and the referent prevalence.

The first observation is that striking properties are relatively rare in
the environment: Most animals do not maul children, eat swimmers, or
carry malaria. Thus, the prevalence prior distributions of these
features may be hard to distinguish from properties that are just
generally rare and distinctive, as we have assumed in our worked example
of \enquote{Mosquitos carry malaria}. Cimpian et al. (2010)'s
experiments did not measure the prevalence prior distribution, but we
have found in pilot work that the prevalence prior distribution changes
to resemble that of a distinctive property when participants are
supplied information about the dangerousness of a property.

The second mechanism by which strikingness could influence generic
endorsement in our model is by speakers having a distorted perspective
of how prevalent these features are within the referent category. That
is, striking properties may be projected more strongly (i.e., higher
predictive probability that future instances will have the property)
than neutral properties.\footnote{This observation is also made by
  Leslie (2008) (p.42), who uses it as motivation for elevating striking
  properties to their special status in her theory.} Interestingly,
evidence suggests this enhanced projectibility holds for both dangerous
properties (e.g., people doing criminal actions) as well as neutral,
distinctive properties (e.g., people taller than 6'5", Rothbart et al.,
1978 Expts. 2 \& 3). Thus, there is evidence for the influence of
strikingness on predictive probability, which we posit is an
intermediate representation between conceptual knowledge and generic
endorsements. It is beyond the scope of this paper to directly attempt
to empirically distinguish these two potential mechanisms, though our
experiments use a number of striking properties for which we measure the
prevalence prior and referent prevalence.

\hypertarget{principled-and-statistical-connections}{%
\subsubsection{Principled and statistical
connections}\label{principled-and-statistical-connections}}

Leslie (2008)'s construct of characteristic properties is similar to
Prasada and Dillingham (2006)'s notion of a \emph{k-property} (\emph{k}
for kind), a property that bears a \emph{principled connection} to the
kind. A property bears a principled connection to the kind if a
(generic) sentence that appeals to the kind to explain the existence of
the property makes sense (e.g., \enquote{Dogs, by virtue of being the
kind of things that they are, are four-legged}). When such sentences are
not endorsed (e.g., \enquote{Barns, by virtue of being the kind of
things that they are, are red}) while the simple generic (e.g.,
\enquote{Barns are red}) is endorsed, then the property is thought to
bear a \emph{statistical connection} to the kind (a so-called
\emph{t-property}, for emphasizing that it is the tokens having the
property which matters for generic endorsement). In addition to the
difference in endorsements for \emph{by virtue of} statements, Prasada
et al. (2013) showed that different kinds of generic statements (e.g.,
striking properties, characteristic properties with low prevalence)
support different kinds of inferences (e.g., \emph{normative}:
\enquote{Dogs are supposed to have four legs}, \emph{aspect}:
\enquote{Having four legs is one aspect of being a dog}). Additionally,
there is some evidence that the impact of principled connections on
interpretations is separate from their influence on beliefs about
prevalence (Prasada \& Dillingham, 2006).

Both Prasada and Dillingham (2006) and Leslie (2008) surmise that a
property may qualify as characteristic (or having a principled
connection) if there exists an overhypothesis about that property for a
relavant superordinate category (e.g., each kind of animal has a means
for self-locomotion; Goodman, 1955; Shipley, 1993). Overhypotheses are
naturally formalized as hierarchical Bayesian models, wherein a learner
acquires knowledge at multiple levels of abstraction (e.g., learning
from the same event about a particular dog, dogs in general, and animals
in general; Kemp et al., 2007). These Bayesian heirarchical models yield
differences in predictive probability that are particularly robust.
Integrating a hierarchical model of kinds and properties with our model
of generic language is a natural direction to understanding the
computational underpinnings of conceptual relations and generics.
Different conceptual structures may give rise to roughly the same
distributions on prevalence and thus have similar generic endorsement
profiles according to our model. Yet the inferences that can be drawn
from hearing these generics may differ, depending on their interactions
with the putative hierarchical knowledge people bring to bear (a la the
differences observed for ``prevalence-matched items'' in Prasada \&
Dillingham, 2006). Building hierarchical models of kinds and properties
is a major undertaking in its own regard, but our formalism provides a
way of connecting such a model with generic language.

\hypertarget{statistical-accounts}{%
\subsection{Statistical accounts}\label{statistical-accounts}}

\hypertarget{relative-and-absolute-generics}{%
\subsubsection{Relative and absolute
generics}\label{relative-and-absolute-generics}}

Our underspecified threshold model has clear antecedents in other
statistical accounts, most notably Cohen (1999)'s theory of generics as
a frequency adverb (e.g., \enquote{generally}). Cohen treats generics as
a class comprising two qualitatively different types: \emph{relative}
and \emph{absolute} generics. \emph{Absolute generics} use a fixed, 50\%
threshold on prevalence: \(p>0.5\). That is, if a particular instance is
more likely than not to have the feature, then an absolute generic is
true. By contrast, \emph{relative generics} are true based on a
comparison to an alternative set of kinds \(Alt(K)\), analogous to the
categories that comprise the prevalence prior (the so-called
\emph{comparison class}, which we discuss further in the General
Discussion). \enquote{Mosquitos carry malaria} is true because an
arbitrary mosquito is more likely than an arbitrary member of an
alternative kind to have the feature.

In our model, we treat all generics as relative. Attested differences in
endorsement, then, emerge through the interplay of prior knowledge with
our uncertain semantics.\footnote{For a related theoretical argument
  against the \enquote{relative} / \enquote{absolute} distinction for
  gradable adjectives, see Lassiter and Goodman (2015).} Further, though
Cohen's theory is framed in terms of probabilities, it is a fully
deterministic, fixed-threshold account which only makes qualitative
predictions about what is true and what is false (i.e., it has a
deterministic semantics). In contrast, we propose a fully probabilistic
semantics embedded within a Bayesian model that describes how context
resolves the uncertain threshold; our theory is a joint
semantic--pragmatic theory.

Cohen's and other statistical theories employ a mechanism (not currently
required in our account) known as \emph{domain restriction} to explain
the context-sensitivity of generics: contextually restricting the
entities that go into the computation of prevalence (i.e., which robins
do we look at to compute the probability of laying eggs among robins?).
Cohen (1999) posits that prevalence is calculated by only considering
entities that \emph{could have some feature} in a contextually-specified
alternative set of features \(Alt(F)\). For example, the property
\emph{lays eggs} induces a set of alternatives that are associated with
modes of reproduction (e.g., \emph{gives birth to live young},
\emph{undergoes mitosis}, \ldots{}). \enquote{Robins lay eggs} (an
absolute generic for Cohen, 1999) is evaluated by only considering
female members of kinds, because only female members can plausibly
satisfy one of the other reproductive property alternatives (i.e., the
alternative features in \(Alt(F)\)). The inferential machinery behind
domain restriction ---how to determine \(Alt(F)\)---relies upon
conceptual knowledge, but the details remain obscure (Carlson,
1995).\footnote{However, see Cohen (2004) for a discussion of how his
  semantic constraints relate to different kinds of generics and
  different kinds of conceptual representational frameworks used in
  cognitive science.} Our uncertain threshold model can be seen as one
particular mechanism by which the domain may be restricted: The
structure of the prior distribution over the prevalence of \emph{lays
eggs} is a reflection of an intuitive theory of reproduction (i.e., that
only females lay eggs) and the uncertain threshold model uses that
background knowledge to derive a property-specific interpretation. There
may exist a refactorization of the uncertain threshold model to a fixed
threshold where the listener has uncertainty about the relevant domain
of restriction.

\hypertarget{generic-as-indexical}{%
\subsubsection{Generic as indexical}\label{generic-as-indexical}}

Sterken (2015) develops a novel analysis taking the context-sensitivity
of generics as primary. This analysis draws analogy to other, inherently
context-sensitive linguistic expressions: \emph{indexicals} (e.g.,
\enquote{this} or \enquote{I}). Sterken (2015) uses this analogy to
motivate a context-sensitive \emph{quantificational force} as well as
mechanism of domain-restriction (of the kind used by Cohen, 1999 and
others). Our uncertain threshold semantics can be seen be a particular
formalization of Sterken (2015)'s context-sensitive quantificational
force. We have not had need for employing domain restriction on the
categories, though as noted above, it is a potential avenue for future
development.

\hypertarget{normal-accounts}{%
\subsubsection{\texorpdfstring{\enquote{Normal}
accounts}{``Normal'' accounts}}\label{normal-accounts}}

A popular alternative view under the statistical banner draws on the
intuition that generics often express something normative in the world
(Asher \& Morreau, 1995; Nickel, 2008, 2016; Pelletier \& Asher, 1997).
\enquote{Dogs have four legs} is then a good generic not because all
dogs have four legs (regrettably, all do not) but were the world to
function normally (e.g., dogs would not be involved in freak tractor
accidents or be born with strange genetic mutations), then all dogs
would have four legs. The idea that our beliefs about what is normal in
the world influences our judgments about generalizations has intuitive
appeal for rejecting accidentally true generics (e.g., \enquote{Supreme
Court justices have even social security numbers}) and resisting
stereotyped language (e.g., \enquote{Boys are good at math}). Our theory
does not directly formalize what is \emph{normal}, though we argue that
a speaker's beliefs about what is probable (which may relate to what is
normal; see Icard, Kominsky, \& Knobe, 2017) plays a role in endorsing
and interpreting generalizations.

\hypertarget{underquantification}{%
\subsubsection{Underquantification}\label{underquantification}}

A proposal similar to our account concerning underspecification of
generics has been made in the computational linguistics literature
(so-called \emph{underquantification}; Herbelot \& Copestake, 2011). In
their model, generics express an explicit quantified relation,
specifically either \enquote{Some}, \enquote{Most}, or \enquote{All}.
This proposal is used to construct a set of features that accurately
predicts (relative to human judgments) the quantified relationship
expressed by the generic (analogous to a quantifier version of the
\emph{implied prevalence} task used in Gelman \& Raman, 2003; Cimpian et
al., 2010). Our semantic theory can be seen as a generalization of
\emph{underquantification} to a continuous interval of possible
meanings. This distinction is relevant for the acquisition of the
language of generalizations; we do not take as primary the quantified
relations (e.g., \enquote{Some}, \enquote{All}). Additionally, by using
an underspecified threshold on a scale of probability, our formulation
naturally extends to other scales and other kinds of generalizations
(e.g., habitual language), where quantified relations like
\enquote{Most} or \enquote{All} are not directly applicable.

\hypertarget{cue-validity}{%
\subsubsection{Cue validity}\label{cue-validity}}

By encoding knowledge about other categories, the prevalence prior
distributions in the uncertain threshold model are deeply connected to
the construct of \emph{cue validity}, or the probability of the kind
given the feature: \(P(x \in k \mid x \in f)\) (e.g., one's predictions
about whether or not an entity is a mosquito, upon learning that it
carries malaria). Cue validity is believed to play a role in
understanding generic language (Khemlani, Leslie, \& Glucksberg, 2012;
Leslie, 2007; Prasada et al., 2013), but the details remain
underspecified. The strongest view of cue validity is that it
operationalizes an alternative hypothesis about the meaning of generic
statements: \enquote{Mosquitos carry malaria} means \enquote{It is
mosquitos that carry malaria}. Indeed, empirically-elicited cue validity
has been shown to be highly correlated with endorsements of generics
(Khemlani et al., 2012).

Cue validity is inverse prevalence; the two are related via Bayes' Rule:
\(P( k \mid f) = \frac{P( f \mid k) \cdot P( k)}{\sum\limits_{k' \in K} P( f \mid k') \cdot P( k')}\).
Knowledge about other categories \(k'\) enters in the denominator to
compute cue validity for prevalence. Indeed this normalizing constant
(the denominator) is equal to the expected value (i.e., the mean) of the
prevalence prior distribution: \(\mathbb{E}[P(p)]\). Thus cue validity
comes from a point estimate of the prevalence prior distribution, and
information about cue validity can be derived from the constructs
posited in our model. We return to the implications of this relationship
in the General Discussion. For a more detailed, mathematical derivation
of the relationship between cue validity and prevalence priors, see
Appendix A.

\hypertarget{baseline-models-for-quantitative-comparisons}{%
\subsection{Baseline models for quantitative
comparisons}\label{baseline-models-for-quantitative-comparisons}}

In our empirical studies below, we compare our model to three
alternatives. These alternative models do not represent any of the
extant theories of generics described above; no extant theory is
sufficiently precise to yield quantitative predictions. Instead, these
models are designed to interrogate the theoretically-substantive
components of our model. There are three such components: (a) property
knowledge in the form of a prior distribution over prevalence \(P(p)\),
(b) endorsement as a decision-theoretic process of uttering the
generalization vs.~not uttering it, and (c) vagueness in the semantics
of a generalization (i.e., an uncertain threshold). In our empirical
studies, we compare the uncertain threshold model to an alternative,
lesioned model which lacks the vagueness in meaning, assigning a fixed
semantics to the generalization (i.e., analogous to a quantified
statement) but which has the same prevalence prior and the
decision-theoretic architecture. There are no correspondingly simple
ways to lesion the other two components (property knowledge or speaker
decision) that still produces quantitative, context-sensitive
predictions. Instead, we compare our model to two regression models
based on empirically elicited referent prevalence and cue validity,
which provides some interrogation of the necessity of property knowledge
in the form of a full distribution on prevalence. In addition to serving
as alternatives, these baseline regression models help us understand the
statistical properties of our experimental materials and provide a
comparison to standard methods in the psychological literature on
generics (e.g., Khemlani et al., 2012).

\hypertarget{interim-summary-and-overview-of-experiments}{%
\section{Interim summary and overview of
experiments}\label{interim-summary-and-overview-of-experiments}}

We have introduced the first quantitative theory of the language of
generalization, and discussed the relationship of this account to extant
theories of generics. Above, we presented simulations showing how the
model predicts endorsements for classically puzzling generic statments.
These predictions depended upon the background knowledge about the
property \(P(p)\) as well as the referent prevalence \(p\) believed to
be true for the category. For each of these we chose intuitive values
for the parameters of the model (i.e., the prevalence priors and
referent prevalence levels). In what follows, we test this theory
empirically for a wide range of generalizations, including
generalizations of different types (categories, events, and causes). We
do this by both measuring and manipulating background knowledge and
referent prevalence and predicting human endorsements of
generalizations.

In Case Study 1, we examine generalizations about categories expressed
in generic language. We first measure endorsements for thirty generic
statements about familiar categories, revealing an entire continuum of
endorsements (Expt. 1a). We then measure the corresponding prevalence
priors and referent prevalence using a prevalence prior elicitation task
(Expt. 1b). We compare the quantitative fits of our model to the three
alternative models described above. This Case Study is an empirical
version of several of the worked examples in the section above.

In Case Study 2, we examine generalizations about events expressed in
habitual language while manipulating the referent prevalence. We measure
prevalence priors for events of people doing various actions (e.g.,
people running, hiking, climbing mountains; Expt. 2a). We then measure
endorsements of habituals for statements about novel actors (e.g.,
\enquote{John runs}) given referent prevalence information (e.g.,
\enquote{In the last two months, John ran three times.}; Expt. 2b).
Finally, we test whether referent prevalence in our model is best
thought of as a past frequency (e.g., the number of times John has run
in the past) or a prediction about the future (e.g., the number of times
a speaker expects John to run in the future) by experimentally
manipulating future predictions while keeping constant past frequency
(Expt. 2c). We answer this question by testing the quantitative fits of
two versions of our model: one in which the speaker is conveying past
frequency and one in which the speaker is conveying their predictions
about the future.

Case Study 3 experimentally manipulates the prevalence prior in the
domain of causal language (e.g., \enquote{Herb X makes animals sleepy}).
Expt. 3a measures the (manipulated) prevalence prior, confirming that
our manipulation influenced participants' beliefs about the prevalence
of the feature across different categories. Expt. 3b measures the
corresponding influence on causal endorsement, finding an effect of the
manipulated background knowledge in the way predicted by the uncertain
threshold model.

\hypertarget{case-study-1-generic-language}{%
\section{Case Study 1: Generic
Language}\label{case-study-1-generic-language}}

Learning from generic language (i.e., generalizations about categories;
e.g., \enquote{Dogs bark.}) is believed to play a central role in
concept and theory formation (e.g., Gelman, 2004), stereotype
propagation (Rhodes et al., 2012), motivation (Cimpian et al., 2007),
and many other facets of everyday reasoning. In addition, generics have
been the case study of choice for the semantics of the language of
generalization because of their tantalizing similarity to quantified
statements (e.g., \enquote{Most dogs bark}). However, intuitions and
empirical data argue that generics simply do not reduce to quantified
statements in a simple way (e.g., Khemlani et al., 2012; Cimpian et al.,
2010; Prasada et al., 2013).

We first investigate how the uncertain threshold endorsement model
predicts actual human endorsements of generic statements. We measure
endorsement for thirty generic sentences that cover a range of
conceptual distinctions previously discussed in the empirical literature
on generics (Prasada et al., 2013): characteristic features displayed by
a majority (e.g., \enquote{Ducks have wings.}), characteristic features
displayed by a minority (e.g., \enquote{Robins lay eggs.}), features
that are striking or dangerous (e.g. \enquote{Mosquitos carry
malaria.}), noncharacteristic features displayed by a minority (e.g.,
\enquote{Robins are female.}), and features that are totally absent
(e.g. \enquote{Lions lay eggs.}). We further craft sentences with the
goal of eliciting the full range of acceptability judgments
(intuitively: \emph{true}, \emph{false}, and \emph{indeterminate}) for
generics with properties of low, medium, and high referent-prevalence
(Expt. 1a). We examine generics about animal categories in order to
reliably measure the prior belief distribution over the prevalence of
features \(P(p)\). The prevalence elicitation procedure (Expt. 1b)
includes measurements of the referent-prevalence \(p\) for different
categories (e.g., \(P(x \text{ lays eggs} \mid x \text{ is a robin})\)),
allowing us to generate predictions for the endorsement model (Eq.
\ref{eq:S1}) as well as for simpler, alternative models.

\hypertarget{experiment-1a-generic-endorsements}{%
\subsection{Experiment 1a: Generic
endorsements}\label{experiment-1a-generic-endorsements}}

In this experiment, we elicit human endorsements for generic sentences
taken from the linguistic and psychological literatures (Prasada et al.,
2013). The goal of this study is to elicit high variability of
endorsements for generic statements about animal categories.

\hypertarget{method}{%
\subsubsection{Method}\label{method}}

\hypertarget{participants}{%
\paragraph{Participants}\label{participants}}

We recruited 100 participants over Amazon's crowd-sourcing platform
Mechanical Turk (MTurk). Participants were restricted to those with US
IP addresses and with at least a 95\% MTurk work approval rating (the
same criteria apply to all experiments reported). Four participants were
excluded for failing to recall the button corresponding to
\emph{agreement} in the forced-choice task. Five participants
self-reported a native language other than English; removing their data
has no effect on the results reported. The experiment took about 3
minutes and participants were compensated \$0.35.

\hypertarget{procedure-and-materials}{%
\paragraph{Procedure and materials}\label{procedure-and-materials}}

Participants were shown thirty generic sentences in a randomized order.
They were asked to press one of two buttons (\emph{P} or \emph{Q};
randomized between-participants) to indicate whether they agreed or
disagreed with the sentence (see
Figure~\ref{fig:generics-endorsement-figure}A for the full list). The
thirty sentences covered a range of conceptual categories described
above. Approximately ten true, ten false, and ten uncertain \emph{a
priori} truth-value generics were selected. As an attention check,
participants were asked at the end of the trials which button
corresponded to \enquote{Agree}. Four participants were excluded for
failing this trial.

\hypertarget{results}{%
\subsubsection{Results}\label{results}}

As a manipulation check, the first author assigned an \emph{a priori}
truth judgment (true/false/indeterminate) to each stimulus item. As one
would expect, there were substantial differences in empirical
endorsements: true generics were almost universally endorsed (Maximum
A-Posteriori estimate and 95\% credible interval of endorsement
probability: \(0.93\) \([0.91, 0.94]\)); indeterminate generics were
endorsed at a rate \emph{less} likely than chance (\(0.38\)
\([0.35, 0.42])\)) but substantially more than false generics (\(0.08\)
\([0.06, 0.09])\)).

Ideally, a complete theory of genericity should be able to explain
statements that are endorsed completely, rejected completely, and the
gradedness between the extremes. We observe gradedness among our thirty
examples covering a continuum of endorsement values
(Figure~\ref{fig:generics-endorsement-figure}A). Such a continuum of
judgments is already evidence against any theory that only predicts
categorically whether a generic statement is true or false. We next
measure the prevalence prior distributions and use them to articulate a
set of quantitative models that try to predict this quantitative
variability in endorsements.

\hypertarget{experiment-1b-prevalence-prior-elicitation}{%
\subsection{Experiment 1b: Prevalence prior
elicitation}\label{experiment-1b-prevalence-prior-elicitation}}

The prevalence prior \(P(p)\) in Eq. \ref{eq:L0} describes the belief
distribution on the probability of a given feature (e.g.,
\textsc{lays eggs}) across relevant categories. To get an intuition for
the kind of knowledge encoded in this belief distribution, imagine you
are walking outside and come across an instance of your favorite kind of
animal (e.g., a reindeer). What is the chance it is female? Your answer
will probably depend upon the percentage of the category that you
believe to \emph{be female} (e.g., the percentage of female reindeer,
approximately 50\%). What is the chance that it lay eggs? Again, this
depends upon the percentage of the category that you believe \emph{lays
eggs}, which then further depends on the particular kind of animal under
consideration: If you're thinking of a reindeer, the answer is probably
0\%; if you're thinking of a peregrine falcon, the probability is
similar to the \emph{being female} probability (50\%) because female
peregrine falcons lay eggs. That is, the answer to how many of an
arbitrary category is likely to lay eggs is either roughly 50\% or 0\%,
depending on the kind of creature you may bring to mind.\footnote{The
  subjective probability may, in fact, be non-zero and small as opposed
  to 0. Non-zero probabilities allow for the intuitive possibility of a
  reindeer that, by some terribly improbable set of circumstances such
  as a genetic mutation, lays eggs.}

The thought experiment decomposes the prevalence prior \(P(p)\) into a
prior distribution on kinds \(P(k)\) and then a conditional probability
of the prevalence given the kind \(P(p \mid k)\). This decomposition can
be used to measure the prevalence prior for familiar properties
\(P(p) = \int_{k} P(p \mid k) \mathop{}\!\mathrm{d}k\) as a stand-in for
a richer intuitive theory that could give rise to prevalence judgments.
We measured prior distributions empirically for the set of properties
(e.g., \emph{lays eggs, carries malaria}; 21 in total) used in our
generic sentences in Expt. 1a. To create a larger set of properties, we
reverse-code responses for five properties to create their corresponding
negative properties (e.g., we create a property \enquote{doesn't have
beautiful feathers} by subtracting from 100\% the responses for
\enquote{has beautiful feathers}).\footnote{This reverse-coding assumes
  for these properties that logical negation tokenizes its own threshold
  instead of being derived compositionally via bivalent logical
  negation. For related empirical and modeling investigations regarding
  resolving compositional vs.~non-compositional negation (in the context
  of gradable adjectives, e.g., \enquote{not happy} vs.
  \enquote{unhappy}), see Tessler and Franke (2018).}

\hypertarget{method-1}{%
\subsubsection{Method}\label{method-1}}

\hypertarget{participants-1}{%
\paragraph{Participants}\label{participants-1}}

We recruited 60 participants over Amazon MTurk. Three participants were
accidentally allowed to complete the experiment for a second time, so we
excluded their second responses (resulting in \(n=57\)). Two
participants self-reported a native language other than English;
removing their data (\(n=55\)) has no effect on the results reported.
The experiment took about 10 minutes and participants were compensated
\$1.00.

\hypertarget{procedure-and-materials-1}{%
\paragraph{Procedure and materials}\label{procedure-and-materials-1}}

\begin{figure}
\centering
\includegraphics{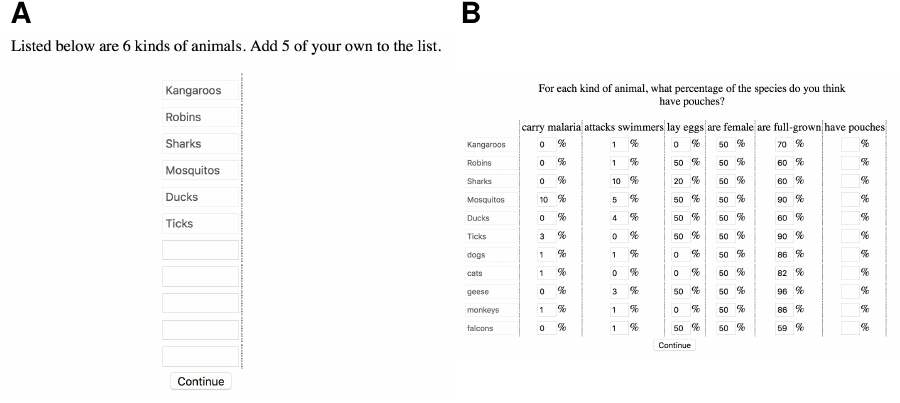}
\caption{\label{fig:generics-prior-task}Prior elicitation task. A:
Participants first generated animal names after seeing six example
categories. B: One feature at a time, participants estimated the
percentage of the category with the feature, for each category.}
\end{figure}

On each trial of the experiment, participants filled out a table where
each row was an animal category and each column was a property
(Figure~\ref{fig:generics-prior-task}). Participants first were shown
six animal categories randomly sampled from a set corresponding to
referent-categories of the generic sentences used in Expt. 1a (e.g.,
\textsc{robins, mosquitos}) and were asked to generate five animal kinds
of their own (Figure~\ref{fig:generics-prior-task}A). A column then
appeared to the right of the animal names with a property label in the
column header (e.g., \emph{lays eggs}). Participants were asked to fill
in each cell with the percentage of members of each of the species that
had the property (e.g., \enquote{50\%};
Figure~\ref{fig:generics-prior-task}B). Eight property--columns in total
appeared in the table. This whole procedure was repeated two times (two
trials). In total, each participant generated ten animal names and
reported on the prevalence of sixteen properties for twenty-two animals
(their own ten and the experimentally-supplied twelve).

\hypertarget{qualitative-results}{%
\subsubsection{Qualitative results}\label{qualitative-results}}

\begin{figure}[!h]
\includegraphics[width=\textwidth]{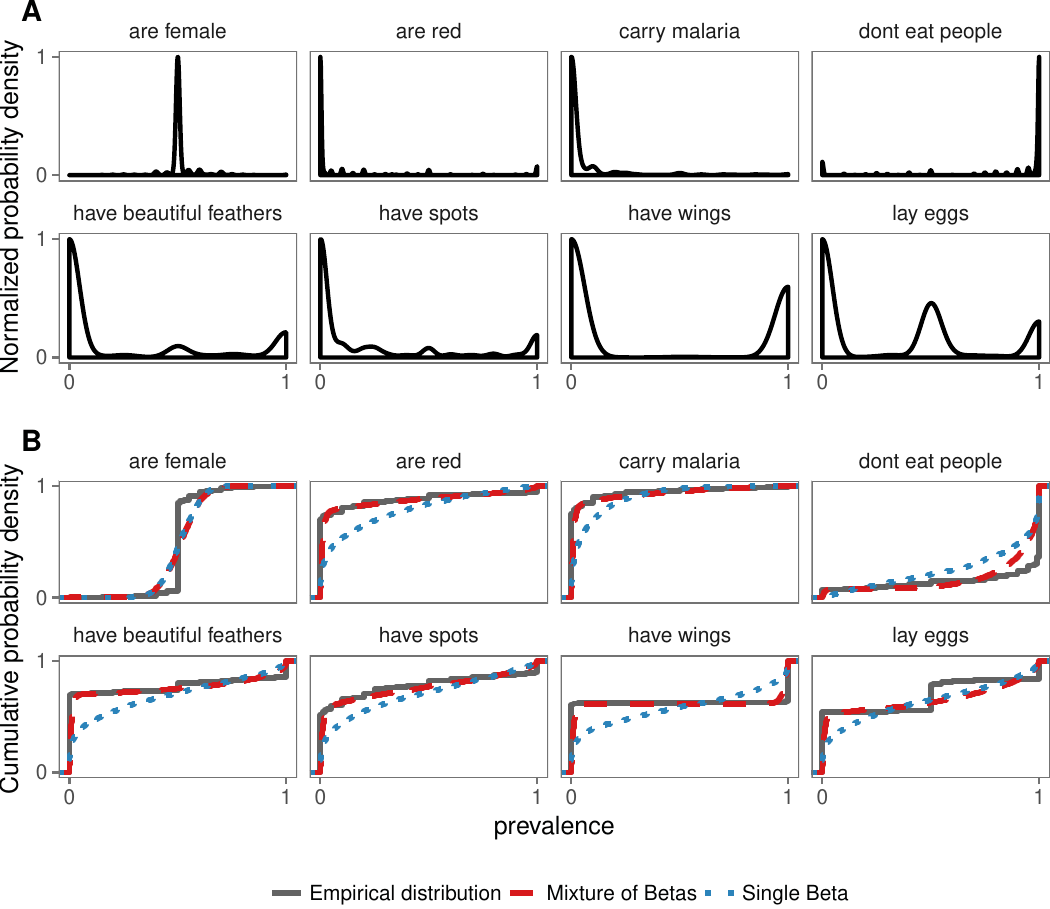} \caption{A: Empirically elicited prior distributions over prevalence for eight properties. B: Cumulative density plots reveal that a model of a mixture of two Beta distributions does substantially better at capturing the structure of the priors than a single Beta distribution. Distributions are the posterior predictive distributions for the models of the prior and the raw empirical distribution. A completely uniform distribution would be represented as the y = x line.}\label{fig:generic-endorsement-priors-figure}
\end{figure}

\begin{figure}[!h]
\includegraphics[width=\textwidth]{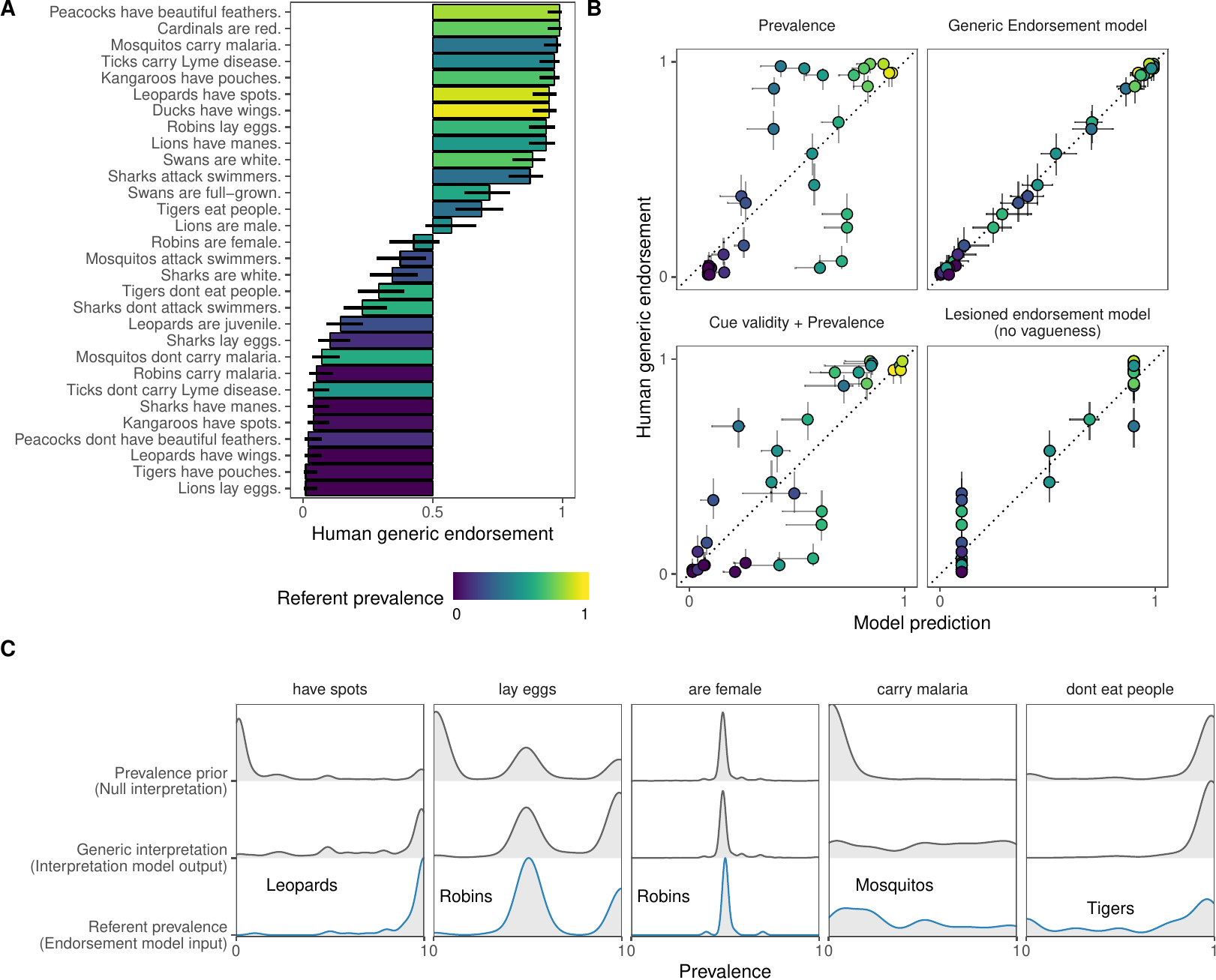} \caption{Endorsing generalizations about categories. A: Human elicited endorsements for thirty generic sentences reveal a continuum of endorsements. B: Model fits for the uncertain semantics speaker model (upper right), a fixed semantics speaker model (lower right), and regression models based on referent prevalence alone (upper left) and prevalence + cue validity (lower left). C: Five example empirical prevalence priors, model-predicted generic interpretations, and empirical referent prevalence (speaker belief) distributions..}\label{fig:generics-endorsement-figure}
\end{figure}

The elicited prior distributions have a diversity of shapes (8 examples
shown in Figure~\ref{fig:generic-endorsement-priors-figure}A) that are
qualitatively consistent with the schematic prior distributions used in
the \emph{Worked Examples} section (Figure~\ref{fig:simulations}A). The
property \emph{being female} is present in almost all categories in
almost exactly the same proportion, whereas priors for properties such
as \emph{laying eggs} or \emph{having spots} exhibit more structure
represented by the multimodality of these distributions. \emph{Being
red} exists mostly at extremely low prevalence levels (i.e., 0
prevalence) but also at high prevalence levels (e.g., 50\% or 100\%),
whereas \emph{carrying malaria} is really only present at low prevalence
levels. This diversity is relevant because our endorsement model makes
different predictions depending on the shape of these distrubutions.

\hypertarget{modeling-the-prevalence-priors}{%
\subsubsection{Modeling the prevalence
priors}\label{modeling-the-prevalence-priors}}

In order to incorporate the uncertainty in our measurement of the
prevalence prior into the endorsement model's predictions, we build a
Bayesian statistical model of the prior elicitation data. We approximate
the prevalence distribution for each property (e.g., \textsc{lay eggs})
with a Mixture of Betas model, which assumes that the data generated for
each kind comes from one of two underlying Beta distributions.\footnote{The
  Beta distribution is chosen because the support of this distribution
  is numerical values between 0 - 1, exactly the form of the response
  data in the prior elicitation task.} We specify one of these
distributions \emph{a priori} to represent kinds of animals who \emph{do
not have} a stable causal mechanism that could give rise to the property
(e.g., \textsc{lions} and \textsc{lay eggs}), which results in
prevalence or prevalence values close to or equal to 0. This \emph{null
distribution} is potentially present for all features and acts in
exactly the same way (i.e., the lack of producing the
feature).\footnote{This assumption is similar in spirit to that employed
  by \emph{Hurdle Models} of epidemiological data, where the observed
  count of zeros is often substantially greater than one would expect
  from standard models, such as the Poisson (e.g., when modeling adverse
  reactions to vaccines; Rose, Martin, Wannemuehler, \& Plikaytis, 2006)}
The second distribution represents kinds of animals who \emph{could}
have such a mechanism, and the two parameters of this distribution are
not specified \emph{a priori} and are not the same for all properties,
but are inferred on a property-wise basis from participants' responses.
The \emph{Mixture of Betas} distribution has a third free parameter (for
each property), the relative contribution of the null distribution (for
example: we expect the null distribution to not contribute at all to
properties like \emph{being female}, for which almost all categories
have at least some members with the property). To ensure the
\emph{Mixture of Betas} model of the prior is not overly complex, we fit
an additional model that represents only a single underlying
distribution (\emph{Single Beta}) for a comparison. For more details
about model implementation and inference, see Appendix C.

The prior distributions over prevalence are well modeled as a mixture of
two Beta distributions and not as a single Beta distribution (Figure~
\ref{fig:generic-endorsement-priors-figure}B; red vs.~blue lines). The
\emph{Single Beta} model provides a good fit for the \emph{being female}
distribution, but overly smooths the other distributions, washing out
the latent structure in participants' responses. One property that the
\emph{mixture of Betas} model does not perfectly capture is the prior
distribution over the feature \emph{lays eggs}. The empirical
distribution is tri-modal, with reliable modes at 0\%, 50\%, and 100\%;
a simple two-component mixture model has no way to account for such a
tri-modal distribution.\footnote{The third mode at 100\% is not
  attributable to categories for which all members could be female
  (e.g., chickens). Instead, it appears that some participants are
  responding that 100\% of several different kinds of birds (e.g.,
  robins) lay eggs. This may result from participants implicitly only
  considering female members of the category as relevant to answer a
  question about a reproductive capactiy like \emph{lays eggs}; this
  restriction of what enters into the prevalence computation is known as
  \emph{domain restriction}, is posited in several theories of generics
  (e.g., Cohen, 1999), and has been observed in other prevalence
  elicitation tasks (e.g., Prasada et al., 2013).} A more complex model
(i.e., one with three mixture components) would be necessary to
perfectly account for this item. Using a three-component model for this
distribution does not change the resulting model predictions and we
maintain the simpler two-component mixture for uniformity.

\hypertarget{endorsement-model-comparison}{%
\subsection{Endorsement model
comparison}\label{endorsement-model-comparison}}

We can now compute the predictions of our endorsement model. We describe
first the behavior of a set of alternatives models that have been
previously proposed in the literature and an alternative form of our
proposed endorsement model before proceeding to the results of the
uncertain threshold endorsement model.

\hypertarget{baseline-models}{%
\subsubsection{Baseline models}\label{baseline-models}}

We present two baseline, quantitative models that have previously been
used in the empirical literature on generic language. In addition to
serving as alternatives, these regressions model help us understand the
statistical properties of our experimental items. First, we estimate how
well referent prevalence itself predicts generic endorsement (e.g., does
the fraction of robins that lay eggs predict the felicity of
\enquote{Robins lay eggs}?). Second, we include \emph{cue
validity}---the probability of a kind given the feature---as a second
predictor in a linear model. We fit these models using standard
maximum-likelihood techniques and model uncertainty in the input
measurements (i.e., referent prevalence, cue validity) by bootstrapping
those data.

\hypertarget{referent-prevalence}{%
\paragraph{Referent prevalence}\label{referent-prevalence}}

From the prevalence prior data (Expt. 1b), we estimate participants'
beliefs about the referent prevalence (e.g., the percentage of
\textsc{robins} that \textsc{lay eggs}) and use it to predict
endorsement. We find a little over half of the variance in the
endorsement data is explained this way (\(r^2(30) = 0.51\);
MSE=\(0.08\); Figure ~\ref{fig:generic-endorsement-priors-figure}B
upper-left facet). Referent prevalence alone predicts a fair amount of
variance because our stimulus set includes generics that are true with
high prevalence properties (e.g., \enquote{Leopards have spots.}) and
false with low prevalence properties (e.g., \enquote{Leopards have
wings.}).

Large deviations from an account based purely on referent prevalence
remain: Generics in which the referent-category has intermediate
prevalence (prevalence quartiles 2 and 3:
\(16\% < \text{prevalence} < 64\%\)), are not at all explained by
referent prevalence (\(r_{Q2,3}^2(15) = 0.01\); MSE = \(0.14\)). This
includes generics that are judged true with relatively low referent
prevalence (e.g., \enquote{Mosquitos carry malaria}) and false with
relatively high referent prevalence (e.g., \enquote{Sharks don't eat
people}).

\hypertarget{cue-validity-and-referent-prevalence}{%
\paragraph{Cue validity and referent
prevalence}\label{cue-validity-and-referent-prevalence}}

Cue validity indexes the diagnosticity of the feature for a kind, given
formally by the probability of a kind given the feature \(P(k \mid f)\).
As discussed in the \emph{Relationship to Other Semantic Theories}
section above, cue validity is linearly related to expected value of the
prevalence prior distribution (see Appendix A for derivation). Cue
validity thus acts using a \emph{point estimate} of the prevalence prior
distribution, a single metric that summarizes the prior distribution.

Though the cue validity of a property for a category can be derived from
the prevalence prior distribution, previous empirical studies of
generics have estimated cue validity from different empirical sources
(Cree, McNorgan, \& McRae, 2006; Khemlani et al., 2012). In the
empirical literature on generics, researchers often ask directly about
the cue validity probability (e.g., ``There is an animal that lays eggs.
What is the probability that it is a robin?''; Khemlani et al., 2012),
though in the broader literature on semantic memory a \emph{free
production} paradigm is often employed (e.g., ``X lays eggs. What do you
think X is?''; Cree et al., 2006). We found that these two ways of
estimating cue validity diverge for a number of key cases (Appendix B).
Most notably, undiagnostic features (e.g., \emph{is female}), which in
theory have a cue validity close to zero, were rated as having
intermediate cue validity (around 0.5) in the \emph{direct question}
paradigm, but not in the \emph{free production} paradigm. In response to
a direct question such as \enquote{There is an animal that is female.
What is probability that it is a robin?}, participants seem to want to
respond \enquote{I don't know} by placing the slider bar at the
midpoint, rather than reporting their intuitive base rate that a random
animal would be a robin. The free production paradigm did not produce
artifacts such as this one, and we chose to use it as the more veridical
estimate of cue validity. For a detailed analysis of the different cue
validity measurements and comparison to cue validity derived from the
prevalence prior, see Appendix B.

A linear model that uses predictors for both referent-prevalence and cue
validity does a better job at explaining the endorsement data than just
prevalence alone (\(r^2(30) = 0.73\); MSE=\(0.04\)). This model is able
to account for the endorsements of examples like \enquote{Mosquitos
carry malaria} (model endorsement and bootstrapped-95\% confidence
interval = 0.85 {[}0.72, 0.86{]}) and \enquote{Lions have manes} (0.79
{[}0.61, 0.84{]})), as these features are very diagnostic of the kind
(generic endorsement both \(> 0.9\)). Deviations, however, still remain.
For example, \enquote{Robins lay eggs} still receives only intermediate
endorsement by this model (0.68 {[}0.56, 0.69{]}; human endorsement =
0.94 {[}0.87, 0.97{]}), and \enquote{Mosquitos don't carry malaria} is
misjudged to be a pretty good statement (0.58 {[}0.41, 0.58{]}; human
endorsement = 0.07 {[}0.04, 0.14{]}).

\enquote{Robins lay eggs} and \enquote{Mosquitos don't carry malaria}
highlight a shortcoming of reducing structured prevalence prior
distributions to single point-estimates of cue validity. \emph{Lays
eggs} is a somewhat diagnostic feature for birds, but there are many
kinds of birds, and the feature is not itself diagnostic for a
particular kind of bird like robins. Thus, the cue validity is low even
though robins are in the distinctive part of the \emph{lays eggs}
prevalence prior distribution (Figure~\ref{fig:simulations}A bottom).
Furthermore, cue validity cannot distinguish \emph{undiagnostic}
features (features present in almost every category; e.g., \emph{not
carrying malaria}) from \emph{false} features (features that are absent
a particular category; e.g., \emph{lions} and \emph{lay eggs}; see
Appendix B for more discussion of this distinction); the cue validity
can be near-zero for different reasons. Such a model makes the wrong
prediction for non-distinctive properties with high referent prevalence
(e.g., \enquote{Mosquitos don't carry malaria}).

\hypertarget{communicative-endorsement-models}{%
\subsubsection{Communicative endorsement
models}\label{communicative-endorsement-models}}

Our underspecified-threshold model considers how well the generalization
would bring a naive interpreter's prior distribution on prevalence
\(P(p)\) (Eq. \ref{eq:L0}; e.g., the prevalence of laying eggs among
other animals) in line with the speaker's belief about the referent
prevalence (\(p\) in Eq. \ref{eq:S1}; e.g, the prevalence of laying eggs
among robins). As described in the \emph{Baseline Models for
Quantitative Comparisons}, there are several substantive components to
this hypothesis: (a) property knowledge in the form of a prior
distribution over prevalence \(P(p)\), (b) endorsement as a
decision-theoretic process of uttering the generalization vs.~not
uttering it, and (c) vagueness in the semantics of a generalization. We
construct an alternative, lesioned model by removing the vagueness in
meaning, assigning a fixed semantics to the generalization (i.e.,
analogous to the quantified statements) but keeping the prior and the
decision-theoretic architecture in place. There are no correspondingly
simple ways to lesion the other two components (context or speaker
decision) while still producing a model that makes quantitative,
context-sensitive predictions.

For both the fixed-threshold and full uncertain-threshold endorsement
models, we build joint Bayesian data analysis models of the referent
prevalence \(p\), prevalence priors \(P(p)\) (both from Expt. 1b data),
and the endorsement data (Expt. 1a). Predicting the data from both Expt.
1a and 1b by a single, joint-inference model makes our assumptions
explicit about how these data were generated and is the proper way to
represent the uncertainty in our measurement of the prior elicitation
data (see Appendix C and Figure~\ref{fig:genericsModelDiagram} for
further model specification details). Empirically elicited referent
prevalence and prevalence prior data (Expt. 1b) directly constrain the
parameters that generate those quantities in the model (\(p\) in Eq.
\ref{eq:S1} and \(P(p)\) in Eq. \ref{eq:L0}, respectively). The
prevalence prior \(P(p)\) is modeled as a mixture of Betas, and referent
prevalence is modeled by a single Beta distribution. The endorsement
data (Expt. 1a) is modeled by our endorsement model (Eq. \ref{eq:S1}),
which has one free parameter \(\lambda\). To learn about the credible
values of the parameters of the joint-inference model and resulting
model predictions, we ran an incrementalized version of MCMC (Ritchie,
Stuhlmüller, \& Goodman, 2016) for 3 chains of 150,000 iterations,
discarding the first 50,000 for burn-in.

\hypertarget{lesioned-model-no-vagueness}{%
\paragraph{Lesioned model (no
vagueness)}\label{lesioned-model-no-vagueness}}

For a strong alternative model, we lesion the uncertain threshold model
so that it has a fixed threshold \(\theta\). We test the strongest,
possible fixed-threshold model by searching for the best possible single
threshold that fits the data. However, a fixed-threshold model will have
to accommodate responses that are literally inconsistent with the
threshold (i.e., a participant endorsing a generic when the referent
prevalence is less than \(\theta\)); we thus outfit this model with an
additional extrinsic noise parameter, to allow for random guessing.
Thus, the fixed-threshold alternative model claims that participants
make an information-theoretic decision, taking into account the
interpreter's prior distribution on prevalence \(P(p)\), using a
fixed-threshold semantics, where deviations from a pure
information-theoretic decision are accounted for by noise. This
alternative model has 2 additional parameters to our uncertain threshold
model (the value of the fixed-threshold and the proportion of noise
responses).

To evaluate the fixed-threshold model, we examine model predictions as
well as the posterior distribution over latent parameters of the model
(referent prevalence, prevalence priors, the optimality,
fixed-threshold, and noise parameters) given the observed data. The
Maximum A-Posteriori value and 95\% highest probability density interval
for the inferred (fixed) threshold and noise parameters are 0.34
{[}0.25, 0.37{]} and 0.20 {[}0.18, 0.22{]}, respectively. The inferred
optimality parameter in Eq. \ref{eq:S1} is 0.34 {[}0.25, 0.37{]}.
Figure~\ref{fig:generics-endorsement-figure}B (bottom right subplot)
shows the fixed-threshold model's ability to predict the generic
endorsement data (\(r^2(30) = 0.9329\); MSE = \(0.010964\)). Though the
model is able to capture a lot of the variance, it only makes three
kinds of judgments: true, false, or neither, similar to purely semantic
accouts of generics. It treats \enquote{Tigers eat people} (0.90
{[}0.89, 0.91{]}) as good a statement as \enquote{Peacocks have
beautiful feathers} (0.90 {[}0.89, 0.91{]}), though participants give a
substantially weaker endorsement of the former (\enquote{Tigers eat
people} = 0.69 {[}0.59, 0.77{]}; \enquote{Peacocks have beautiful
feathers} = 0.99 {[}0.94, 1{]}). Similarly, \enquote{Lions lay eggs}
(0.10 {[}0.09, 0.11{]}) is judged to be just as bad as
\enquote{Mosquitos attack swimmers} (0.10 {[}0.09, 0.11{]}), though
participants rate the former as completely false (0.01 {[}0, 0.06{]})
while the latter is kind of true (0.38 {[}0.28, 0.48{]}). The
fixed-threshold alternative model is unable to make these fine-grained
distinctions because it uses the same semantic threshold in all
contexts.

\hypertarget{uncertain-threshold-model}{%
\paragraph{Uncertain threshold model}\label{uncertain-threshold-model}}

Our underspecified threshold model is the same as the fixed-threshold
model, except that rather than having a fixed \(\theta\) for all
contexts, the model infers property-specific \(\theta\)'s. We use the
same Bayesian data analysis approach, dropping the additional parameters
required for the fixed-threshold model (the fixed-threshold and noise
parameters). Thus, this model has two fewer parameters than the
fixed-threshold model above.

We first examine the posterior predictive distribution on the prevalence
prior and referent prevalence data to ensure that the joint-inference
model does not distort these parameters at the service of predicting the
endorsement data (e.g., such a distortion could manifest by the
joint-inference model inferring that 100\% of mosquitos carry malaria in
order to predict that \enquote{Mosquitos carry malaria} is a good
utterance). This is an important step in model validation because it
tells whether the model's predictions are derived from intuitively
plausible values of the parameters (e.g., that not all mosquitos carry
malaria). Importantly, the joint-inference model captures the prior
elicitation data (e.g., the probability of carry malaria among various
species) and the referent prevalence data (e.g., the prevalence of
carrying malaria among mosquitos) as well as it did when we analyzed
these data in isolation (without conditioning on the endorsement data)
(\(r_{\text{prevalence prior parameters}}^2(60) = 0.9997\);
\(r_{\text{referent prevalence parameters}}^2(63) = 0.9989\); see
Appendix D, Figure 14). This result confirms that the
theoretically-interesting predictions of this model --- predictions of
generic endorsement --- are based on intuitively meaningful model
components (i.e., the shapes of the prevalence distributions). Finally,
the inferred optimality parameter in the endorsement model (Eq.
\ref{eq:S1}) is 2.47 {[}2.18, 2.75{]}, a range consistent with the
literature on similar models.

As we see in Figure~\ref{fig:generics-endorsement-figure}B (top right),
the uncertain threshold endorsement model explains nearly all of the
variance in human endorsements (\(r^2(30) = 0.9978\); MSE =
\(0.00035185\)). Examining the relevant model components that give rise
to these predictions further reveals the intuition for why the model
makes the predictions that it does (Figure
\ref{fig:generics-endorsement-figure}C). Recall the endorsement model's
alternative utterance is a null or silent utterance and thus, the
prevalence prior distributions are exactly what the interpreter model
would believe if the speaker does not produce the generic. The
endorsement model then decides whether maintaining the listener's prior
or updating it with the uncertain threshold semantics would better get
the listener to guess the correct prevalence for the category (correct
in the mind of the speaker). The model rates \enquote{Robins lay eggs}
as a good utterance because the prevalence posterior implied by the
generalization is similar to the referent prevalence.\footnote{We again
  note that the empirical prevalence prior for \emph{lays eggs} is
  tri-model with peaks at 0\%, 50\%, and 100\%. As a result, the generic
  interpretation is bi-modal: the listener is led to believe either 50\%
  or 100\% of robins lay eggs. Interestingly, the endorsement model
  still makes the correct prediction because the referent-prevalence for
  laying eggs among robins is also bi-modal (peaks at 50\% and 100\%).
  That is, domain-restriction (resulting in a response of 100\%)
  seemingly occurs for some participants for \emph{both} the prevalence
  prior and referent prevalence measurements, and this allows the model
  to predict correctly that \enquote{Robins lay eggs} is a good generic
  utterance.} On the other hand, the hypothetical interpretation for
\enquote{Robins are female} is almost indistinguishable from the
prevalence prior, because the prevalence prior has such low variance
(almost all animals have female members in exactly the same proportion);
the endorsement model then has no basis to prefer silence or the generic
statement and the model predicts that the utterance should be neither
good nor bad---endorsement around 0.5---also approximately the
proportion of participants who endorse the statement. \enquote{Mosquitos
carry malaria} is an interesting case because the prevalence prior has
high variance (i.e., participants are highly uncertain about the
prevalence of carrying malaria among categories). As a result, the
generic interpretation also has high variance; still, the generic
interpretation is more consistent with the referent prevalence than the
prevalence prior, and the endorsement model predicts \enquote{Mosquitos
carry malaria} is a good utterance. Finally, an utterance with high
referent prevalence, such as \enquote{Tigers don't eat people}, is
predicted to have low endorsement because the generic would be
misleading; even if most tigers don't eat people, saying \enquote{Tigers
don't eat people} implies that all don't eat people, which is too
strong.

\hypertarget{discussion}{%
\subsection{Discussion}\label{discussion}}

Generic language is the premier case study for generalizations in
language. Generics have been studied extensively in the cognitive and
developmental psychological literatures and have deep implications for
wide ranging phenomena from stereotype propagation (Rhodes et al., 2012)
to motivation (Cimpian et al., 2007). Heretofore, no models have been
articulated with enough precision to make quantitative predictions about
endorsement decisions, deciding whether a statement is true or false.
This empirical case study demonstrates that a semantics based on the
prevalence of the feature is tenable despite of alleged counterexamples
(e.g., \enquote{Robins lay eggs} vs. \enquote{Robins are female}). The
key theoretical insight is that the truth-functional semantics is
underspecified, or vague, and resolved in context by a process of
probabilistic inference. Our model provides a clear delineation of world
knowledge (formalized as a prevalence prior) from the semantics of
generics. We return to this point, and its implications for
theory-building, in the general discussion.

In explaining the variable endorsements of generics, we related the
referent prevalence (e.g., the percentage of robins that lay eggs) and
the prevalence prior (e.g., the prevalence of laying eggs for different
kinds of animals) to the endorsement of the generalization (e.g.,
\enquote{Robins lay eggs}) via an information-theoretic communicative
model where the meaning of a generic is simple but underspecified. In
this case study, we used generic statements about familiar animal
categories, which has long been the cleanest domain for testing semantic
theories of generics by providing minimal comparison like
\enquote{Robins lay eggs} vs. \enquote{Robins are female}. However,
modeling familiar category generics is a correlational analysis: The
relevant quantities in the model were measured rather than manipulated.
We now seek to demonstrate how these quantities are causally related to
endorsement, by manipulating referent prevalence (Case Study 2) and
prevalence priors (Case Study 3). In addition, we take this opportunity
to highlight the generality of the theory, by performing these
additional empirical tests in different domains for generalization:
events and causes.

\hypertarget{case-study-2-habitual-language}{%
\section{Case Study 2: Habitual
Language}\label{case-study-2-habitual-language}}

As with instances of categories, particular events like \enquote{Mary
smoked yesterday} can be generalized into \emph{habitual sentences}:
\enquote{Mary smokes}. It is believed that an analysis of generics
should lend itself naturally to be extended to an analysis of habituals
(e.g., Carlson, 2005; Leslie, 2008), but no such analysis or empirical
data has directly connected the two. In our second case study, we focus
on habituals about people's behaviors that take the form: \emph{singular
noun phrase} \(+\) \emph{present tense simple verb phrase} (e.g.,
\enquote{Mary smokes cigarettes}). Learning about the behaviors of
others is useful because they tell us about what that person is like
more generally (e.g., Repacholi \& Gopnik, 1997; Seiver, Gopnik, \&
Goodman, 2013). When children describe their lives to others, a
surprisingly large amount of the language produced concerns the actions
of people close to them (e.g., ``My brother works part-time at the
restaurant''; McGuire \& McGuire, 1986).

To test the generality of our theory, we use the same computational
model and follow the same general experimental structure as in the first
case study. We take the event analogue of prevalence to be the
\emph{rate} with which the event occurs (e.g., how often Mary
smokes).\footnote{Specifically, for the generalization \enquote{Mary
  smokes}, the instances being generalized are \emph{instances of Mary}.}
We test the model by first measuring the prevalence (rate) prior
distribution for various actions (e.g., how often different people smoke
cigarettes; Expt. 2A). We then measure endorsements of habitual
statements while manipulating the referent prevalence (Expt. 2B), and
use our computational model to predict habitual endorsements. By
describing novel characters to participants, we are able to directly the
manipulate the referent prevalence, which we were unable to do for
familiar categories in Case Study 1.

Finally, if habituals (and generics) are truly language for conveying
generalizations, they should reflect speakers' expectations, not merely
their observations. This intuition is sometimes expressed as an
intensional meaning component of generics (Dahl, 1975). For example,
imagine a very small town where by total coincidence, all residents chew
sugarless gum. Endorsing the sentence \enquote{Residents of this town
chew sugarless gum} seems to commit the speaker to believing that it is
not sheer happenstance, but that there is some underlying cause that
supports the counterfactual implication that were a new person to become
a resident of the town, they too would likely chew sugarless gum.

Our computational model predicts endorsement rates given a referent
prevalence \(p\). Using this model, we can ask quantitatively how well
prevalence conveyed by a speaker represents the actual, objective
frequency in the world (e.g., the rate at which a person has smoked
cigarettes in the past) or a subjective, predictive belief in the head
(e.g., the rate at which a person is expected to smoke in the future)?
Such a distinction would support the intuitions about sugarless gum
residents and could explain why \enquote{Supreme Court Justices have
even social security numbers} sounds strange even if nine out of the
nine current justices have even social security numbers (Cohen, 1999):
Our predictions about the evenness of the \emph{next} justice's social
security number are driven by strong prior beliefs that selection for
the Supreme Court is uncorrelated from the numerical properties of one's
social security number; the current observations are not enough in this
case to sway those beliefs. We examine this aspect of the theory by
measuring endorsements of habituals when causal forces intervene on the
world (e.g., the person buys a pack of cigarettes; Expt. 2C) as well as
participants' predictions about the likely frequency of the event in the
future. We then compare habitual endorsement models based on speakers
aiming to convey the objective, past frequency or their subjective,
future expectation.

\hypertarget{experiment-2a-measuring-the-prevalence-prior-for-events}{%
\subsection{Experiment 2A: Measuring the prevalence prior for
events}\label{experiment-2a-measuring-the-prevalence-prior-for-events}}

In order to generate model predictions for habitual endorsements, we
first elicit the prior distributions over rates for different events.
For language about the behaviors of people, \(P(p)\) represents a
language user's background knowledge about the rates with which people
perform a behavior; this prior can be constructed as a distribution over
\emph{different people}, each of whom do the behavior with a different
rate. We designed our elicitation task to take advantage of the
mixture-model representation of the prevalence prior used in Case Study
1. In particular, we assume, to a first approximation, that the
distribution over prevalence can be represented as a mixture of those
who tend to perform the action with a stable rate and those who do not
perform the action. With the further assumption that, all else being
equal, past is predictive of future behavior, we operationalize these
two kinds of people as \emph{people who have done the action before} and
\emph{people who have not the action before}. We design this experiment
to measure participants' beliefs about the relative proportion of these
two kinds of people (as a measure of the mixture parameter in the
prevalence prior model) as well as the rate at which people (who have
done the action before) do the action. We will assume for simplicity
that people who have never done the action before will probably never do
the action.\footnote{It is likely that more than just these two
  possibilities are represented in people's intuitive theories,
  corresponding to individuals with additional traits or demographics,
  and influenced by the types of people a speaker knows and interacts
  with. We assume here a simple two-component structure so as to not
  make the specification of the prior overly complex.}

\hypertarget{method-2}{%
\subsubsection{Method}\label{method-2}}

\hypertarget{participants-2}{%
\paragraph{Participants}\label{participants-2}}

We recruited 40 participants from Amazon's Mechanical Turk. Participants
were restricted to those with U.S. IP addresses and who had at least a
95\% work approval rating. The experiment took on average 12 minutes and
participants were compensated \$1.25 for their work.

\hypertarget{materials}{%
\paragraph{Materials}\label{materials}}

To construct our stimulus set, we choose actions from five categories of
typical human behaviors having to do with food and drug, work, clothing,
entertainment, and hobbies. For each category, we created pairs or
triplets of events that shared a superordinate action (e.g.
\emph{writing} poems vs.~novels). The events were chosen to intuitively
cover a range of likely frequencies. In total, thirty-one events were
used. For a full list of the stimuli used in Expts. 2A-C, see Appendix
D.

\hypertarget{procedure}{%
\paragraph{Procedure}\label{procedure}}

For each event, participants were asked two questions, with different
dependent measures. These questions were designed to measure the two
components of the prevalence prior distribution. We anticipated there to
be different beliefs about the rates and relative proportions of men
vs.~women, so we asked about both genders separately. The two questions
were schematically:

\begin{enumerate}
\item ``How many \{men, women\} have \emph{done action} before?'' \\

Participants responded ``N out of every J.'' by entering a number for N and choosing J from a drop-down menu (options: \{1000 - 10 million\}, incremented by 10x; default setting: 1000).

\item ``For a typical \{man, woman\} who has \emph{done action}  before, how frequently does he or she \emph{do action}?''\\  

Participants responded ``M times in K.'' by entering a number for M and choosing K from a drop-down menu (options: \{week, month, year, 5 years\}; default setting: year).
\end{enumerate}

For example, one set of prompts read: \enquote{How many women have
smoked cigarettes before?}; \enquote{For a typical woman who has smoked
cigarettes before, how frequently does she smoke cigarettes?}
Participants answered both questions for both genders on each slide (4
questions total per slide, order of male / female randomized
between-subjects), and every participant completed all 31 items in a
randomized order. The difference in meaning of these questions was
explained to participants on an instructions page before the
experimental trials and tested for recall on a subsequent trial.
Participants responded to this attention check by selecting an option
from a drop-down menu consisting of four options (one correct
description of the questions and three distractors).

\begin{figure}[!h]
\includegraphics[width=\textwidth]{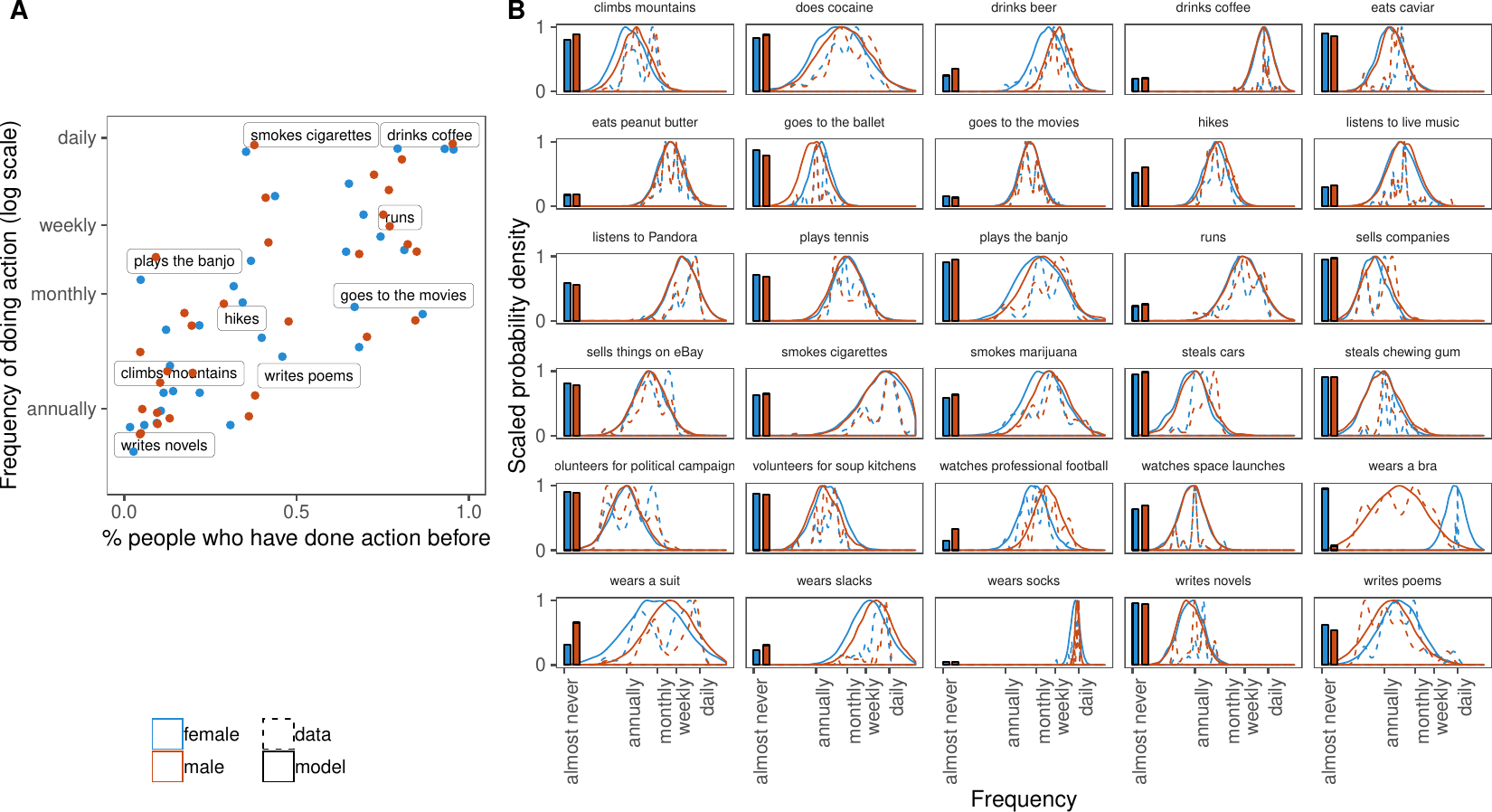} \caption{Prevalence priors for events (Expt. 2a). A: Maximum A-Posteriori (MAP) estimates of parameters of prevalence priors for the 31 items in Expt. 2. Items cover much of the range of possible parameter values. B: Reconstructed prevalence priors. In order to show frequencies for events that are very rare across people (e.g., "writes novels"), extremely low frequencies (\emph{almost never}) are omitted. Instead, height of the bars on left denote the MAP values of the mixture component (in terms of \emph{proportion of people who have never done action before}), reflecting the (inverse) popularity of the event across people.}\label{fig:habituals-prior-figure}
\end{figure}

\hypertarget{data-analysis-and-results}{%
\subsubsection{Data analysis and
results}\label{data-analysis-and-results}}

All participants responded correctly to both questions in the attention
check trial, so all collected data were used in the analysis. Question 1
elicits the proportion of people who have done an action before. We
rescale this to be a number between 0 and 1, and model it as generated
from a Beta distribution: \(d_{1} \sim \text{Beta}(\gamma, \xi)\).
Question 2 elicits the rate with which a person (who has done the action
before) does the action. We model this as generated by a log-normal
distribution: \(\ln d_{2} \sim \text{Gaussian}(\mu, \sigma)\). Each item
was modeled independently for each gender. We learned about the credible
values of the parameters by running MCMC for 100,000 iterations,
discarding the first 50,000 for burn-in.

The priors elicited cover a range of possible parameter values as
intended (Figure~\ref{fig:habituals-prior-figure}A): We observe a
correlation in our items between the mean \% of Americans who have
\textsc{done action} before (Question 1) and the mean log-frequency of
action (Question 2) (\(r_{1,2}(62) = 0.73\)). Items in our data set that
tend to be more popular actions also tend to be more frequent actions
(e.g., \emph{wears socks}) and visa-versa (e.g., \emph{steals cars}),
though there are notable exceptions (e.g., \emph{plays the banjo} is not
popular but done frequently when done at all, as is
\emph{smokes cigarettes}; \emph{goes to the movies} is a popular
activity though not done particularly often). This diversity is relevant
because the speaker model (Eq. \ref{eq:S1}) will endorse habitual
sentences (e.g., \emph{Sam goes to the movies vs. the ballet.})
contingent on these details of the prior distribution.

To generate prevalence prior distributions, we built a Bayesian
mixture-model for this prior elicitation task, analogous to that used in
Case Study 1 (Expt. 1A). The only difference is that we estimate the
mixture component \(\phi\) directly from responses to Question 1. We
assume that those who have not done the action before will probably not
do the action in the future. With these assumptions, the prevalence
distribution is given by:

\begin{align}
\phi & \sim \text{Beta}(\gamma_{Q1}, \xi_{Q1}) \nonumber \\ 
\ln p & \sim \begin{cases}
        \text{Gaussian}(\mu_{Q2}, \sigma_{Q2}) &\mbox{if } \text{Bernoulli}(\phi) = \textsc{T} \label{eq:priorModel}  \\
                \text{Delta}(p=0.01) &\mbox{if } \text{Bernoulli}(\phi) = \textsc{F} \\
        \end{cases}
\end{align}

Figure~\ref{fig:habituals-prior-figure}B shows example reconstructed
priors. In addition to specifying the correct way to combine our two
prior-elicitation questions, using this inferred prior resolves two
technical difficulties. First, it smooths effects that are clearly
results of the response format. For example, a very common rating for
certain events is \emph{1 time per year}. Presumably participants would
be just as happy reporting \emph{approximately} 1 time per year (e.g.,
on average, 1.2 times per year); the raw data does not reflect this due
to demands of the dependent measure. Second, this methodology better
captures the tails of the prior distribution (i.e., very frequent or
very infrequent rates) which have relatively little data and need to be
regularized by the analysis. Now that we have modeled the prevalence
prior data, we see whether our endorsement model can accurately predict
endorsement rates for habitual sentences about these actions.

\hypertarget{experiment-2b-habitual-endorsements}{%
\subsection{Experiment 2b: Habitual
endorsements}\label{experiment-2b-habitual-endorsements}}

In this experiment, we elicit human endorsements for generalizations
about events (\emph{habituals}; e.g., \enquote{Mary smokes cigarettes})
while manipulating the frequency with which the referent event occurs
(e.g., how often Mary smokes cigarettes).

\hypertarget{method-3}{%
\subsubsection{Method}\label{method-3}}

\hypertarget{participants-3}{%
\paragraph{Participants}\label{participants-3}}

We recruited 150 participants from MTurk. To arrive at this number, we
performed a Bayesian precision analysis to determine the minimum sample
size necessary to reliably ensure 95\% posterior credible intervals no
larger than 0.3 for a parameter whose true value is 0.5 and for which
the data is a 2-alternative forced choice. This analysis revealed a
minimum sample size of 50 per item; since participants only completed
about one third of the items, we recruited 150 participants. The
experiment took 4 minutes on average and participants were compensated
\$0.55 for their work.

\hypertarget{materials-1}{%
\paragraph{Materials}\label{materials-1}}

Each event from Expt. 2A was paired with between two to four
frequencies, for which the habitual statement would be evaluated.
Frequencies were presented in terms of a character performing the action
\enquote{three times in the past \emph{time interval}}. We chose to
always have the character perform the action three times to provide a
strong test of a baseline hypothesis that the habitual encodes the
person has done the action several (at least 3) times in the past.

Different time intervals were chosen for each event in order to maximize
the variability of responses within each item. Specifically, we used the
endorsement model to generate predictions based on the prior elicitation
data (Expt. 2A) for each item, and chose between two and four time
intervals across which maximal variability was predicted. For example,
relatively high frequencies were chosen (e.g., time intervals of weeks
and months) for items expected to occur rather often (e.g.,
\emph{runs}); for an item that was expected to occur infrequently (e.g.,
\emph{climbs mountains}), lower frequencies were chosen (e.g., time
intervals of years or longer) because the model predicted that much of
the variability in endorsement would occur in those respective ranges.
In total, 93 unique items were created by pairing frequencies with
events. The full list of items and frequencies can be found in Appendix
D.

\hypertarget{procedure-1}{%
\paragraph{Procedure}\label{procedure-1}}

On each trial, participants were presented with a \emph{past frequency
statement} for a given event of the form: \enquote{In the past \{week, 2
weeks, month, 5 years, \ldots{}\}, \emph{Person did X} 3 times}. For
example, \enquote{In the past month, Bill smoked cigarettes 3 times}.
Participants were asked whether they agreed or disagreed with the
corresponding habitual sentence: \enquote{Person does X}
(e.g.,\enquote{Bill smokes cigarettes.}). Participants completed
thirty-seven trials, which were composed of the thirty-one items from
the prior elicitation task randomly paired with either a male or female
character name. Six of these items were then also paired with a name of
the opposite gender (e.g., participants rated both a female character
and a male character who drank beer). These were used for an exploratory
analysis on differences in endorsements by gender of the target
character.

\hypertarget{results-1}{%
\subsubsection{Results}\label{results-1}}

The goal of this experiment was to elicit variability in habitual
endorsements. Consistent with this goal, we found habitual sentences
were endorsed for a wide range of frequencies. When actions are very
infrequent (3 times in a 5-year interval), habituals can receive strong
agreement (e.g., \emph{writes novels}, \emph{climbs mountains}). When
actions are relatively frequent (e.g., 3 times in a one month interval),
habitual sentences can receive less than full endorsement (e.g.,
\emph{wears socks}, \emph{drinks coffee}). In our data, actions
completed with a relatively high frequency (3 times in a one week
interval) receive at a minimum 75\% endorsement, though there is still
variability among them (e.g., between 10-25\% disagree that people who
wore a watch or wore a bra 3 times in the past week \emph{wear a watch}
or \emph{wear a bra} habitually). Finally, we observe that none of our
items receive less than 25\% endorsement (i.e., a maximum of about 75\%
of participants disagree with the habitual utterances), reflecting the
fact that these statements are not altogether \emph{false} even though
the action may be done very rarely.

\hypertarget{endorsement-model-comparison-1}{%
\subsubsection{Endorsement model
comparison}\label{endorsement-model-comparison-1}}

\begin{figure}[!h]
\includegraphics[width=\textwidth]{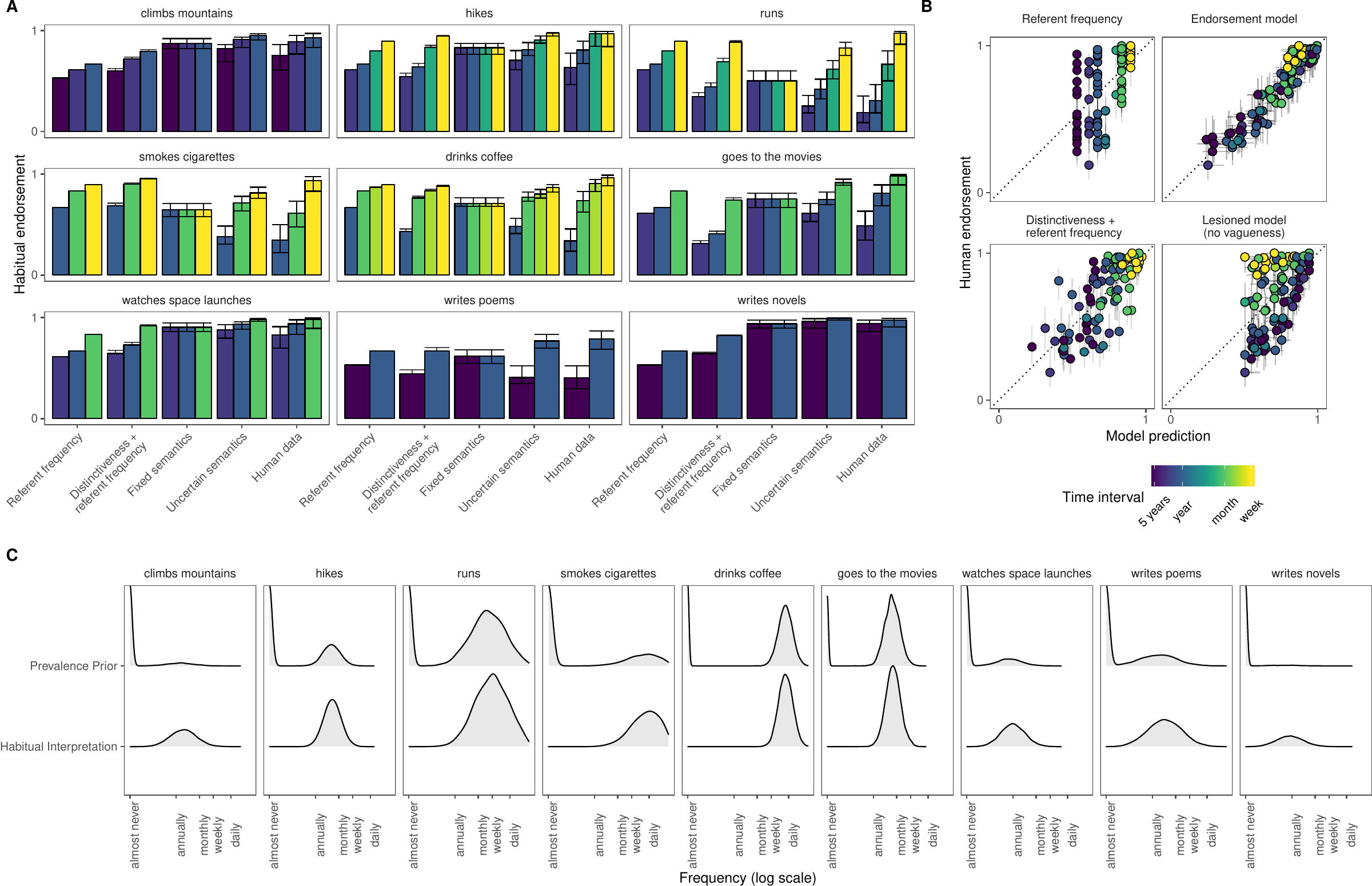} \caption{Endorsing generalizations about events. A: Endorsements for nine events given different frequencies of action. B: Top right: Model fits for all ninety-three habitual sentences by each model. C: Nine example frequency priors and posteriors upon hearing the habitual. These distributions are inferred using both data sources from Expts. 2a and 2b.}\label{fig:habituals-endorsement-figure}
\end{figure}

In an exploratory analysis, we found no differences between endorsements
of the habitual of characters with male and female names, and overall,
the mean endorsements by gender were strongly correlated
\(r(93) = 0.91\). Endorsements are even more highly correlated for the
six events we anticipated differences by referent-gender:
\(r(21) = 0.97\). This lack of a difference may be because the felicity
of habitual sentences depends on a comparison to individuals of both
genders (i.e, habituals are evaluated with respect to \emph{other
people}, not just other men or other women). Less interestingly, the
lack of a difference may be the result of gender being not very salient
in our paradigm, perhaps because the names used were not sufficiently
gendered.

We now turn to our model-based analyses to better understand the
endorsement data and the contribution of our model. For all analyses, we
collapse across gender of the referent character for endorsement
judgments. Parallel to our analysis of generic language endorsements, we
articulate a set of simple regression models and a fixed-threshold
alternative to our uncertain-threshold endorsement model. Analyses which
use the prevalence prior distribution \(P(p)\) (all models except
\enquote{referent frequency} regression) use a 50\% mixture of the
inferred priors for each gender to construct a single prevalence prior
distribution. Parallel to our analysis in Case Study 1, we model
uncertainty in the input measurements by bootstrapping the data for the
regression models and constructing joint-inference, Bayesian data
analytic models for the information-theoretic endorsement models
(fixed-threshold and uncertain-threshold; see \emph{Supplementary Model
Criticism} in Appendix C for a justification).

\hypertarget{referent-frequency}{%
\paragraph{Referent frequency}\label{referent-frequency}}

To understand the role of frequency in habitual endorsement, we use the
frequency supplied to the participants in our experiment as a predictor
in a linear model. This model predicts the same endorsement level for
two actions done with the same rate. Obvious counterexamples exist in
our data set: While participants are willing to endorse that a person
\enquote{\ldots{} climbs mountains} having done it 3 times in the past
\emph{year}, they are less willing to say that a person
\enquote{\ldots{} hikes} and not willing to say that a person
\enquote{\ldots{} runs}. Some actions done with a relative high rate
(e.g., 3 times in a month) do not receive full endorsement (e.g.,
\emph{smokes cigarettes}; Figure
\ref{fig:habituals-endorsement-figure}A). Overall, the
referent-frequency (in log-scale) predicts only a fraction of the
variability in responses (\(r^2(93) = 0.325\); MSE=\(0.0355\)). In
addition, for actions that are done on the time scale of years or longer
(lower median of frequency), referent frequency no longer explains
endorsements (\(r^2(50) = 0.0662\); MSE =\(0.0477\)). The prevalence
baseline does appreciably worse in this data set in comparison to the
generics data set (Case Study 1) because we were able to independently
manipulate the referent-frequency separate from the prevalence priors,
which we could not do for generics about familiar categories.

\hypertarget{distinctiveness-and-referent-frequency}{%
\paragraph{Distinctiveness and referent
frequency}\label{distinctiveness-and-referent-frequency}}

In our empirically elicited priors, items differ in the proportion of
people who have done the action before (the \emph{mixture parameter} of
the mixture model; Figure \ref{fig:habituals-prior-figure}A x-axis).
This mixture parameter is a major contributor to the mean of the
prevalence prior distribution, and thus relates to the \emph{cue
validity} of a particular feature for a particular individual (Appendix
A). Thus, we take this parameter as an index of the
\emph{distinctiveness} of the action, analogous to \emph{cue validity}
in the case of generic language. We construct a regression model that
treats endorsement as a linear combination of the frequency given to
participants and participants' responses to the question about the
mixture parameter \(\phi\) (i.e., the proportion of people who have done
the action before) as an index of distinctiveness.

This model is able to explain more of the variance in endorsements
(\(r^2(93) = 0.583\); MSE=\(0.022\)). It can differentiate events done
with the same frequency (e.g., \emph{writing poems} vs. \emph{novels}, 3
times in the past 5 years) by increasing endorsement of the more rare
action (\emph{novels}). Still, this model fails to capture fine-grained
differences in endorsement. For example, going to the movies is a
relatively nondistinctive action (many people do it) and going three
times in a year is not very frequent, and yet people still strongly
endorse the habitual (mean endorsement and 95\% CI:
\(0.81 [0.69, 0.89]\)), while this regression model predicts quite lower
judgments (\(0.41 [0.41, 0.43]\)). On the other hand, playing the banjo
three times in the past two years is not strong evidence for the
habitual, according to participants (\(0.45 [0.33, 0.59]\)).
Nevertheless, because playing the banjo is a distinctive action, the
regression model wants to endorse the habitual strongly in this case
(\(0.75 [0.75, 0.76]\)).

\hypertarget{lesioned-model-no-vagueness-1}{%
\paragraph{Lesioned model (no
vagueness)}\label{lesioned-model-no-vagueness-1}}

We next examine an information-theoretic endorsement model based on a
fixed-threshold semantics. This model is identical to the full
endorsement model, but is lesioned to not have vagueness, or uncertainty
about the meaning. A fixed-threshold model commits the habitual to
conveying literally that a person \emph{does the action with some
frequency}, and that threshold on frequency is the same for all actions.
As in Case Study 1, we incorporate this model into a Bayesian
joint-inference model to infer the fixed-threshold and simultaneously
predict both the priors data and the endorsement data (for more details
on model implementation, see Appendix C). We assume the
referent-prevalence \(p\) being conveyed by the endorsement model (Eq.
\ref{eq:S1}) is the frequency provided to participants (e.g., 3 times in
the past year). Additionally, to account for statements that would be
literally false under this model (frequencies that fall below the fixed
threshold), we include an additional noise parameter, as we did for the
fixed-threshold model in Case Study 1. To learn about the credible
values of the model's parameters and generate predictions given those
inferred parameter values, we collected 2 MCMC chains of 100,000
iterations, discarding the first 50,000 iterations for burn in.

The data analytic model infers that a low threshold is likely: the
Maximum A-Posteriori threshold and 95\% credible interval in units of
number of times per year is 0.01 {[}0.01, 0.37{]}. Compare this with the
lowest referent-frequency used in our data set: \(0.6\) times per year
(3 times every 5 years). Thus, all of the utterances evaluated under
this fixed-threshold model were literally true. As a result, the lowest
endorsement this model can apply to an utterance is 0.5 (since both the
habitual and silence are always true). As a result, the fixed-threshold
model exhibits a small dynamic range of endorsements, similar to the
referent-frequency model
(Figure~\ref{fig:habituals-endorsement-figure}B).

The fixed-threshold habitual updates the interpreter model's prior
beliefs differentially depending on the item. For instance, because of
the distinctiveness of \emph{climbs mountains}, the fixed-threshold
endorsement model fully endorses the habitual (\enquote{Mary climbs
mountains}) even at a low frequency (Figure
\ref{fig:habituals-endorsement-figure}A). However, the model cannot
differentiate among different frequencies of doing the same action,
because they are all above the truth-functional threshold. It is equally
true that a person does an action with non-zero frequency for any
frequency greater than zero, analogous to how \enquote{Some dogs are
friendly} is equally true whether 20\% or 50\% or 80\% of dogs are
friendly.

Overall, the fixed-threshold model is able to predict only a fraction of
the variance in human endorsements (\(r^2(93) = 0.299\);
MSE=\(0.0369\)). The model does this by inferring that 10\% {[}1, 19{]}
of the data is noise and that the speaker optimality parameter is 0.77
{[}0.70, 1.25{]}.

\hypertarget{uncertain-threshold-model-1}{%
\paragraph{Uncertain threshold
model}\label{uncertain-threshold-model-1}}

We used the same data analytic approach for the uncertain threshold
endorsement model and performed the same Bayesian statistical inference
over the model to learn about its parameters and predictions. Again,
this model has two fewer parameters than the fixed-threshold model (no
fixed-threshold parameter and no extrinsic noise process). As shown in
Figure~\ref{fig:habituals-endorsement-figure}B, the uncertain-threshold
endorsement model does a good job of accounting for the variability in
responses (\(r^2(93) = 0.894\); MSE=\(0.00598\)), including actions done
on the time scale of years or more (\(r^2(50) = 0.903\);
MSE=\(0.00617\)).

Figure~\ref{fig:habituals-endorsement-figure}C provides insight into how
the uncertain-threshold model is able to match human judgments. The
endorsement model simulates how an interpreter would understand the
habitual sentence. A habitual is interpreted relative the prior
distribution over frequencies, and the comparison between the frequency
implied by the habitual vs.~staying silent results in different
frequencies at which the generalization is good to assert. Climbing
mountains three times in the past year is good evidence that you
\emph{climb mountains} because it is approximately the frequency that a
listener would infer given the utterance; going for a hike three times
in the past year is correspondingly less convincing that you
\emph{hike}; and if you went for a run three times in the past year, you
not a person who \emph{runs}. Only the uncertain-threshold model is able
to draw these subtle distinctions.

\hypertarget{discussion-1}{%
\subsubsection{Discussion}\label{discussion-1}}

Habitual language exhibits context-sensitivity directly parallel to that
of generic language (Case Study 1). Habituals are endorsed for a wide
range of frequencies, but show systematic patterns relative to the prior
distribution of frequencies, as formalized by the uncertain-threshold
model. Again, we articulated a number of alternative models and found
that only the underspecified threshold model was able to explain the
variability in endorsements.

In this case study, we manipulated rather than measured the referent
frequency (e.g., the frequency with which a person drinks coffee). By
manipulating the target frequency, we have shown that it is causally
related to habitual endorsements in the way predicted by our model (and
in a way that a fixed-threshold model cannot account for). The
relationship is not linear, however; habitual endorsements vary in
complex ways that reflect interpreters' prior knowledge about the event
in question.

In Expt. 2B, participants were given a statement about how often a
person has done the action in the past and asked to judge the
corresponding habitual statement. This design potentially confounds an
important distinction for the language of generalization: Does the
prevalence communicated by a generalization indicate an objective, past
frequency or a subjective, future expectation? In Expt. 2C, we
investigate this question by teasing apart \emph{past} from
\emph{predictive frequency} and measuring its influence on habitual
endorsement.

\hypertarget{experiment-2c-what-is-prevalence}{%
\subsection{Experiment 2c: What is
prevalence?}\label{experiment-2c-what-is-prevalence}}

While past frequency is often a good indicator of future tendency, the
future is under no obligation to mimic the past. Does habitual language
communicate probabilities in terms of past frequency or future
expectations? On one hand, speakers can only be certain about what has
happened in the past. On the other hand, it is important for speakers to
be able to convey their predictions of what they believe will be the
case in the future.

People can change their behavior abruptly due to a variety outside
events (e.g., developing an allergy) or intend to do an action without
actually completing it (e.g., by making a resolution). We introduce
these causal events into our experimental paradigm to measure their
influence on endorsement. To provide the appropriate model-based
analysis, participants in one condition make a prediction about the
future (\emph{predictive frequency}). In another condition, participants
decide whether or not to endorse the habitual sentence (as in Expt. 2b).
We then compare two uncertain-threshold models: one which uses
participants' ratings of \emph{predictive frequency} as the referent
prevalence and one which uses the \emph{past frequency} (as was done for
Expt. 2B). In addition, we compare to a baseline linear model that uses
only the \emph{predictive frequency} (no priors) to model endorsement.

\begin{figure}[!h]
\includegraphics[width=\textwidth]{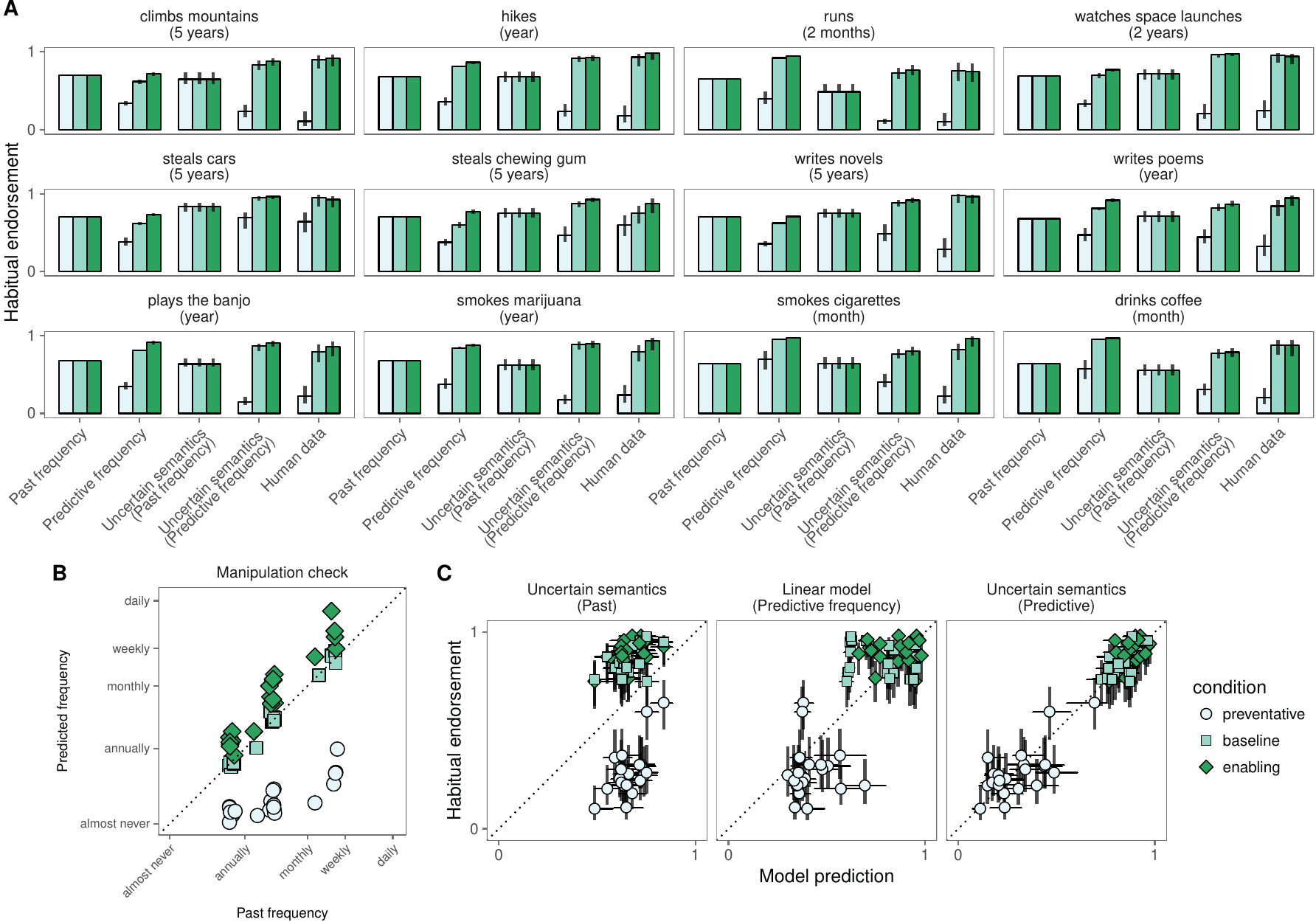} \caption{A: Example empirical and predicted endorsements for habituals in three conditions. B: Predicted frequency as a function of past frequency and condition manipulation (enabling, preventative, and baseline). C: Model fits for the uncertain threshold model using past frequency, linear model based on future frequency, and uncertain threshold model using future frequency.}\label{fig:habituals-predictive-figure}
\end{figure}

\hypertarget{method-4}{%
\subsubsection{Method}\label{method-4}}

\hypertarget{participants-4}{%
\paragraph{Participants}\label{participants-4}}

We recruited 270 participants from MTurk, using the same criteria as
Expt. 2b. 120 were assigned to the \emph{predictive frequency} condition
and 150 were assigned to the \emph{habitual endorsement} condition. The
experiment took on average 3.50 minutes (\emph{predictive frequency})
and 2 minutes (\emph{habitual endorsement}). Participants were
compensated \$0.40.

\hypertarget{materials-2}{%
\paragraph{Materials}\label{materials-2}}

The events used were a subset of those used in Expts. 2A \& B (21 of the
original 31). In addition, we crafted statements that were intended to
either increase the frequency (\emph{enabling}; e.g.,
\enquote{Yesterday, Bill bought a pack of cigarettes.}) or decrease the
frequency (\emph{preventative}; \enquote{Yesterday, Bill quit smoking.})
of the event in the future. In order to increase the potential
variability of responses across the experimental conditions,
participants only saw the frequencies that led to the most intermediate
endorsement of the habitual in Expt. 2B. We did not include separate
trials for both male and female names for the select items we did in
Expt. 2B, since we saw no differences in their endorsements of the
habitual. See Appendix D for a full list of the items and frequencies
used, as well as the enabling and preventative information.

\hypertarget{procedure-2}{%
\paragraph{Procedure}\label{procedure-2}}

The procedure was identical to Expt. 2B except for the inclusion of a
second sentence on a subset of trials (\emph{preventative} and
\emph{enabling} trials). On all trials, participants were presented with
a \emph{past frequency sentence} (same as Expt. 2B). Additionally,
trials either included \emph{preventative} information, \emph{enabling}
information, or no additional information (identical to Expt. 2B), in
equal proportions. See Table 2 for example trials.

\begin{table}[ht]
\centering
\begingroup\fontsize{9pt}{10pt}\selectfont
\begin{tabular}{ |p{1in}|p{1.75in}| p{1.75in}|p{1.75in} |}
  \hline
{\bfseries Habitual} & {\bfseries Baseline} & {\bfseries Preventative} & {\bfseries Enabling} \\ 
  \hline
John smokes cigarettes. & In the past month, John smoked cigarettes 3 times. & In the past month, John smoked cigarettes 3 times. Yesterday, John quit smoking cigarettes. & In the past month, John smoked cigarettes 3 times. Yesterday, John wanted a smoke and bought a pack of cigarettes. \\ 
   \hline
Tina volunteers at soup kitchens. & In the past five years, Tina volunteered for soup kitchens 3 times. & In the past five years, Tina volunteered for soup kitchens 3 times. Yesterday, Tina grew disillusioned with the soup kitchen system and wants nothing to do with it anymore. & In the past five years, Tina volunteered for soup kitchens 3 times. Yesterday, Tina researched a new soup kitchen in the area and is going to volunteer with them. \\ 
   \hline
\end{tabular}
\endgroup
\caption{Example stimuli used in Expt. 2c.} 
\end{table}

In the \emph{predictive frequency} condition, participants were asked
``In the next \emph{time interval}, how many times do you think
\emph{person does action}?'', where the \emph{time interval} was the
same as given in the past frequency statement. In the \emph{habitual
endorsement} condition, participants were asked if they agreed or
disagreed with the corresponding habitual sentence (as in Expt. 2B).

\hypertarget{predictive-frequency-results}{%
\subsubsection{Predictive frequency
results}\label{predictive-frequency-results}}

Figure \ref{fig:habituals-predictive-figure}B shows the mean predicted
future frequency as a function of the past frequency given to the
participant and the type of causal information given. We observe in the
baseline condition that future frequency perfectly tracks past frequency
(\(r(21) = 0.994\)). That is, participants believe if a person smoked
cigarettes 3 times last month, they will smoke cigarettes 3 times next
month. This result implies that our model makes identical predictions
for Expt. 2B whether the referent is past frequency or expected future
frequency (indicating, as expected, that we must look to new data to
distinguish these models). Critically, we observe the preventative
information strongly decreases and the enabling information slightly
increases predicted frequency (Figure
\ref{fig:habituals-predictive-figure}B, white and dark green dots).

We confirmed these observations using a linear mixed-effects model,
predicting the log-transformed responses from the log-transformed past
frequency and the experimental condition (baseline, preventative,
enabling). To account for participant and item variability in this
analysis, we also include random effects of intercept and condition for
both participants and items. Confirming that our manipulation worked as
intended, the preventative information led to significantly lower
predictions for future frequency, relative to the baseline condition
(\(\beta = -3.18\); \(SE = 0.27;\) \(t = -11.96\)). There was also a
tendency for the enabling information to lead to higher predictions for
future frequency, relative to baseline (\(\beta = 0.96\); \(SE = 0.11;\)
\(t = 8.54\)). Finally, past frequency was a significant predictor of
predicted future frequency (\(\beta = 1.01\); \(SE = 0.02;\)
\(t = 40.80\)).

\hypertarget{habitual-endorsement-results}{%
\subsubsection{Habitual endorsement
results}\label{habitual-endorsement-results}}

There is a clear and consistent negative effect of preventative
information on endorsements for the habitual sentences (Figure
\ref{fig:habituals-predictive-figure}A; white bars). When collapsing
across items, the Bayesian Maximum A-Posteriori estimate and 95\%
highest probability density interval for the true endorsement
probabilities per condition are: baseline = 0.85 {[}0.83, 0.87{]},
enabling = 0.90 {[}0.88, 0.92{]}, preventative = 0.29 {[}0.26, 0.32{]}.
Still, frequency --- even predictive frequency --- does not perfectly
explain the endorsements (\(r^2(63) = 0.518\); MSE = \(4.29\); Figure
\ref{fig:habituals-predictive-figure}B).

We use our formal model to test whether past or predictive frequency
matters for endorsement. To formalize the predictive frequency speaker
model, we use the mean predictive frequency as the referent-prevalence
\(p\) that the endorsement model (Eq. \ref{eq:S1}) aims to convey. The
past frequency model is constructed using the past frequency supplied to
participants as the referent-prevalence. We analyze this model in the
same Bayesian data analysis regime as for our previous models. We use
the same priors over the parameters as before and learn about the
posterior distribution by collecting three independent MCMC chains of
100,000 iterations (removing the first 50,000 for burn-in). Figure
\ref{fig:habituals-predictive-figure}C shows the resulting model
predictions for the past frequency and the predictive frequency
endorsement models. Participants' judgments of the habitual statements
was indeed influenced by the causal manipulations in the way predicted
by the endorsement model that uses the predictive frequency as the
referent prevalence (\(r^2(63) = 0.931\); MSE = \(0.00594\)). The model
based on past frequency does not make different predictions for the
different causal manipulation conditions and does a poor job at
explaining the endorsements (\(r^2(63) = 0.0334\); MSE = \(0.0833\)).
This result strongly suggests that prevalence represents a predictive
belief about the future.

\hypertarget{discussion-2}{%
\subsection{Discussion}\label{discussion-2}}

Habitual language conveys generalizations about events. Our model
decides if a habitual sentence is a pragmatically useful way to describe
the rate at which a person does an action, taking into account a naive
interpreter's prior beliefs about the event (measured in Expt. 2A). Our
computational model endorses statements that communicate generalizations
about events with the same sensitivity to context and frequency that
people exhibit (Expt. 2B \& C). In Expt. 2B, we varied the type of event
and the past frequency with which the person did the action, and found
graded endorsements of the corresponding habitual sentences. By
manipulating (rather than measuring) the referent frequency, we showed
how alternative models were unable to account for the gradience in
endorsement. In particular, we show that prior knowledge in an
information-theoretic, communicative model is not sufficient to produce
gradience in endorsement: The fixed-threshold model, which has these
components, does not make different predictions for different
frequencies. Only our uncertain threshold model was able to precisely
account for the wide range of endorsements.

In Expt. 2C, we further investigated the nature of the underlying
prevalence scale by introducing causal information that enabled and
prevented future occurrences of the action. We used the
empirically-measured predicted future frequency as the object of
communication for our endorsement model. We found that the endorsement
model that seeks to communicate its predictions (rather than its
observations) is a better model of habitual endorsements under these
situations. That is, habitual language (and generalization language more
generally) is fundementally about conveying people's predictive beliefs,
not what has actually happened.

In these experiments, we introduced participants to novel actors and, by
doing so, were able to directly manipulate the referent frequency. The
kinds of events we used were familiar to participants (e.g.,
\emph{running}) and thus we measured the prevalence priors for those
events. In our final case study, we experimentally manipulate the
prevalence priors, testing their causal influence over endorsements. In
addition, we further we extend our theory to the language of causal
relationships.

\hypertarget{case-study-3-causal-language}{%
\section{Case Study 3: Causal
Language}\label{case-study-3-causal-language}}

Language about causal relationships manifests in generalization. The
utterance \enquote{Fire causes smoke} relates to \enquote{This fire
caused this smoke} in a way analogous to how \enquote{John runs} relates
to \enquote{John ran yesterday}. We explore this hypothesis in our third
case study: causal language or \emph{causals} (e.g., \enquote{A causes
B}).

The problem of \emph{causal induction}---knowing that one thing causes
another---has been studied extensively in human psychology (Cheng, 1997;
Cheng \& Novick, 1992; Griffiths \& Tenenbaum, 2005, 2009). Classically,
this is cast as a problem of inducing an unobservable relation
(\emph{type causation}) from observable events or contingency data. We
take a different approach, examining \emph{type causation} by the
language used to describe it (e.g., \enquote{A causes B}). We explore
the idea that such language conveys a generalization about \emph{token}
or \emph{actual causation} (e.g., \enquote{A caused B, in this
instance}) and that our theory of the language of generalizations
extends in a natural way to describe \emph{causal language}. Ascribing
causation to an individual event (\emph{token or actual causation}) is
itself a complex, inferential process (e.g., depending on counterfactual
reasoning; Gerstenberg, Goodman, Lagnado, \& Tenenbaum, 2015), which we
do not try to model here.

In this paper, we posit that prevalence priors are a mediating
representation between abstract conceptual structure and the language of
generalization. In this last set of experiments, we explicitly test the
relationship between the prevalence priors and endorsements of
generalizations about causes by manipulating the priors.

\hypertarget{experiment-3a-manipulating-prevalence-priors}{%
\subsection{Experiment 3a: Manipulating prevalence
priors}\label{experiment-3a-manipulating-prevalence-priors}}

In this experiment, we manipulate participants' background knowledge,
measuring these beliefs in order to check whether the manipulation was
successful. Experiment 3B (\emph{causal endorsement}) will then use a
very similar experimental procedure in exploring the language of causal
generalizations.

\hypertarget{method-5}{%
\subsubsection{Method}\label{method-5}}

\hypertarget{participants-5}{%
\paragraph{Participants}\label{participants-5}}

We recruited 160 participants from Amazon's Mechanical Turk.
Participants were restricted to those with U.S. IP addresses and who had
at least a 95\% work approval rating. The experiment took on average
1.70 minutes and participants were compensated \$0.50 for their work.

\hypertarget{materials-3}{%
\paragraph{Materials}\label{materials-3}}

Participants were told a story of a scientific experiment testing
different substances to produce an effect (either to make animals sleepy
or make plants grow tall). Our cover stories were constructed so that
the potential cause could have some plausible intuitive mechanism that
could give rise to the property (e.g., a naturally occurring herb
causing animals to be sleepy). The two cover stories can be seen in
Table 3.

Participants were then shown \enquote{previous experimental results},
which followed one of four distributions represented as a table of
numbers. In two of the conditions, participants saw results that came
from a single underlying distribution (\emph{common} conditions). In one
of these conditions, all causes produced a strong effect (average
efficacy approximately 98\%; the \emph{common strong} condition). In the
second of these conditions, all causes produced a weak effect (average
efficacy approximately 20\%; the \emph{common weak} condition). The two
other conditions used distributions in which some experiments resulted
in either no or very few successes (i.e., produced 0, 1, or 2
successes), and others that either had strong or weak effects as above.
These are the \emph{rare strong} and \emph{rare weak} distributions.

\begin{table}[ht]
\centering
\begingroup\fontsize{9pt}{10pt}\selectfont
\begin{tabular}{ |p{1.75in}|p{2in}|p{2in} |}
  \hline
{\bfseries .} & {\bfseries Plants} & {\bfseries Animals} \\ 
  \hline
Cover story & On this planet, there is a plant called feps and your team wants to figure out how to make these plants grow tall. Your team runs experiments trying to make feps grow tall with different fertilizers. & On this planet, there are animals called cheebas and your team of scientists wants to figure out how to make these animals sleepy. Your team runs experiments trying to make cheebas sleepy with different naturally occurring herbs. \\ 
   \hline
Evidence statement & Your team gave fertilizer B to 100 different feps. Of those 100 treated, 2 feps grew tall. & Your team gave herb C to 100 different cheebas. Of those 100 treated, 98 cheebas were made sleepy. \\ 
   \hline
\end{tabular}
\endgroup
\caption{Cover stories and example evidence statements for the two sets of materials used in Expt. 3} 
\end{table}

\begin{figure}[!h]
\includegraphics[width=\textwidth]{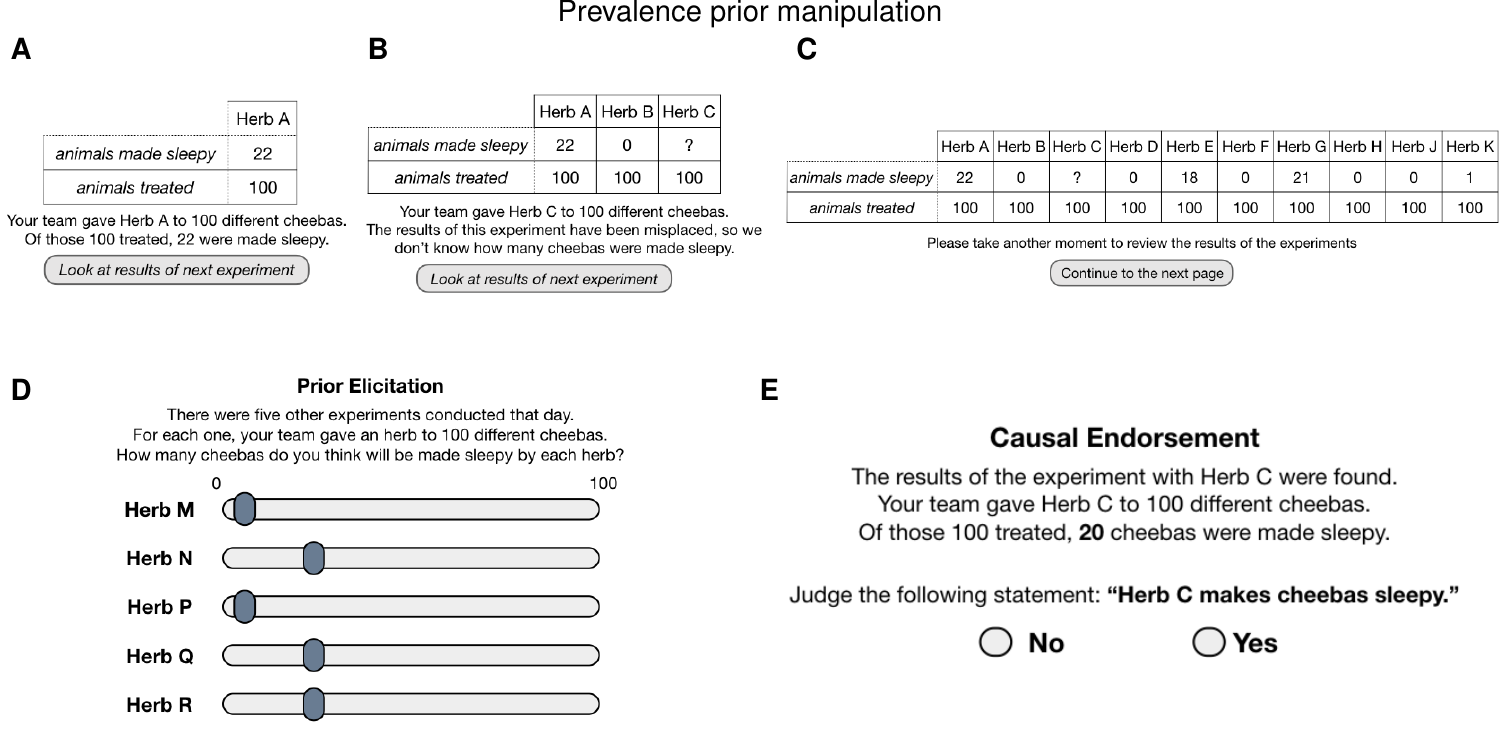} \caption{Overview of Experiment 3. A-C: Results of previous experiments are shown one at a time, described in text and displayed in a table. One of the results was lost. Participants are asked to review previous results once all displayed. D: Prior elicitation task: Participants predict the results of the next 5 experiments. E: Causal endorsement task: Results of previously lost experiment are found and participants are asked to evaluate the causal generalization.}\label{fig:causalExpt}
\end{figure}

\hypertarget{procedure-3}{%
\paragraph{Procedure}\label{procedure-3}}

The experiment was a single trial: Each participant saw only one cover
story with one distribution of previous experiments. Participants were
told that they were an astronaut-scientist on a distant planet trying to
figure out how some system works (i.e., how to make a certain kind of
animal sleepy with different herbs or how to make a plant grow tall with
different fertilizers). The story for the \emph{sleepy animals}
condition read:

\begin{quotation}
You are an astronaut-scientist exploring a distant planet. 
On this planet, there are animals called cheebas and your team of scientists wants to figure out how to make these animals sleepy.
Your team runs experiments trying to make cheebas sleepy with different naturally occurring herbs.
The results are shown below:
\end{quotation}

Participants then clicked a button to show the results of the
experiments, which appeared one at a time in a random order (following a
particular distribution). Experimental results were also are described
linguistically (e.g., \enquote{Your team gave herb A to 100 different
cheebas. Of those 100 treated, 98 cheebas were made sleepy.}) as well as
displayed in a table showing the number of successes (e.g.,
\enquote{animals made sleepy}) per number of attempts (always 100 per
experiment; Figure \ref{fig:causalExpt}A). We described the results of
these individual experiments using token-level causal language (e.g.,
\enquote{98 cheebas were made sleepy}) to imply that actual causation
occurred in these cases. Participants see the results of eleven
experiments, though they are told the results of one experiment were
lost and a \enquote{?} was placed in the table (Figure
\ref{fig:causalExpt}B). (These lost results would be found in the
\emph{causal endorsement} task, Expt. 3B). After participants viewed the
results of the 10 experiments (and 1 missing experiment), they are told
to review the results of the experiments before continuing (Figure
\ref{fig:causalExpt}C).

Upon clicking the continue button, the table of experiment results is
removed and participants are told that five more experiments were
conducted that day and asked to predict the results of those experiments
(Figure \ref{fig:causalExpt}D). Participants were given five slider bars
ranging from 0 - 100 to rate the number of predicted successes out of
100 attempts. After responding, participants then completed an attention
check survey where they were asked what the team of scientists was
investigating (choosing a response from a drop-down menu with 12
options) and to input one of the numerical results they saw on the
previous screen. This attention check served to confirm that
participants had encoded both relevant aspects of the experiment (the
domain and the frequencies).

\hypertarget{results-2}{%
\subsubsection{Results}\label{results-2}}

\begin{figure}[!h]
\includegraphics[width=\textwidth]{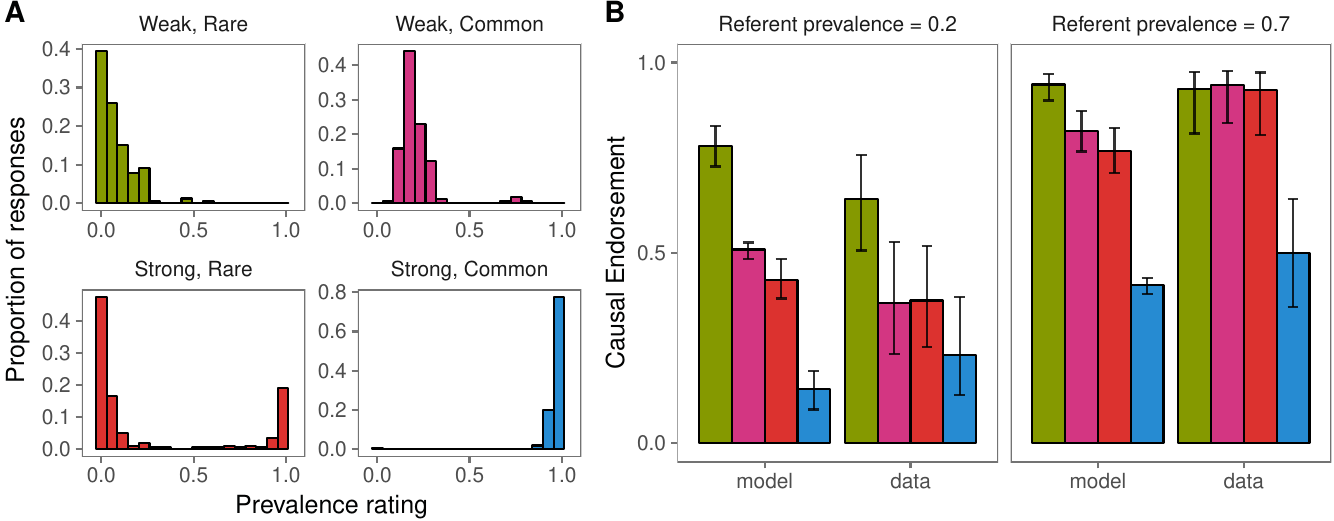} \caption{Endorsing generalizations about causes. A: Empirical prevalence prior distributions elicited following prior manipulation cover story (Expt. 3a). B: Endorsement model predictions and human elicited endorsements for four manipulated prevalence prior distributions (colors) and two referent prevalence levels (facets).}\label{fig:figure-causals}
\end{figure}

20 participants were excluded from the analysis for failing to answer
both of the attention check questions correctly, leaving a total of 140
responses for analysis. The empirically elicited distributions of
responses were not appreciably different for our two cover stories
(herbs making animals sleepy, fertilizer making plants grow tall) and
thus we collapse the data across these two stories. The distributions
that resulted from participants predicting the causal efficacy of the
new substances are shown in Figure~\ref{fig:figure-causals}A. As is
visually apparent, the empirical prevalence distributions differ between
conditions and nicely recapitulate the distributions supplied in the
different experimental conditions, suggesting that the manipulation does
indeed change participants' representations of what probabilities are
likely to occur in each experimental condition. This diversity is
important because the model of generalizations predicts differences in
endorsement---for the same referent prevalence---depending on these
priors.

\hypertarget{experiment-3b-causal-endorsements}{%
\subsection{Experiment 3b: Causal
Endorsements}\label{experiment-3b-causal-endorsements}}

In this experiment, we tested whether the manipulated priors of Expt. 3A
are causally related to the endorsement of causal statements. Most of
the experimental design was identical to that of Expt. 3A.

\hypertarget{method-6}{%
\subsubsection{Method}\label{method-6}}

\hypertarget{participants-6}{%
\paragraph{Participants}\label{participants-6}}

We recruited 400 participants from Amazon's Mechanical Turk.
Participants were restricted to those with U.S. IP addresses and who had
at least a 95\% work approval rating. None of the participants had
participated in Experiment 3a. The experiment consisted of one trial and
took on average 1.40 minutes; participants were compensated \$0.25 for
their work.

\hypertarget{procedure-and-materials-2}{%
\paragraph{Procedure and materials}\label{procedure-and-materials-2}}

The materials were the same as in Expt. 3A.

The first part of the experimental trial was the same as in Expt. 3A
(the table of \enquote{previous experiments}; Figure
\ref{fig:causalExpt}A-C). Upon continuing beyond the first part of the
trial, the table of results and background story were removed from the
screen and the participant is told that the results of the \enquote{lost
experiment} were found (the experiment with a \enquote{?} in the
table).\footnote{We chose to have the scientists report on an earlier
  \enquote{lost experiment} to suggest a binomial generative process for
  the experiments wherein the scientists planned to perform 11
  experiments, as opposed to alternative design wherein an 11th
  experiment is reported. Such a continuation of the series of
  experiments could imply a generative model following a geometric
  distribution where the scientists repeat the experiment until they
  reach one that is successful.} The results are reported to the
participant in terms of how many out of 100 of the attempts were
successful. Participants saw one of two reported frequencies: 20\% or
70\% (randomized between-subjects). Participants were then asked to
judge the causal sentence (e.g., \enquote{Herb X makes the animals
sleepy}) by either clicking \enquote{Yes} or \enquote{No} (Figure
\ref{fig:causalExpt}E). This allows us to test whether endorsement of a
causal sentence for a given actual frequency is affected by the causal
priors induced by our manipulation. After responding, participants
completed the same attention check as Expt. 3a.

\hypertarget{results-3}{%
\subsubsection{Results}\label{results-3}}

42 participants were excluded from the analysis for failing to answer
both of the attention check questions correctly, leaving a total of 358
responses for analysis. As in our other analyses of endorsement
responses, we computed the Bayesian Maximum A-Posteriori (MAP) estimate
and 95\% highest probability density interval of the true population
probability of endorsing the statement, assuming a uniform prior. These
are shown for the different experimentally-manipulated priors and
referent prevalences in Figure~\ref{fig:figure-causals}B.

As predicted by our model, endorsements for a causal statement were
sensitive to the referent prevalence of causal events and, critically,
to the background distribution of other causes. When many other causes
produced the effect very reliably (\emph{common strong} condition), very
few participants endorsed the causal statement for a causal frequency of
0.2, and were at chance when the causal frequency was 0.7
(Figure~\ref{fig:figure-causals}, blue bars). By contrast, when many
other causes failed to produce the effect and those that did were not
very reliable (\emph{rare weak} condition; green in figure), at least
half of participants endorsed the causal statement for a cause with
causal prevalence of 0.2, and were at ceiling when the prevalence was
0.7. The other two conditions (\emph{rare deterministic} and
\emph{common weak}) led to endorsements intermediate between these two
conditions. These effects were predicted by our model with strong
quantitative accuracy (\(r^2(8) = 0.835\); MSE = \(0.0123\)).

\hypertarget{discussion-3}{%
\subsection{Discussion}\label{discussion-3}}

In our third case study, we applied our model to generalizations about
causal events, without any changes. In this domain, we successfully
manipulated participants' beliefs about the expected prevalence of a
causal relationship in a domain (Expt. 3A). This was done using both
unimodal (\emph{common weak}, \emph{common strong}) and bimodal
(\emph{rare weak}, \emph{rare strong}) distributions. In Expt. 3B, we
showed that these manipulated priors influenced endorsements of the
corresponding causal statements. In addition to further demonstrating
the generality of this theory, these experiments show that the
prevalence prior \(P(p)\) is causally related to endorsements of
generalizations in language.

In these experiments, we used two cover stories that described plausible
causal events: herbs making animals sleepy and fertilizers making plants
grow tall. We chose these items because there was a plausible causal
mechanism that could give rise to the property and these causal events
could have ambiguous causal power associated with them (e.g., it is
plausible that there are herbs that only weakly make animals sleepy and
it is also plausible that there are herbs that almost deterministically
make animals sleepy). These two features of the domains make them
particularly amenable to manipulation. It is likely that other domain
knowledge would interact with the experimentally-supplied
\enquote{experimental data} to form a hybrid belief
distribution.\footnote{If this were happening in our domains, we would
  expect this to show up in the results of Expt. 3A. Participants'
  predictions about the likely causal power of new causes would be
  expected to show a mixture of their abstract, intuitive theories and
  the experimentally supplied data.} For example, physical causal
systems (e.g., billiard balls hitting each other) could strongly induce
near-deterministic notions of causality, analogous to our \emph{strong}
priors conditions. Causal systems that demonstrate surprising or \emph{a
priori} unlikely effects (e.g., liquids melting concrete) could induce
rarity about the existence of a non-zero causal power, analogous to our
\emph{rare} prior conditions. Our theory would predict that differences
in endorsement in these cases would be mediated by differences in the
corresponding prevalence prior distributions.

\hypertarget{general-discussion}{%
\section{General Discussion}\label{general-discussion}}

The human species is remarkable not only because we can extract useful
generalizations from the world, but because we can convey these
generalizations to each other succinctly using language. Generalizations
expressed in language (e.g., \enquote{John runs.}, \enquote{Dogs are
friendly.}, \enquote{Fire causes smoke.}) are a premier example of how
simple statements---statements understood by even the youngest language
users---can convey rich meanings and display complex sensitivities to
context. We have argued that the core meaning of such linguistic
expressions can be understood by three ingredients: Probability,
vagueness, and context. That generics are vague does not preclude them
from being treated by formal models. A vague threshold meaning operating
over a speaker's inductive beliefs, formalized by predictive
probability, leads to complex, quantitative interactions with background
knowledge, which closely tracks human judgments.

We tested this theory by exploring its implications for the simplest
judgments a person can make about a sentence: an endorsement or truth
judgment. Truth judgments are the standard measurement for semantic
theories, but are often limited to the intuitions of a single or a few
trained theorists, often producing only binary or ternary judgments
(i.e., true, false, and possibly indeterminate). Our model accounted for
these standard linguistic intuitions about a number of philosophically
puzzling generic statements (\emph{Worked Examples}). We went beyond
intuitions, however, measuring truth judgments of naive language users,
which revealed substantial gradience in endorsements. These
fine-grained, quantitative measurements pose real challenges for verbal
theories of generics, which can only predict qualitative differences.
The framework we propose is sufficiently precise to predict quantitative
gradience, and we showed that the gradience in truth judgments depends
in systematic ways on an interpreter's prior beliefs about prevalence.
We provided strict tests of our formal model by both measuring and
manipulating listeners' background knowledge as well as comparing our
model to a number of alternatives. In each case, our model provided a
strong quantitative fit to human judgements while other models fell
short.

The fact that our model applies equally as well to generic and habitual
language sheds light on a long-attested relationship. The analogy of
habituality (e.g., \enquote{John runs}) and genericity (e.g.,
\enquote{Dogs have four legs}) has typically been assumed in the
literature following the original suggestion by Carlson (1977). The
relationship, however, has been never been empirically tested nor
formally described. This paper presents the first empirical evidence
that the reasoning involved in understanding generic and habitual
sentences are similar: The same uncertain threshold mechanism operating
over prior beliefs explains judgments of both kinds of statements
(Expts. 1 \& 2).

In our model, an utterance conveying a generalization updates a
listener's \emph{a priori} beliefs about the prevalence of the property
for that category. Since the prevalence prior includes the information
necessary to compute cue validity (see Appendix A), the cue validity of
the feature for the new category is added to the common ground between
interlocutors. Thus, given the knowledge represented in a prevalence
prior, generics communicate information both about the prevalence of the
feature in the new category (via the literal semantics) as well as the
cue validity of the feature for the new category. For example, if a
listener hears \enquote{Alligators grow to be 10-feet long}, they will
update their personal beliefs both about the prevalence of 10-footedness
among alligators as well as the probability that an animal is an
alligator given that it is 10 feet long. To our knowledge, this is the
first theory of generics that describes how the informational
contribution of a generic relates to cue validity.\footnote{We are
  grateful to an anonymous reviewer for drawing our attention to this. }
This also means that our computational framework makes predictions about
the cue validity implied by different generalizations, without any
further assumptions.

We propose that the language of generalization conveys predictive
probabilities. Communicating probabilities might seem contrary to the
long attested failures in reasoning about probability (Tversky \&
Kahneman, 1974). Our theory suggests that the problems with
understanding probabilities observed in classic, cognitive psychology
paradigms are a problem in understanding explicit probabilities
expressed using \emph{numerical language}, a historically quite recent
innovation (cf., Levinson, 1995). Rather than conveying probabilities
explicitly, the language of generalization conveys them implicitly. In
other words, we argue that the utterance \enquote{70\% of birds fly} is
a precise statement about how many birds fly in the way that
\enquote{John is 6'3}" is a precise statement about the height of John,
whereas \enquote{Birds fly} is a vague statement more akin to the
utterance \enquote{John is tall}. The latter statements are easier to
process, understood at an earlier age, and may be more useful for human
reasoning. Indeed, even infants are actually quite good with reasoning
about probabilities, but in ways that are not explicit or tied to the
numerical language of probability (Xu \& Denison, 2009; Xu \& Garcia,
2008).

In the rest of this discussion, we further discuss the contribution of
our modeling framework to empirical and theoretical work on the language
of generalization, elaborate further on the role of conceptual knowledge
in our modeling framework, describe how our model relates to language
understanding more generally including theories of vagueness, and sketch
an argument for how our formal representation of the language of
generalization could provide a relatively simple acquisition problem for
a language learner.

\hypertarget{generics-and-genericity}{%
\subsection{Generics and genericity}\label{generics-and-genericity}}

The statistical vs.~conceptual distinction for theories of generics,
described in detail earlier in this paper, is roughly co-extensive with
the distinction between theories of the truth conditions of generics and
the mental phenomenon of \emph{genericity}, respectively (Nickel, 2016).
Truth conditions are unified criteria that all generics must satisfy in
order to be true. \emph{Genericity}, on the other hand, is thought to be
a non-linguistic, psychological phenomenon previously explained either
in terms of pragmatics (Declerck, 1991) or meta-semantics (Leslie, 2007;
Liebesman, 2011; Nickel, 2016; Sterken, 2015). A full account of generic
language must include both a theory of generics (i.e., truth conditions)
and a complementary theory of genericity (Nickel, 2016).

Our model leverages the insights of formal semantics and computational
cognitive modeling to articulate how an agent should update their
beliefs based on a generic sentence. The Bayesian model provides a clean
separation of the semantics of a generic statement (represented formally
by a threshold function) from world knowledge (represented by the
prevalence prior). We, thus, chart-out a single answer to the semantic
question \enquote{what do generics mean?}, while also formalizing how
background knowledge influences generic understanding (a question about
genericity). Previous theoretical accounts have either aimed to account
for the context-sensitivity of generics by positing distinct semantic
constructs (e.g., \emph{relative} vs \emph{absolute} generics; Cohen,
1999) or positing a semantics that cannot be separated from world
knowledge (Leslie, 2008). The modeling approach we take here
demonstrates how the context-sensitivity of generics can emerge via the
interplay of a stable (but vague) semantics and diverse background
knowledge. This model is thus the first step towards unifying conceptual
approaches to genericity with an inherently quantitative natural
language semantics for generics.

\hypertarget{relation-to-intuitive-theories}{%
\subsection{Relation to intuitive
theories}\label{relation-to-intuitive-theories}}

Our theory assumes that listeners and speakers employ real-world
knowledge of objects and events in order to use and interpret the
language of generalization. It is this real-world knowledge (presently
formalized in terms of a prevalence prior) where issues about domain
specific beliefs, interpretation of properties (Prasada \& Dillingham,
2006), over-hypotheses (Leslie, 2008), and essentialism (Gelman, 2003)
can all play a role. We hypothesize that all of this knowledge feeds
into participants' predictions, measured through judgments about
prevalence. That is, even though our underlying semantics is defined in
terms prevalence of individual properties, this does not imply that the
concepts involved in generic language understanding are defined merely
by their properties. Instead, complex intuitive theories (that may
reflect, for example conceptual role, see Goodman et al., 2015) give
rise to prevalence priors, which then yield judgements about generic
statements. Thus, it is the output of these cognitive processes and
conceptual understanding---intuitions about prevalence---to which this
framework brings precision and clarity. Applying the same precision and
clarity to the more abstract aspects of conceptual knowledge including
property knowledge, essentialism, etc. is an important next step for
this line of work. Probabilistic causal models (Gopnik, 2003; Pearl,
1988) and their generalization in probabilistic programs (Goodman et
al., 2015) are obvious starting places to look for such a formalism for
conceptual structure and higher-order abstractions.

The fact that conceptual knowledge is not a part of our formal semantics
does not imply that such knowledge is unrelated to communicating
generalizations. In fact, we saw in Expt. 2C that it was participants'
predictions about the future that led to the prevalence computation
involved in the endorsement decision. Intuitive theories guide these
predictions, and might provide new insight into classic puzzles of
generics. We described in the Worked Examples how our model treats
\enquote{Supreme Court Justices have even social security numbers} as
infelicitous even when 100\% of justices have even SSNs, because a
speaker's subjective probability that the \emph{next justice} would have
an even SSN is likely 50\%. However, were we to learn much more
surprising information---for instance, if every Supreme Court Justice in
U.S. history had a social security number which was a \emph{prime
number} (a much lower probability outcome)---the shear suspiciousness
could compel an observer to revise their theory of the domain (appealing
perhaps to a conspiracy), update their subjective probability of future
instances, and endorse such a generic.

Not only does conceptual knowledge influence interpretation via beliefs
about prevalence, but conceptual knowledge may very well be what a
speaker intends to communicate. Like other models in the probabilistic
pragmatics tradition, our model distinguishes the relevant variables for
the truth-functional semantics (e.g., a threshold and a prevalence) from
those that impact the speaker's utility in producing the utterance. A
speaker's utility comes from addressing an implicit goal of
communication, sometimes referred to as the \emph{Question Under
Discussion} or QUD (Roberts, 1996). The separation of the QUD from the
literal, truth-functional denotation of the utterance is an important
theoretical distinction made explicit in our formal modeling approach
(Goodman \& Frank, 2016). For simplicity, the model presented here
assumes the QUD is about prevalence. However, the model makes nearly
identical predictions if the QUD concerns abstract parameters of the
prevalence distributions (e.g., if the speaker intends to convey that
``K is the kind of thing that Fs''; cf., Prasada \& Dillingham, 2006),
something much closer to communicating conceptual understanding. This
observation may prove useful for constructing models of speakers whose
goals are to convey abstract relations between kinds and properties.

Conceptual knowledge concerning the strikingness or dangerousness of
properties and their relation to generic language has been of particular
interest to psychologists and philosophers since Leslie (2007). We
described before that the strikingness of certain properties plausibly
influences prevalence priors or a speaker's predictions about future
prevalence. We have collected pilot data suggesting these constructs are
indeed altered as a result of introducing information about the
dangerousness of features to participants (e.g., as in Cimpian et al.,
2010). More empirical work is needed to understand whether and how
strikingness influences generic understanding above and beyond the
probabilistic constructs we have posited in our theory.

\hypertarget{the-comparison-class}{%
\subsection{The comparison class}\label{the-comparison-class}}

The probabilistic communicative model we introduce in this paper assumes
shared background knowledge about the statistics of an event or property
in question, represented by the prior belief distribution over the
prevalence \(P(p)\). We constructed prevalence priors for a property,
event, or cause by considering other possible categories having the
property, people doing the action, or causes producing the effect,
respectively. In Expt. 3, we empirically demonstrated the influence of
other categories on endorsements of causal generalizations.
Collectively, these other kinds, people, or causes form \emph{comparison
classes} against which the referent-category is evaluated. Thus, the
prevalence prior \(P(p)\) is actually a conditional distribution
constructed with respect to some comparison class \(C\):
\(P(p \mid C)\).

In this paper, we have assumed particular values of \(C\) for our three
case studies (\emph{animals}, \emph{people}, and \emph{possible causes},
respectively). We think the choice of these classes is intuitive, but it
is a limitation of this work that we do not derive these choices from
more general information-theoretic considerations. The problem of
choosing a comparison class, however, is not unique to the language of
generalization, but is a problem for any theory of \emph{vague} or
\emph{underspecified} language (e.g., gradable adjectives like
\emph{tall} and vague quantifiers like \emph{many}; Bale, 2011; Qing \&
Franke, 2014; Schmidt, Goodman, Barner, \& Tenenbaum, 2009; Solt, 2009;
Solt \& Gotzner, 2012). We have begun to explore how pragmatic reasoning
and world knowledge can flexibly adjust the comparison class to
appropriately suit the context (e.g., how a ``short'' basketball player
is \emph{short for a basketball player} while a ``tall'' basketball
player is \emph{tall for a person}; Tessler, Lopez-Brau, \& Goodman,
2017). It remains to be seen how such principles could operate on richly
structured, hierarchical knowledge about categories and properties that
would be important for interpreting generic language.

The employment of comparison classes in our model provides additional
flexibility towards modeling different communicative goals. The
comparison classes used in our case studies were constructed with
respect to the category (\emph{other animals}, \emph{other people},
\emph{other possible causes}). Thus, our endorsement model assumes that
the QUD was \enquote{\emph{what} has this feature?}. Any part of a
sentence, however, can be brought into focus (e.g., by prosody) and
turned into an answer to a QUD: \enquote{\emph{Dogs} have four legs} (as
opposed to other animals), \enquote{Dogs \emph{have} four legs} (as
opposed to eating or doing other things with four legs), \enquote{Dogs
have \emph{four} legs} (as opposed to two legs, three legs, six legs),
\enquote{Dogs have four \emph{legs}} (as opposed to having four of other
limbs/body parts), \enquote{Dogs \emph{have four legs}} (as opposed to
having other features). It has long been noted that the same
generalization can be used to address multiple QUDs (Krifka, 1995), and
we posit that differences in interpretation are the result of multiple
distinct comparison classes competing for influence. Prosody provides a
good cue towards resolving the comparison class, and pragmatic reasoning
is likely relevant here as well (Declerck, 1986; Tessler et al., 2017).
In this work, we focused on category-wise comparison classes purely for
methodological convenience. Future work should investigate the factors
that give rise to feature-wise interpretations of generalizations (e.g.,
\enquote{Dogs \emph{have four legs}}, as opposed to having other
features), other QUDs generics can address, how the relevant comparison
classes are constructed, and how these inferences interact with
threshold inference.

\hypertarget{genericity-and-vagueness}{%
\subsection{Genericity and vagueness}\label{genericity-and-vagueness}}

We have argued that generalizations in language are vague descriptions
of prevalence in a way analagous to how gradable adjectives like
\enquote{tall} are vague descriptions of some underlying degree scale
(e.g., height). Vague predicates like \enquote{tall}, however, exhibit a
number of additional phenomena that we do not suppose generics or
generalizations would necessarily exhibit. For example, vague predicates
admit borderline cases (e.g., a person who is neither tall nor not
tall). \emph{A priori}, it is unclear whether one could construct
borderline cases with generics or habituals without sounding like a
contradiction (e.g., \enquote{John both runs and doesn't run}).
Additionally, sorites paradoxes can be constructed out of arguments
using vague predicates, and it is not clear that generics or habituals
could be used to construct such arguments (e.g., a person who runs one
day a week less than John {[}who runs habitually{]} is still a person
who runs). No experimental data yet bears on these ideas, but would be
useful for further understanding the relationship between the language
of generalization and vague predicates.

For now, we understand both genericity and vagueness as analogous
linguistic phenomena insofar as they can both be thought of as the
result of a listener being uncertainty as to the exact truth conditions
(e.g., an uncertain threshold). This sort of \enquote{lexical
uncertainty} is a kind of \emph{parameter learning} problem (e.g., the
listener knows the form of the truth conditions, but not the values of
some variables) which can be formally distinguished from other sorts of
uncertain truth conditions better modeled as a \emph{structure learning}
problem (e.g., lexical ambiguity). Further work should be done to better
understand the kinds of context-sensitive linguistic phenomena that are
modeled as parameter vs.~structure learning problems to construct a
richer typology of underspecified linguistic expressions.

Our view of understanding the language of generalization involving
reasoning about an uncertain threshold is reminiscent of Williamson and
Simons (1992)'s epistemicist view on vagueness. This view holds that
when a speaker utters a vague statement, there exists an objective set
of criteria determining whether the entity in question satisfies the
vague predicate (i.e., whether or not the utterance is true). Vagueness
emerges from uncertainty about the details of those objective criteria
(e.g., the precise value of a threshold). This uncertainty in turn could
be the result of a population of speakers using slightly different
criteria. Listeners then will not know exactly what kind of speaker they
are dealing with \emph{a priori} and normatively should maintain
uncertainty about the truth conditions (see Lassiter \& Goodman, 2015
for an extended discussion of vagueness involving resolving the value of
a free semantic variable in context).

\hypertarget{acquiring-the-language-of-generalization}{%
\subsection{Acquiring the language of
generalization}\label{acquiring-the-language-of-generalization}}

Perhaps the most surprising aspect of the language of generalization is
how difficult it is to formalize, given how common it is in
child-directed and child-produced speech (Gelman, Coley, Rosengren,
Hartman, \& Pappas, 1998; Gelman et al., 2004). Generics are often
contrasted with quantifier language (e.g., \enquote{some},
\enquote{most}, \enquote{all}), whose truth conditions are easy to
formalize but which pose difficulty for young children to acquire
(Brandone, Gelman, \& Hedglen, 2014; Gelman et al., 2015). Leslie (2008)
argues that the necessary complexity of a formal account of generics and
the simplicity with which young children acquire generics imply that the
normal tools for describing the semantics of quantified utterances
(i.e., a truth-functional threshold) are inappropriate for
generics.\footnote{This argument is a purely semantic argument which
  ignores evidence that children's difficulty with quantifiers stems
  from pragmatic issues (Musolino \& Lidz, 2006).}

Rejecting the tools of truth-functional semantics for the language of
generalization would be throwing the baby out with the bathwater: In
fact, from a learning perspective, our semantics of generics can be
viewed as \emph{simpler} than that of quantifier semantics. With some
reasonable assumptions about the hypothesis space of semantic meanings,
the acquisition of a threshold-based truth-functional meaning
\(\mbox{ $[\![ u ]\!]$}(p, \theta) := p > \theta\) requires learning
three distinct aspects of meaning: (1) the \emph{dimension} being
described (i.e., prevalence \(p\)), (2) the \emph{polarity} of the
relation (i.e., \(>\) vs. \(<\)), and (3) the value of the
\emph{threshold} (e.g., \(\theta = 0\) for \enquote{some},
\(\theta = 0.5\) for \enquote{most}).\footnote{This particular framing
  of the acquisition process ignores the potentially different impact of
  pragmatic reasoning on generics vs.~quantifiers. } If a learner first
acquires the dimension and the polarity (i.e., generics have to do with
prevalence in some positive way), a rational learner should then
represent uncertainty over possible thresholds \(\theta\). For
quantifier semantics, a learner would then need to learn the
context-invariant value of the threshold \(\theta\), but for generics,
she could have an adult-like semantics for generics by maintaining
uncertainty about the threshold. That is, the language of generalization
is learned once aspects (1) and (2) are understood. Thus, under our
framework, generics should not only be learned before quantifiers but
could also facilitate the acquisition of quantifiers because of their
shared logical form.

There is a secondary argument for why our semantics presents an easy
learning problem. We have assumed the truth functional threshold
\(\theta\) comes from a uniform distribution over the unit interval
\([0, 1]\), which is mathematically equivalent to a
\emph{continuously-valued} or \emph{soft semantics} wherein the degree
to which the utterance is true is proportional to the degree itself, in
this case prevalence \(p\):
\(\int_{0}^{1} \delta_{p > \theta} \mathop{}\!\mathrm{d}\theta = p\).
The model for generic interpretation (Eq. \ref{eq:L0}) then becomes:
\(L(p \mid u) \propto p \cdot P(p)\) This \emph{soft semantics}
(corresponding intuitively to a meaning like \enquote{the higher the
prevalence, the better}, or simply \enquote{more is better}) is perhaps
the simplest quantitative semantics one could posit. The difficulty in
acquiring the meaning of quantifiers, then, is a difficulty in
recognizing a fixed-threshold semantics as a special case of this
\emph{more is better} semantics. We leave for future work the precise
implementation of such an acquisition model.

\hypertarget{conclusion}{%
\section{Conclusion}\label{conclusion}}

It might seem paradoxical that a part of language that is so common in
communication and central to learning should be vague. Shouldn't
speakers and teachers want to express their ideas as crisply as
possible? To the contrary, underspecification can be efficient, given
that context can be relied upon to resolve uncertainty in the moment
(Piantadosi, Tily, \& Gibson, 2012). In our work, context takes the form
of shared beliefs between speakers and listeners. By leveraging this
common ground, the language of generalizations provides a powerful way
to communicate and learn abstract knowledge, which would otherwise be
difficult or costly information to acquire through direct experience.

Categories are inherently unobservable. You cannot see the category
\textsc{dog}, only some number of instances of it. Yet we easily talk
about these abstractions, conveying hard-won generalizations to each
other and down through generations. The theory presented here provides
the first computational perspective on how we communicate
generalizations, illustrating how beliefs play a central role in
understanding the meaning of words.

\newpage

\hypertarget{references}{%
\section{References}\label{references}}

\hypertarget{refs}{}
\leavevmode\hypertarget{ref-Asher1995}{}%
Asher, N., \& Morreau, M. (1995). What some generic sentences mean.
\emph{The Generic Book}, 300--338.

\leavevmode\hypertarget{ref-Bale2011}{}%
Bale, A. C. (2011). Scales and comparison classes. \emph{Natural
Language Semantics}, \emph{19}, 169--190.

\leavevmode\hypertarget{ref-Behrens2005}{}%
Behrens, L. (2005). Genericity from a cross-linguistic perspective.
\emph{Linguistics}, \emph{43}(2), 275--344.

\leavevmode\hypertarget{ref-Brandone2014}{}%
Brandone, A. C., Gelman, S. A., \& Hedglen, J. (2014). Children's
Developing Intuitions About the Truth Conditions and Implications of
Novel Generics Versus Quantified Statements. \emph{Cognitive Science},
1--28.

\leavevmode\hypertarget{ref-Carlson1977}{}%
Carlson, G. N. (1977). \emph{Reference to kinds in english}
(PhD thesis). University of Massachusetts, Amherst.

\leavevmode\hypertarget{ref-Carlson1995essay}{}%
Carlson, G. N. (1995). Truth conditions of generic sentences: Two
contrasting views. In G. N. Carlson \& F. J. Pelletier (Eds.), \emph{The
generic book} (pp. 224--38). Chicago: University of Chicago Press.

\leavevmode\hypertarget{ref-Carlson2005}{}%
Carlson, G. N. (2005). Generics, Habituals and Iteratives. In \emph{The
encyclopedia of language and linguistics} (2nd ed.). Elsevier.
doi:\href{https://doi.org/10.1007/s13398-014-0173-7.2}{10.1007/s13398-014-0173-7.2}

\leavevmode\hypertarget{ref-Carlson1995}{}%
Carlson, G. N., \& Pelletier, F. J. (1995). \emph{The generic book.}
Chicago, IL: Chicago University Press.

\leavevmode\hypertarget{ref-Cheng1997}{}%
Cheng, P. W. (1997). From Covariation to Causation: A Causal Power
Theory. \emph{Psychological Review}, \emph{104}(2), 367--405. Retrieved
from \url{http://psycnet.apa.org/fulltext/1997-03612-007.pdf}

\leavevmode\hypertarget{ref-Cheng1992}{}%
Cheng, P. W., \& Novick, L. R. (1992). Covariation in Natural Causal
Induction. \emph{Psychological Review}, \emph{99}(2), 365--382.
Retrieved from \url{http://psycnet.apa.org/fulltext/1992-26096-001.pdf}

\leavevmode\hypertarget{ref-Cimpian2007}{}%
Cimpian, A., Arce, H.-M. C., Dweck, C. S., \& Markman, E. M. (2007).
Subtle Linguistic Cues Affect Children's Motivation. \emph{Psychological
Science}, \emph{18}(4), 314--316.

\leavevmode\hypertarget{ref-Cimpian2010}{}%
Cimpian, A., Brandone, A. C., \& Gelman, S. A. (2010). Generic
statements require little evidence for acceptance but have powerful
implications. \emph{Cognitive Science}, \emph{34}(8), 1452--1482.

\leavevmode\hypertarget{ref-Cimpian2008}{}%
Cimpian, A., \& Markman, E. M. (2008). Preschool children's use of cues
to generic meaning. \emph{Cognition}, \emph{107}, 19--53.
doi:\href{https://doi.org/10.1016/j.cognition.2007.07.008}{10.1016/j.cognition.2007.07.008}

\leavevmode\hypertarget{ref-Clark1996}{}%
Clark, H. H. (1996). \emph{Using language}. Cambridge University Press.

\leavevmode\hypertarget{ref-Cohen1999}{}%
Cohen, A. (1999). Generics, Frequency Adverbs, and Probability.
\emph{Linguistics and Philosophy}, \emph{22}.

\leavevmode\hypertarget{ref-Cohen2004}{}%
Cohen, A. (2004). Generics and Mental Representations. \emph{Linguistics
and Philosophy}, \emph{27}(5), 529--556.
doi:\href{https://doi.org/10.1023/B:LING.0000033851.25870.3e}{10.1023/B:LING.0000033851.25870.3e}

\leavevmode\hypertarget{ref-Cree2006}{}%
Cree, G. S., McNorgan, C., \& McRae, K. (2006). Distinctive features
hold a privileged status in the computation of word meaning:
Implications for theories of semantic memory. \emph{Journal of
Experimental Psychology. Learning, Memory, and Cognition}, \emph{32}(4),
643--658.
doi:\href{https://doi.org/10.1037/0278-7393.32.4.643}{10.1037/0278-7393.32.4.643}

\leavevmode\hypertarget{ref-Dahl1975}{}%
Dahl, Ö. (1975). On generics. In E. Keenan (Ed.), \emph{Formal semantics
of natural language} (pp. 99--111). Cambridge: Cambridge University
Press.

\leavevmode\hypertarget{ref-Declerck1986}{}%
Declerck, R. (1986). The manifold interpretations of generic sentences.
\emph{Lingua}, \emph{68}, 149--188.

\leavevmode\hypertarget{ref-Declerck1991}{}%
Declerck, R. (1991). The origins of genericity. \emph{Linguistics},
\emph{29}, 79--102.

\leavevmode\hypertarget{ref-Degen2014}{}%
Degen, J., \& Goodman, N. D. (2014). Lost your marbles? The puzzle of
dependent measures in experimental pragmatics. In \emph{Proceedings of
the thirty-sixth annual conference of the Cognitive Science Society}.

\leavevmode\hypertarget{ref-Frank2012}{}%
Frank, M. C., \& Goodman, N. D. (2012). Predicting pragmatic reasoning
in language games. \emph{Science}, \emph{336}(6084).

\leavevmode\hypertarget{ref-Franke2014cogsci}{}%
Franke, M. (2014). Typical use of quantifiers: A probabilistic speaker
model. In \emph{Proceedings of the 36th annual meeting of the cogntiive
science society} (pp. 1--6). Retrieved from
\href{http://staff.science.uva.nl/\%7B~\%7Dmfranke/Papers/Franke\%7B/_\%7D2014\%7B/_\%7DTypical\%20use\%20of\%20quantifiers\%20A\%20probabilistic\%20speaker\%20model.pdf}{http://staff.science.uva.nl/\{\textasciitilde{}\}mfranke/Papers/Franke\{\textbackslash{}\_\}2014\{\textbackslash{}\_\}Typical use of quantifiers A probabilistic speaker model.pdf}

\leavevmode\hypertarget{ref-Gelman2003}{}%
Gelman, S. A. (2003). \emph{Essential child: Origins of
essentialntialism in everyday thought.} Oxford University Press.

\leavevmode\hypertarget{ref-Gelman2004}{}%
Gelman, S. A. (2004). Learning words for kinds: Generic noun phrases in
acquisition. In \emph{Weaving a lexicon} (pp. 445--484). MIT Press.

\leavevmode\hypertarget{ref-Gelman2009}{}%
Gelman, S. A. (2009). Learning from others: Children's construction of
concepts. \emph{Annual Review of Psychology}, \emph{60}, 115--140.
doi:\href{https://doi.org/10.1146/annurev.psych.59.103006.093659.LEARNING}{10.1146/annurev.psych.59.103006.093659.LEARNING}

\leavevmode\hypertarget{ref-Gelman1998}{}%
Gelman, S. A., Coley, J. D., Rosengren, K. S., Hartman, E., \& Pappas,
A. (1998). Beyond labeling: the role of maternal input in the
acquisition of richly structured categories. \emph{Monographs of the
Society for Research in Child Development}, \emph{63}(1), I--V,
1--148;discussion 149--157.

\leavevmode\hypertarget{ref-Gelman2008}{}%
Gelman, S. A., Goetz, P. J., Sarnecka, B. W., \& Flukes, J. (2008).
Generic Language in Parent-Child Conversations. \emph{Language Learning
and Development}, \emph{4}(1), 1--31.
doi:\href{https://doi.org/10.1080/15475440701542625.Generic}{10.1080/15475440701542625.Generic}

\leavevmode\hypertarget{ref-Gelman2015}{}%
Gelman, S. A., Leslie, S.-J., Was, A. M., \& Koch, C. M. (2015).
Children's interpretations of general quantifiers, specific quantifiers
and generics. \emph{Language, Cognition and Neuroscience}, \emph{30}(4),
448--461.
doi:\href{https://doi.org/10.1080/23273798.2014.931591}{10.1080/23273798.2014.931591}

\leavevmode\hypertarget{ref-Gelman2003b}{}%
Gelman, S. A., \& Raman, L. (2003). Preschool children use linguistic
form class and pragmatic cues to interpret generics. \emph{Child
Development}, \emph{74}(1), 308--325.

\leavevmode\hypertarget{ref-GelmanEtAl2004}{}%
Gelman, S. A., Taylor, M. G., Nguyen, S. P., Leaper, C., \& Bigler, R.
S. (2004). Mother-child conversations about gender: Understanding the
acquisition of essentialist beliefs. \emph{Monographs of the Society for
Research in Child Development}, \emph{69}(1), vii, 116--127.
doi:\href{https://doi.org/10.1111/j.1540-5834.2004.06901001.x}{10.1111/j.1540-5834.2004.06901001.x}

\leavevmode\hypertarget{ref-Gerstenberg2015how}{}%
Gerstenberg, T., Goodman, N. D., Lagnado, D. A., \& Tenenbaum, J. B.
(2015). How, whether, why: Causal judgments as counterfactual contrasts.
In D. C. Noelle, R. Dale, A. S. Warlaumont, J. Yoshimi, J. Matlock T.,
C. D., \& P. P. Maglio (Eds.), \emph{Proceedings of the 37th Annual
Conference of the Cognitive Science Society} (pp. 782--787). Austin, TX:
Cognitive Science Society.

\leavevmode\hypertarget{ref-Goodman1955}{}%
Goodman, N. (1955). \emph{Fact, fiction, and forecast}. Harvard
University Press.

\leavevmode\hypertarget{ref-Goodman2016}{}%
Goodman, N. D., \& Frank, M. C. (2016). Pragmatic language
interpretation as probabilistic inference. \emph{Trends in Cognitive
Sciences}.

\leavevmode\hypertarget{ref-GoodmanLassiter}{}%
Goodman, N. D., \& Lassiter, D. (2015). Probabilistic semantics and
pragmatics: Uncertainty in language and thought. In S. Lappin \& C. Fox
(Eds.), \emph{The handbook of contemporary semantic theory, 2nd
edition}. Wiley-Blackwell.

\leavevmode\hypertarget{ref-dippl}{}%
Goodman, N. D., \& Stuhlmüller, A. (2014). The Design and Implementation
of Probabilistic Programming Languages. \url{http://dippl.org}.

\leavevmode\hypertarget{ref-Goodmanconcepts}{}%
Goodman, N. D., Tenenbaum, J. B., \& Gerstenberg, T. (2015). Concepts in
a probabilistic language of thought. In \emph{The conceptual mind: New
directions in the study of concepts}. MIT Press.

\leavevmode\hypertarget{ref-Gopnik2003theory}{}%
Gopnik, A. (2003). The theory theory as an alternative to the innateness
hypothesis. \emph{Chomsky and His Critics}, 238--254.

\leavevmode\hypertarget{ref-Grice1975}{}%
Grice, H. P. (1975). Logic and conversation. In \emph{Readings in
language and mind}. Blackwell.

\leavevmode\hypertarget{ref-Griffiths2005}{}%
Griffiths, T. L., \& Tenenbaum, J. B. (2005). Structure and strength in
causal induction. \emph{Cognitive Psychology}, \emph{51}(4), 334--84.
doi:\href{https://doi.org/10.1016/j.cogpsych.2005.05.004}{10.1016/j.cogpsych.2005.05.004}

\leavevmode\hypertarget{ref-Griffiths2009}{}%
Griffiths, T. L., \& Tenenbaum, J. B. (2009). Theory-Based Causal
Induction. \emph{Psychological Review}, \emph{116}(4), 661--716.
doi:\href{https://doi.org/10.1037/a0017201}{10.1037/a0017201}

\leavevmode\hypertarget{ref-Henrich2015}{}%
Henrich, J. (2015). \emph{The secret of our success: how culture is
driving human evolution, domesticating our species, and making us
smarter}. Princeton University Press.

\leavevmode\hypertarget{ref-Herbelot2011}{}%
Herbelot, A., \& Copestake, A. (2011). Formalising and specifying
underquantification Quantification resolution. In \emph{IWCS} (pp.
165--174).

\leavevmode\hypertarget{ref-HumeTHN}{}%
Hume, D. (1888). (L. A. Selby Bigge, Ed.). Oxford, Clarendon Press.

\leavevmode\hypertarget{ref-Icard2017}{}%
Icard, T. F., Kominsky, J. F., \& Knobe, J. (2017). Normality and actual
causal strength. \emph{Cognition}, \emph{161}, 80--93.
doi:\href{https://doi.org/10.1016/j.cognition.2017.01.010}{10.1016/j.cognition.2017.01.010}

\leavevmode\hypertarget{ref-Kemp2007}{}%
Kemp, C., Perfors, A., \& Tenenbaum, J. B. (2007). Learning
overhypotheses with hierarchical Bayesian models. \emph{Developmental
Science}, \emph{10}(3), 307--21.
doi:\href{https://doi.org/10.1111/j.1467-7687.2007.00585.x}{10.1111/j.1467-7687.2007.00585.x}

\leavevmode\hypertarget{ref-Kennedy2007}{}%
Kennedy, C. (2007). Vagueness and grammar: the semantics of relative and
absolute gradable adjectives. \emph{Linguistics and Philosophy},
\emph{30}, 1--35.
doi:\href{https://doi.org/10.1007/s10988-006-9008-0}{10.1007/s10988-006-9008-0}

\leavevmode\hypertarget{ref-Khemlani2012}{}%
Khemlani, S., Leslie, S.-J., \& Glucksberg, S. (2012). Inferences about
members of kinds: The generics hypothesis. \emph{Language and Cognitive
Processes}, \emph{27}(6), 887--900.
doi:\href{https://doi.org/10.1080/01690965.2011.601900}{10.1080/01690965.2011.601900}

\leavevmode\hypertarget{ref-KrifkaGenericBookFocus}{}%
Krifka, M. (1995). Focus and the interpretation of generic sentences. In
G. N. Carlson \& F. J. Pelletier (Eds.), \emph{The generic book} (pp.
238--264). Chicago: University of Chicago Press.

\leavevmode\hypertarget{ref-Lassiter2013}{}%
Lassiter, D., \& Goodman, N. D. (2013). Context, scale structure, and
statistics in the interpretation of positive-form adjectives. In
\emph{Semantics and Linguistic Theory (SALT) 23}.

\leavevmode\hypertarget{ref-Lassiter2015}{}%
Lassiter, D., \& Goodman, N. D. (2015). Adjectival vagueness in a
bayesian model of interpretation. \emph{Synthese}.

\leavevmode\hypertarget{ref-Leslie2007}{}%
Leslie, S.-J. (2007). Generics and the Structure of the Mind.
\emph{Philosophical Perspectives}, \emph{21}(1), 375--403.

\leavevmode\hypertarget{ref-Leslie2008}{}%
Leslie, S.-J. (2008). Generics: Cognition and acquisition.
\emph{Philosophical Review}, \emph{117}(1).

\leavevmode\hypertarget{ref-Leslie2012}{}%
Leslie, S.-J., \& Gelman, S. A. (2012). Quantified statements are
recalled as generics: evidence from preschool children and adults.
\emph{Cognitive Psychology}, \emph{64}(3), 186--214.
doi:\href{https://doi.org/10.1016/j.cogpsych.2011.12.001}{10.1016/j.cogpsych.2011.12.001}

\leavevmode\hypertarget{ref-Levinson1995}{}%
Levinson, S. C. (1995). Interactional biases in human thinking. In E. N.
Goody (Ed.), \emph{Social intelligence and interaction} (pp. 221--260).
Cambridge University Press.

\leavevmode\hypertarget{ref-Liebesman2011}{}%
Liebesman, D. (2011). Simple generics. \emph{Noûs}, \emph{45}(3),
409--442.

\leavevmode\hypertarget{ref-McGuire1986}{}%
McGuire, W. J., \& McGuire, C. V. (1986). Differences in conceptualizing
self versus conceptualizing other people as manifested in contrasting
verb types used in natural speech. \emph{Journal of Personality and
Social Psychology}, \emph{51}(6), 1135--43.
doi:\href{https://doi.org/10.1037/0022-3514.51.6.1135}{10.1037/0022-3514.51.6.1135}

\leavevmode\hypertarget{ref-Montague1973}{}%
Montague, R. (1973). The Proper Treatment of Quantification in Ordinary
English. In \emph{Philosophy, language, and artificial intelligence}
(pp. 141-----162). Springer. Retrieved from
\url{http://semantics.uchicago.edu/kennedy/classes/s08/semantics2/montague73.pdf}

\leavevmode\hypertarget{ref-Musolino2006}{}%
Musolino, J., \& Lidz, J. (2006). Why children aren't universally
successful with quantification. \emph{Linguistics}, \emph{44}(4),
817--852.

\leavevmode\hypertarget{ref-Nickel2008}{}%
Nickel, B. (2008). Generics and the ways of normality. \emph{Linguistics
and Philosophy}, \emph{31}(6), 629--648.
doi:\href{https://doi.org/10.1007/s10988-008-9049-7}{10.1007/s10988-008-9049-7}

\leavevmode\hypertarget{ref-Nickel2016}{}%
Nickel, B. (2016). \emph{Between logic and the world}. Oxford: Oxford
UP; Oxford UP.

\leavevmode\hypertarget{ref-Nisbett1983}{}%
Nisbett, R. E., Krantz, D. H., Jepson, C., \& Kunda, Z. (1983). The use
of statistical heuristics in everyday inductive reasoning.
\emph{Psychological Review}, \emph{90}(4), 339--363.
doi:\href{https://doi.org/10.1037/0033-295X.90.4.339}{10.1037/0033-295X.90.4.339}

\leavevmode\hypertarget{ref-Orvell2017}{}%
Orvell, A., Kross, E., \& Gelman, S. A. (2017). How ``you'' makes
meaning. \emph{Science}, \emph{355}(6331). Retrieved from
\url{http://science.sciencemag.org/content/355/6331/1299.full}

\leavevmode\hypertarget{ref-pearl1988probabilistic}{}%
Pearl, J. (1988). Probabilistic reasoning in intelligent systems:
Networks of plausible reasoning. Morgan Kaufmann Publishers, Los Altos.

\leavevmode\hypertarget{ref-Pelletier1997}{}%
Pelletier, F. J., \& Asher, N. (1997). Generics and defaults. In J. van
Benthem \& A. Ter Meulen (Eds.), \emph{Handbook of logic and language}
(pp. 1125--1177). Cambridge: The MIT Press.

\leavevmode\hypertarget{ref-Piantadosi2012}{}%
Piantadosi, S. T., Tily, H., \& Gibson, E. (2012). The communicative
function of ambiguity in language. \emph{Cognition}, \emph{122}(3),
280--291.
doi:\href{https://doi.org/10.1016/j.cognition.2011.10.004}{10.1016/j.cognition.2011.10.004}

\leavevmode\hypertarget{ref-Prasada2000}{}%
Prasada, S. (2000). Acquiring generic knowledge. \emph{Trends in
Cognitive Sciences}, \emph{4}(2), 66--72.
doi:\href{https://doi.org/10.1016/S1364-6613(99)01429-1}{10.1016/S1364-6613(99)01429-1}

\leavevmode\hypertarget{ref-Prasada2006}{}%
Prasada, S., \& Dillingham, E. M. (2006). Principled and statistical
connections in common sense conception. \emph{Cognition}, \emph{99}(1),
73--112.
doi:\href{https://doi.org/10.1016/j.cognition.2005.01.003}{10.1016/j.cognition.2005.01.003}

\leavevmode\hypertarget{ref-Prasada2012}{}%
Prasada, S., Hennefield, L., \& Otap, D. (2012). Conceptual and
Linguistic Representations of Kinds and Classes. \emph{Cognitive
Science}, \emph{36}(7), 1224--1250.
doi:\href{https://doi.org/10.1111/j.1551-6709.2012.01254.x}{10.1111/j.1551-6709.2012.01254.x}

\leavevmode\hypertarget{ref-Prasada2013}{}%
Prasada, S., Khemlani, S., Leslie, S.-J., \& Glucksberg, S. (2013).
Conceptual distinctions amongst generics. \emph{Cognition},
\emph{126}(3), 405--22.
doi:\href{https://doi.org/10.1016/j.cognition.2012.11.010}{10.1016/j.cognition.2012.11.010}

\leavevmode\hypertarget{ref-Qing2014}{}%
Qing, C., \& Franke, M. (2014). Gradable adjectives, vagueness, and
optimal language use: A speaker-oriented model. In \emph{Semantics and
Linguistic Theory (SALT) 24}.

\leavevmode\hypertarget{ref-Repacholi1997}{}%
Repacholi, B. M., \& Gopnik, A. (1997). Early reasoning about desires:
evidence from 14- and 18-month-olds. \emph{Developmental Psychology},
\emph{33}(1), 12--21.
doi:\href{https://doi.org/10.1037/0012-1649.33.1.12}{10.1037/0012-1649.33.1.12}

\leavevmode\hypertarget{ref-Rhodes2012}{}%
Rhodes, M., Leslie, S.-J., \& Tworek, C. M. (2012). Cultural
transmission of social essentialism. \emph{Proceedings of the National
Academy of Sciences}, \emph{109}(34), 13526--13531.
doi:\href{https://doi.org/10.1073/pnas.1208951109}{10.1073/pnas.1208951109}

\leavevmode\hypertarget{ref-Ritchie2016}{}%
Ritchie, D., Stuhlmüller, A., \& Goodman, N. D. (2016). C3: Lightweight
incrementalized mcmc for probabilistic programs using continuations and
callsite caching. In \emph{AISTATS 2016}.

\leavevmode\hypertarget{ref-roberts1996qud}{}%
Roberts, C. (1996). Information structure in discourse: Towards an
integrated formal theory of pragmatics. \emph{Working Papers in
Linguistics-Ohio State University Department of Linguistics}, 91--136.

\leavevmode\hypertarget{ref-hurdleModels}{}%
Rose, C. E., Martin, S. W., Wannemuehler, K. A., \& Plikaytis, B. D.
(2006). On the use of zero-inflated and hurdle models for modeling
vaccine adverse event count data. \emph{Journal of Biopharmaceutical
Statistics}, \emph{16}(4), 463--481.

\leavevmode\hypertarget{ref-Rothbart1978}{}%
Rothbart, M., Fulero, S., Jensen, C., Howard, J., \& Birrell, P. (1978).
From individual to group impressions: Availability heuristics in
stereotype formation. \emph{Journal of Experimental Social Psychology},
\emph{14}(3), 237--255.
doi:\href{https://doi.org/10.1016/0022-1031(78)90013-6}{10.1016/0022-1031(78)90013-6}

\leavevmode\hypertarget{ref-Schmidt2009}{}%
Schmidt, L. A., Goodman, N. D., Barner, D., \& Tenenbaum, J. B. (2009).
How Tall Is Tall? Compositionality, Statistics, and Gradable Adjectives.
In \emph{Proceedings of the 31st annual conference of the cognitive
science society}.

\leavevmode\hypertarget{ref-Seiver2013}{}%
Seiver, E., Gopnik, A., \& Goodman, N. D. (2013). Did She Jump Because
She Was the Big Sister or Because the Trampoline Was Safe? Causal
Inference and the Development of Social Attribution. \emph{Child
Development}, \emph{84}(2), 443--454.
doi:\href{https://doi.org/10.1111/j.1467-8624.2012.01865.x}{10.1111/j.1467-8624.2012.01865.x}

\leavevmode\hypertarget{ref-Shepard1987}{}%
Shepard, R. N. (1987). Toward a universal law of generalization for
psychological science. \emph{Science (New York, N.Y.)},
\emph{237}(4820), 1317--1323.
doi:\href{https://doi.org/10.1126/science.3629243}{10.1126/science.3629243}

\leavevmode\hypertarget{ref-Shipley1993}{}%
Shipley, E. F. (1993). Categories, hierarchies, and induction. \emph{The
Psychology of Learning and Motivation}, \emph{30}, 265--301.

\leavevmode\hypertarget{ref-Solt2009}{}%
Solt, S. (2009). Notes on the Comparison Class. In \emph{International
workshop on vagueness in communication} (pp. 189-----206). Springer.

\leavevmode\hypertarget{ref-Solt2012}{}%
Solt, S., \& Gotzner, N. (2012). Experimenting with degree.
\emph{Proceedings of SALT}, \emph{22}, 166--187.

\leavevmode\hypertarget{ref-Sterken2015}{}%
Sterken, R. K. (2015). Generics in Context. \emph{Philosophers'
Imprint}, \emph{15}(i), 1--30.

\leavevmode\hypertarget{ref-Tenenbaum2001}{}%
Tenenbaum, J. B., \& Griffiths, T. L. (2001). Generalization, similarity
and Bayesian inference. \emph{Behavioral and Brain Sciences}, \emph{24},
629--640.
doi:\href{https://doi.org/10.1017/S0140525X01000061}{10.1017/S0140525X01000061}

\leavevmode\hypertarget{ref-Tenenbaum2011}{}%
Tenenbaum, J. B., Kemp, C., Griffiths, T. L., \& Goodman, N. D. (2011).
How to grow a mind: statistics, structure, and abstraction.
\emph{Science (New York, N.Y.)}, \emph{331}(6022), 1279--85.
doi:\href{https://doi.org/10.1126/science.1192788}{10.1126/science.1192788}

\leavevmode\hypertarget{ref-Tessler2018negant}{}%
Tessler, M. H., \& Franke, M. (2018). Not unreasonable: Carving vague
dimensions with contraries and contradictions.

\leavevmode\hypertarget{ref-Tessler2017}{}%
Tessler, M. H., Lopez-Brau, M., \& Goodman, N. D. (2017). Warm (for
winter): Comparison class understanding in vague language. In
\emph{Proceedings of the thirty-ninth annual conference of the cognitive
science society}.

\leavevmode\hypertarget{ref-Tomasello1999}{}%
Tomasello, M. (1999). \emph{The cultural origins of human cognition}.
Cambridge, MA: Harvard University Press.

\leavevmode\hypertarget{ref-Tversky1974}{}%
Tversky, A., \& Kahneman, D. (1974). Judgment under Uncertainty:
Heuristics and Biases. \emph{Science}, \emph{185}(4157), 1124--31.
doi:\href{https://doi.org/10.1126/science.185.4157.1124}{10.1126/science.185.4157.1124}

\leavevmode\hypertarget{ref-Williamson1992}{}%
Williamson, T., \& Simons, P. (1992). Vagueness and ignorance.
\emph{Proceedings of the Aristotelian Society, Supplementary Volumes},
\emph{66}, 145--177.

\leavevmode\hypertarget{ref-xu2009statistical}{}%
Xu, F., \& Denison, S. (2009). Statistical inference and sensitivity to
sampling in 11-month-old infants. \emph{Cognition}, \emph{112}(1),
97--104.

\leavevmode\hypertarget{ref-Xu2008}{}%
Xu, F., \& Garcia, V. (2008). Intuitive statistics by 8-month-old
infants. \emph{Proceedings of the National Academy of Sciences of the
United States of America}, \emph{105}(13), 5012--5.
doi:\href{https://doi.org/10.1073/pnas.0704450105}{10.1073/pnas.0704450105}

\newpage

\hypertarget{appendix-a-the-relationship-between-the-prevalence-prior-and-cue-validity}{%
\section{Appendix A: The Relationship between the Prevalence Prior and
Cue
Validity}\label{appendix-a-the-relationship-between-the-prevalence-prior-and-cue-validity}}

Cue validity is defined for a particular category--property pair (e.g.,
\emph{mosquitos} and \emph{carry malaria}), and relates to the referent
prevalence (e.g., how many mosquitos carry malaria) via Bayes' Rule:
\[ P(k \mid f) = \frac{P(f \mid k) \cdot P(k)}{Z} \] where
\(Z = \sum_{k' \in K} P( f \mid k') \cdot P( k')\), the average or
marginal prevalence of the feature (e.g., \emph{carrying malaria}) in
other categories \(k'\).

The prevalence prior \(P(p)\) using the generic interpretation model
(Eq. \ref{eq:L0}) is a probability distribution over prevalence for
different categories \(k'\).

\underline{Claim:} The normalizing constant for computing cue validity
is equal to the expected value of the prevalence prior distribution:
\(\mathbb{E}[P(p)] = Z\)

\underline{Proof:}

By the definition of the expectation of a distribution:

\begin{eqnarray} \label{eq:expectation}
\mathbb{E}[P(p)] & = & \sum\limits_{p} p \cdot P(p) 
\end{eqnarray}

The probability of a prevalence \(p\) can be decomposed into the prior
probability of a category \(k\) and the likelihood of the prevalence
\(p\) given that category \(k\): \(P(p) = P(p \mid k) \cdot P(k)\). We
assume here, without loss of generality, that each category corresponds
to one and only one prevalence \(p\). Thus, \(P(p \mid k) = 1\) if
\(k \in K_p\), a set of categories that have a given prevalence:
\(K_p = \{k' : p_{k'} = p\}\). Then, consider the partition of the set
of all categories \(K\) into non-overlapping \(K_p\). Thus:

\begin{eqnarray} \label{eq:prevToKinds}
P(p) & = & \sum\limits_{k' \in K_p} P(p \mid k') \cdot P( k') \nonumber \\
     & = & \sum\limits_{k' \in K_p} P(k')
\end{eqnarray}

since \(\forall k' \in K_{p}: P(p \mid k') = 1\). Returning to Eq.
\ref{eq:expectation}, we have:

\begin{eqnarray} \label{eq:eq3}
\mathbb{E}[P(p)] & = & \sum\limits_{p} p \sum\limits_{k' \in K_p} P(k') \nonumber \\
                  & = & \sum\limits_{p} \sum\limits_{k' \in K_p} p \cdot P(k')
\end{eqnarray}

The set of all partitioned subsets \(K_p\) is in a one-to-one
correspondence with the set of all prevalences \(p\). Thus, we have:

\begin{eqnarray} \label{eq:bijection}
  & = & \sum\limits_{K_p} \sum\limits_{k' \in K_p} p \cdot P( k')
\end{eqnarray}

Then, since \(\cup_{p}{K_p} = K\), we have

\begin{eqnarray} \label{eq:partition}
  & = & \sum\limits_{k' \in K} p \cdot P( k') \nonumber \\ 
  & = & \sum\limits_{k' \in K} P(f \mid k') \cdot P( k') \nonumber \\ 
  & = & Z
\end{eqnarray}

\(\square.\)

\newpage

\hypertarget{appendix-b-measuring-cue-validity}{%
\section{Appendix B: Measuring Cue
Validity}\label{appendix-b-measuring-cue-validity}}

In Experiment 1, we articulated an alternative model by measuring cue
validity (and prevalence) and predicting generic endorsement from a
linear combination of these parameters. In a small review of the
literature, we discovered different methods for measuring cue validity;
in piloting, we found these different methods led to different results.
Therefore, we propose three \emph{a priori} desiderata that a
measurement of cue validity should satisfy. We describe two experiments
that represent the primary methods for measuring cue validity and
compare them with these desiderata in mind. Finally, we compare the cue
validity measured using these different methods to cue validity derived
from our prevalence prior elicitation task (Expt. 1b, main text).

\hypertarget{desiderata}{%
\subsection{Desiderata}\label{desiderata}}

Measuring cue validity involves collecting participants' judgments that
relate to the probability that an exemplar is a member of a kind given
that it has a feature: \(P(x \in k \mid x \in f)\). There are several
ways one could measure cue validity. Here we consider two measures: The
first has participants estimate the cue validity probability
\(P(x \in k \mid x \in f)\) directly, a common technique in the
literature on generic language (e.g., Khemlani et al., 2012; Prasada et
al., 2013), and another has participants freely produce categories given
a feature (i.e., draw a sample from the conditional distribution on
kinds given a feature; Cree et al., 2006). We will refer to the former
as the \emph{direct question} method and the latter as the \emph{free
production} method.

Are the \emph{direct question} and the \emph{free productions} equally
valid for measuring cue validity? We propose \emph{a priori} three
boundary conditions that a measurement of cue validity should satisfy.
For each, we provide four examples from our larger stimulus set on
generics (Case Study 1) which will be used to evaluate each measure.

\begin{enumerate}
\item{Completely diagnostic features: Some features are only present in one (or a very small number) of categories. Examples include: \emph{carrying malaria} (mosquitos), \emph{carrying Lyme disease} (ticks, deer), \emph{having manes} (lions), \emph{having pouches} (marsupials, including most famously: kangaroos). The cue validity of these features for the corresponding categories should be very high (at least 0.5 and possibly close to 1).}
\item{Completely absent features: Many features are completely absent in many kinds. For these, the cue validity should be extremely low or 0. There are infinite examples. The ones we will use are \emph{has wings} (leopard), \emph{has a mane} (shark), \emph{has spots} (kangaroo), \emph{has a pouch} (tiger). }
\item{Completely undiagnostic features: A number of features are shared by almost every category. The cue validity of these features for particular categories should be extremely low or 0. The ones we will use are: \emph{is female} (robin), \emph{is male} (lion), \emph{is juvenile} (kangaroo), \emph{is full-grown} (leopard).  Learning that an entity is female tells you almost nothing about what kind of animal it is.}
\end{enumerate}

We collected cue validity ratings by running both a direct question and
a free production experiment. For the free production experiment, the
cue validity is the proportion of responses of the target category
(e.g., \enquote{mosquitos}) for the property (e.g., \enquote{carries
malaria}). Of primary interest is the measurement for the desiderata
items described above. Links to the experiments can be found on
\url{https://github.com/mhtess/genlang-paper}.

\hypertarget{experimental-materials}{%
\subsection{Experimental materials}\label{experimental-materials}}

Materials were the same for both experiments. These were a collection of
familiar properties and animal categories used in Expt. 1a (endorsement
of generic statements) described in the main text. There were twenty-one
properties in total.

\hypertarget{direct-question-experiment}{%
\subsection{Direct question
experiment}\label{direct-question-experiment}}

\hypertarget{method-7}{%
\subsubsection{Method}\label{method-7}}

\hypertarget{participants-7}{%
\paragraph{Participants}\label{participants-7}}

We recruited 40 participants from Amazon's Mechanical Turk. Participants
were restricted to those with U.S. IP addresses and who had at least a
95\% work approval rating. The experiment took on average 5 minutes and
participants were compensated \$0.75 for their work.

\hypertarget{procedure-4}{%
\paragraph{Procedure}\label{procedure-4}}

Following the procedure in Khemlani et al. (2012) and Prasada et al.
(2013), participants were presented with prompts of the following form:

\begin{quotation}
Imagine you come across a thing that \textsc{f}.
What are the odds that it is a \textsc{k}?
\end{quotation}

Participants responded using a slider bar with endpoints labeled
\enquote{unlikely} and \enquote{likely}. The slider appeared with no
handle present; participants had to click on the slider for the slider
handle to appear.

Participants completed the thirty target trials (corresponding to the
thirty generic statements used in Expt. 1a) in addition to ten filler
trials (total number of trials = 40). The filler trials were made up of
random category--property pairings. Trials were presented in a
randomized order.

\hypertarget{free-production-experiment}{%
\subsection{Free production
experiment}\label{free-production-experiment}}

\hypertarget{method-8}{%
\subsubsection{Method}\label{method-8}}

\hypertarget{participants-8}{%
\paragraph{Participants}\label{participants-8}}

We recruited 100 participants from Amazon's Mechanical Turk.
Participants were restricted to those with U.S. IP addresses and who had
at least a 95\% work approval rating. The experiment took on average 3
minutes and participants were compensated \$0.40 for their work.

\hypertarget{procedure-5}{%
\paragraph{Procedure}\label{procedure-5}}

On each trial, participants were presented with prompts of the following
form:

\begin{quotation}
Imagine you come across a thing (animal or insect) that \textsc{f}.
What do you think it is?
\end{quotation}

Participants responded by filling in a text box with their response for
twenty-one trials in total, one for each property. No filler trials were
used. Trials were presented in a randomized order.

\hypertarget{free-production-data-processing}{%
\subsubsection{Free production data
processing}\label{free-production-data-processing}}

To process the free production, we forced all characters in a response
to lower case, removed spaces, and made all terms into singular terms
(e.g., \enquote{lions} --\textgreater{} \enquote{lion}). As well,
\enquote{mosquito} was a commonly misspelled label; we counted anything
that started with \enquote{mosqu}, \enquote{mesqu}, \enquote{misqu},
\enquote{mosiq} as \enquote{mosquito}.

To calculate confidence intervals for the free production data, we
resampled participants (with replacement) and computed the proportion of
responses that were of the target category (e.g., the proportion of
\enquote{mosquito} responses for the cue \enquote{carries malaria}). We
did this 1000 times to generate an empirical distribution from which
95\% intervals could be calculated.

\hypertarget{results-and-evaluation}{%
\subsection{Results and Evaluation}\label{results-and-evaluation}}

We are interested in the results of each measure (direct question and
free production) for the three conditions corresponding to the
desiderata outlined above. To evaluate each measure, we selected four
example property--category pairs that we believe are unambiguous
instances of the boundary conditions described above (these items are
described above with the desiderata).

\begin{figure}[!h]
\includegraphics[width=\textwidth]{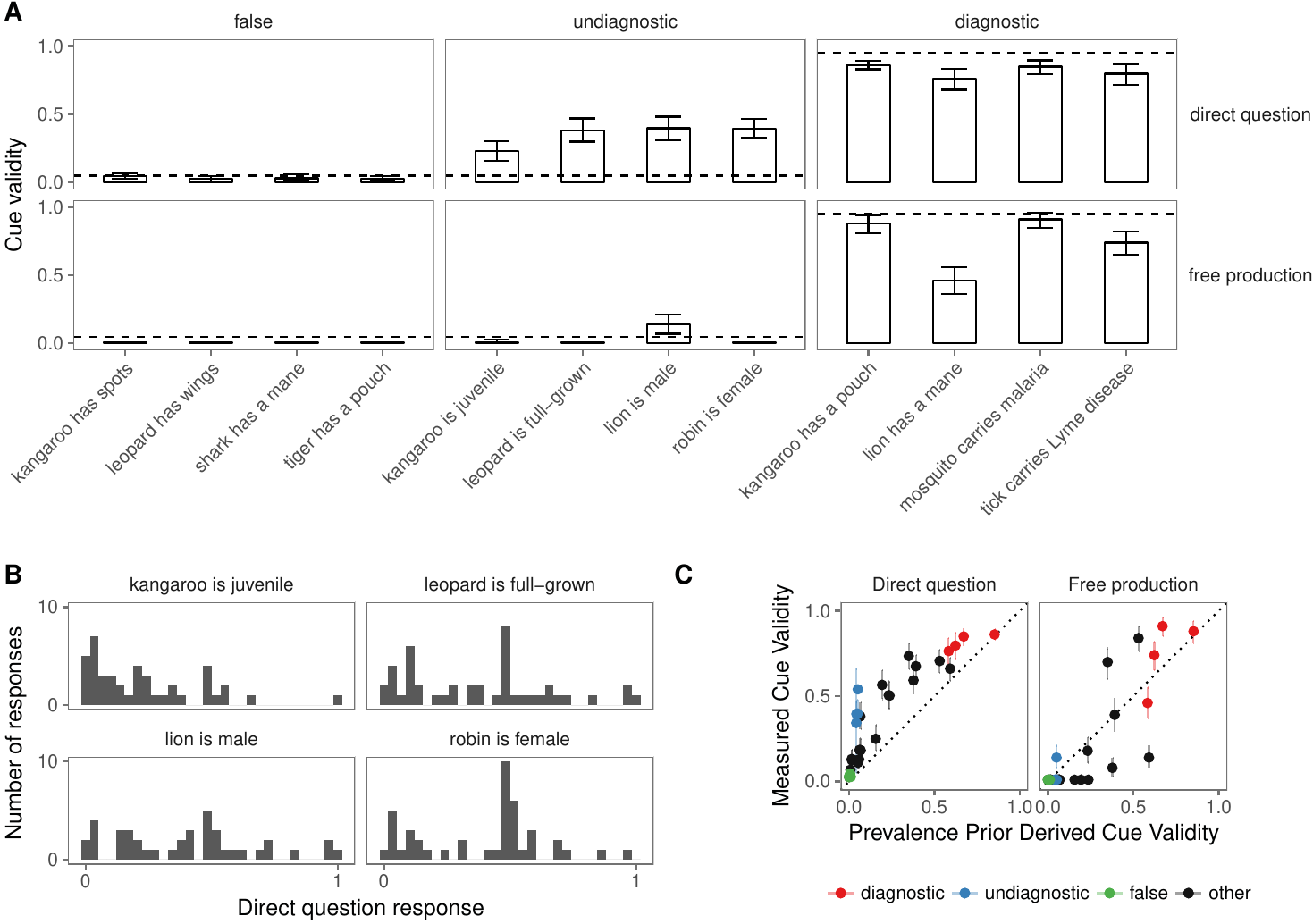} \caption{A: Empirically measured cue validity for two different tasks. Items are grouped by whether the property is never present in the category (false), the property is always present in the category and every other category (undiagnostic), or present in the category and absent from most other categories (diagnostic). Dotted lines denote theoretical cue validity representing the desiderata (see text).  B: Raw empirical distributions for the undiagnostic features in the direct question task. C: Correspondence between measured cue validity and prevalence prior derived cue validity.}\label{fig:cvBothquestionsBarplots}
\end{figure}

Figure~\ref{fig:cvBothquestionsBarplots}A shows the results for the
twelve items of interest for both measurements. We see that for the
\emph{false features}, both measures behave as desired (hypothesized
results shown by the dotted line): The cue validity of a feature that is
not present in the category is zero or near-zero. For \emph{diagnostic
features}, both measures also behave reasonably: Learning that an entity
has malaria strongly implies that it is a mosquito. However, there is
some evidence that the free production measurement is more sensitive
than the direct-question measure. \enquote{Having a mane} is strongly
diagnostic for a \enquote{lion} but also for a \enquote{horse} (and so
the overall cue validity of having a mane for a lion is around 0.5).
\enquote{Carrying Lyme disease} is mostly diagnostic for a
\enquote{tick} but also \enquote{deer} (and thus, the cue validity for
tick is not maximal). These subtle differences among diagnostic features
are picked up by the free-production measure but not by the
direct-question measure.

The free production and direct question measures deviate most strongly
in their characterization of the undiagnostic features. Learning that an
entity is female should not imply that it is a robin, which is
accurately reflected in the free production measure but not in the
direct question measure. Figure~\ref{fig:cvBothquestionsBarplots}B shows
the raw empirical distributions of responses for the direct question
measure for undiagnositic features. We observe that participants respond
to this question for undiagnostic features in one of two ways: (i)
reporting near-0 likelihood (hypothesized response) or (ii) reported
near-0.5 likelihood. This latter response option may reflect
participants \enquote{opting out} of a response (e.g., signalling
\enquote{I don't know}). For example, in response to the question
\enquote{There is a thing that is female. What are the odds that it is a
robin?}, a person could say they have no evidence to suggest that it is,
besides the very fact that the experimenter asked the question.
Participants may cope with the awkwardness of the question by placing
the slider bar in the middle of the scale.

\hypertarget{comparison-with-prevalence-prior-derived-cue-validity}{%
\subsection{Comparison with prevalence prior derived cue
validity}\label{comparison-with-prevalence-prior-derived-cue-validity}}

To further understand these measures of cue validity, we compare them to
cue validity derived from our prevalence prior elicitation task (Expt.
1b). Expt. 1b is not perfectly designed for this comparison, as we
supplied participants with half of the animal categories that they rated
(the other half was freely generated by participants); including these
supplied categories was important to measure the referent-prevalence of
interest (e.g., the percentage of mosquitos that carry malaria).
Including them in this analysis, however, potentially distorts the prior
probability of categories \(P(k)\).

In this analysis, we treat each category entry from Expt. 1b
(participant free production or experimentally supplied category like
mosquitos) as contributing to the prevalence prior. This results in the
prevalence prior favoring kinds that are easy to produce (like dogs and
cats; plausibly a good approximation for \(P(k)\)) as well as favoring
the experimentally supplied kinds (like mosquitos and robins).

We compute cue validity from the prevalence prior using Bayes' rule.
Figure~\ref{fig:cvBothquestionsBarplots}C shows the two measurements of
cue validity as they relate to the prevalence prior derived cue
validity. Across the thirty property--category pairs, the prevalence
prior derived cue validity is highly associated with both measurements:
\(r_{direct}^2(30) = 0.782\); \(MSE_{direct} = 0.0481\) and
\(r_{free}^2(30) = 0.739\); \(MSE_{free} = 0.0244\). Of primary interest
is how the measurements behave for desiderata items. We see that the
prevalence prior derived cue validity converges with the free production
measurement for the desiderata items (\(r_{free}^2(12) = 0.947\);
\(MSE_{free} = 0.00822\)), whereas the direct question measurement
overestimates the cue validity of undiagnositic features
(\(r_{direct}^2(12) = 0.779\); \(MSE_{direct} = 0.0561\)).

The points of largest deviation for the free-production measurement from
the prevalence prior derived measurement occur where the prevalence
prior derived measure rates the cue validity as relatively high when the
free-production measure gives the item low cue validity (two black
points in Figure~\ref{fig:cvBothquestionsBarplots}C, high X-value, low
Y-value). These two items are: (\enquote{is red}, \enquote{cardinal})
and (\enquote{is white}, \enquote{swan}). These items should have
relatively low cue validity and are overestimated by the prevalence
prior because of the prior on categories \(P(k)\) over-represents the
categories that were supplied to every participant and thus get a higher
weighting in the prior for deriving cue validity.\footnote{This
  deviation could be reduced in future experiments by performing a fully
  free production version of the prevalence prior task (i.e., without
  supplying the referent categories).} In comparison, the direct
question measurement overestimates cue validity for almost all of the
items relative to the cue validity derived from the prevalence prior.

\hypertarget{summary}{%
\subsection{Summary}\label{summary}}

\emph{Cue validity} is a commonly used measurement for understanding
generic truth judgments (e.g., Khemlani et al., 2012; Prasada et al.,
2013). We observed different measurements used in the literature and
articulated three \emph{a priori} desiderata to validate a measure of
cue validity. We found that the \enquote{free production} measurement
(i.e., participants freely produced categories given a feature), and not
the direct question measurement (i.e., participants provide a likelihood
judgment of a particular category given the feature), satisfied all
three boundary conditions. In addition, cue validity derived from our
prevalence prior measurement (Expt. 1b) also satisfied these boundary
conditions. Researchers interested in comparing cue validity to generic
truth judgments should use a free production paradigm for measuring cue
validity.

\newpage

\hypertarget{appendix-c-bayesian-data-analysis}{%
\section{Appendix C: Bayesian Data
Analysis}\label{appendix-c-bayesian-data-analysis}}

In our three case studies, we compare an information-theoretic,
computational model of endorsement to human endorsements of the language
of generalization. The model has a single free parameter: the optimality
parameter \(\lambda\) in Eq. \ref{eq:S1}. We analyze this model using a
Bayesian data analytic approach, where we jointly infer the value of
this single model parameter \(\lambda\) together with parameters that
govern the prevalence priors \(P(p)\) (Eq. \ref{eq:L0}) and referent
prevalence \(p\) (Eq. \ref{eq:S1}) for each item. To pin down the
prevalence prior and referent prevalence parameters, we use the data
directly related to those parameters (e.g., prior elicitation data).
Incorporating all data sources into a single Bayesian data analysis
model is the appropriate way to track measurement uncertainty for all
measurements simultaneously. In this appendix, we describe this
procedure in more detail for each case study.

\hypertarget{modeling-prevalence-priors}{%
\subsection{Modeling prevalence
priors}\label{modeling-prevalence-priors}}

In Case Study 1: Generic Language (Expt. 1b), we elicited the prevalence
prior by asking participants about the prevalence of features for
individual categories. We performed an analogous elicitation in Case
Study 3: Causal Language (Expt. 3a). We describe the analysis using
generics as our running example, but a parallel analysis was done for
causal language.

Participants' responses in the prior elicitation task can be thought of
as samples from the prevalence prior distribution. Formally, we assume
the prior data (analyzed independently for each property) was generated
from one of two underlying distributions: a distribution corresponding
to those kinds with a stable causal mechanism that \emph{could} give
rise to the property (\(\mathcal{D}_{stable}\)) and a \enquote{transient
cause} distribution corresponding to those kinds without a stable
mechanism (\(\mathcal{D}_{transient}\)). The \enquote{transient}
distribution intuitively corresponds to categories that do not have the
feature normally, but potentially could acquire the feature by
accidental forces (e.g., a lion, who through some genetic mutation,
reproduces by laying eggs). We model this distribution as a Beta
distribuition that heavily favors probabilities near 0:
\(\text{Beta}(\gamma = 0.01, \delta = 100)\).\footnote{Note that we use
  the noncanonical mean \(\gamma\) and concentration \(\xi\) (or,
  inverse-variance) parameterization of the Beta distribution rather
  than the canonical shape (or pseudocount) parameterization for ease of
  posterior inference. The shape parameterization can be recovered
  using: \(\alpha = \gamma \cdot \xi; \beta = (1 - \gamma) \cdot \xi\).}
The \enquote{stable} distribution is modeled as a Beta distribution with
unknown parameters \(\text{Beta}(\gamma, \xi)\).\footnote{Because the
  Beta distribution is not defined at the points 0 and 1, we add
  \(\epsilon\) to the 0 responses and round 1 to 0.99. Similar results
  can be obtained by rounding 0 to 0.01. Alternatively, the
  \enquote{transient} distribution could be defined as a Delta
  distribution at 0, and 0 responses could remain in their raw form.
  Adjusting 1 to \(1- \epsilon\) leads to improper inferences for this
  simple 2-component model, as \(1 - \epsilon\) is only likely under a
  highly left-skewed distribution; treating 1 as \(1- \epsilon\)
  disproportionately influences the shape of \(\mathcal{D}_{stable}\),
  forcing it to favor probabilities close to 1. This problem does not
  appear for 0 being adjusted to \(\epsilon\) because the
  \enquote{transient} distribution already expects such low values.}
Finally, we assume that these two components combine with mixture
weighting \(\phi\) such that the data we observe is
\[P(d) = \phi\cdot \text{Beta} (d \mid \gamma, \xi) + (1 -  \phi) \cdot \text{Beta}(d \mid \gamma = 0.01, \xi = 100) \].
We put the following priors over the latent parameters of the model:
\begin{eqnarray*}
\phi_i & \sim & \text{Uniform}(0, 1) \\
\gamma_i & \sim & \text{Uniform}(0, 1) \\
\xi_i & \sim & \text{Uniform}(0, 100)
\end{eqnarray*} where \(i\) ranges over the different properties (e.g.,
\emph{lays eggs}, \emph{carries malaria}).

To learn about the credible values of the parameters, we ran separate
MCMC chains for each item, collecting 75,000 samples, removing the first
25,000 for burn-in. To see how well the mixture model fits the
prevalence prior data, we use the inferred parameters to generate new
data. The data generated from the model's posterior is called the
\emph{posterior predictive distribution} and is an important step in
model criticism. If the model is a good representation of the data, the
posterior predictive data will align with the observed experimental
data. We construct a posterior predictive distribution by
\enquote{forward sampling} the model (i.e., generating new data given
the inferred parameter values).\footnote{This forward sampling can be
  described by the following algorithm: First, flip a coin weighted by
  \(\phi\). If it comes up heads, we then sample from the
  \enquote{stable} component: \(\text{Beta}(\gamma, \xi)\). If it comes
  up tails, we sample from the \enquote{transient} component:
  \(\text{Beta}(0.01, 100)\). We do this many times using the posterior
  distibution to generate a distribution over predicted prevalence
  ratings.} Representative posterior predictive results are shown in
Figure~\ref{fig:generic-endorsement-priors-figure}B (main text).

In Case Study 2 on habitual language (Expt. 2a), we asked participants
about parameters of this mixture model (by having participants answer
questions about different kinds of people) rather than having
participants give samples (e.g., by listing their friends and family
members, and rating how often they did certain actions). In pilot
testing, we found these different methodologies to give comparable
results and we opted to ask about hypothetical people to probe about
potentially undesirable habits of participants' friends and familiy
(e.g., how often they use cocaine). The questions used in this
structured elicitation task are described in the main text.

\hypertarget{modeling-referent-prevalence}{%
\subsection{Modeling
referent-prevalence}\label{modeling-referent-prevalence}}

In Case Study 1 (generics), we used participants prevalence ratings for
the category-of-interest in our generic sentences as the
referent-prevalence that is used in the endorsement model (Eq.
\ref{eq:S1}). For a given generic sentence (e.g., \enquote{Robins lay
eggs}), we took the prevalence ratings for the referent-category (e.g.,
the percentage of robins that lay eggs) from the prior elicitation task
(Expt. 1b) and assumed those were generated from a single Beta
distribution. We assumed the following priors on the parameters:
\begin{eqnarray*}
\gamma_i & \sim & \text{Uniform}(0, 1) \\
\xi_i & \sim & \text{Uniform}(0, 100)
\end{eqnarray*} We took samples from the posterior predictive of this
Beta distribution (i.e., reconstructed prevalence ratings) as the
\emph{referent-prevalence} used in the model.

In Expt. 2b (Habitual endorsement), we used the frequency given to
participants in the experimental prompt (e.g., 3 times in the past week)
as the referent-prevalence. In Expt. 2c (\enquote{What is prevalence?}),
we compared two endorsement models that differed in their representation
of referent-prevalence. For one model (past frequency model), the actual
frequency given to participants in the experimental prompt was assumed
to be the referent prevalence (same as in Expt. 2b); for the other model
(predictive frequency model), we used the mean elicited frequency from
the \emph{predictive frequency} condition (participants predictions
about how often the person would do the action in the next time
interval; see main text for details). In Case Study 3 (Causal
endorsement), we used the proportion of successful causal events given
to participants in the experimental prompt (e.g., 70 out of 100 uses of
Herb C made animals sleepy).

\hypertarget{jointly-modeling-referent-prevalence-prevalence-priors-and-generic-endorsements}{%
\subsection{Jointly modeling referent-prevalence, prevalence priors, and
generic
endorsements}\label{jointly-modeling-referent-prevalence-prevalence-priors-and-generic-endorsements}}

To fit the generic endorsement models, we incorporate them into the
Bayesian data analytic model of the prevalence prior data (described
above) to create a single, joint-inference model where the optimality
parameter \(\lambda\) (Eq. \ref{eq:S1}) is inferred jointly with all the
other latent parameters of the full model (the referent-prevalence \(p\)
for each category \(k\) and property \(f\) and the parameters of the
prevalence priors \(P(p)\) for each property \(f\)) using data from
Expt. 1a \& b (Figure \ref{fig:genericsModelDiagram}). For the
parameters of the prevalence priors, we use the same priors described in
Expt. 1b; for the speaker optimality parameter, we use a prior with a
range consistent with the previous literature that uses the same model
class: \(\lambda \sim \text{Uniform}(0,5)\). We learn about the
\emph{a posteriori} credible values of the joint inference models by
collecting samples from 3 MCMC chains of 150,000 iterations removing the
first 50,000 iterations for burn-in, using an incrementalized version of
the Metropolis-Hastings algorithm (Ritchie et al., 2016). This algorithm
is useful for models with many variables that only affect a subset of
the full model's predictions (e.g., models with by-item or
by-participant parameters, wherein those additional parameters mostly
only influence predictions for those items or participants).

\begin{figure}

{\centering \includegraphics[width=0.8\textwidth]{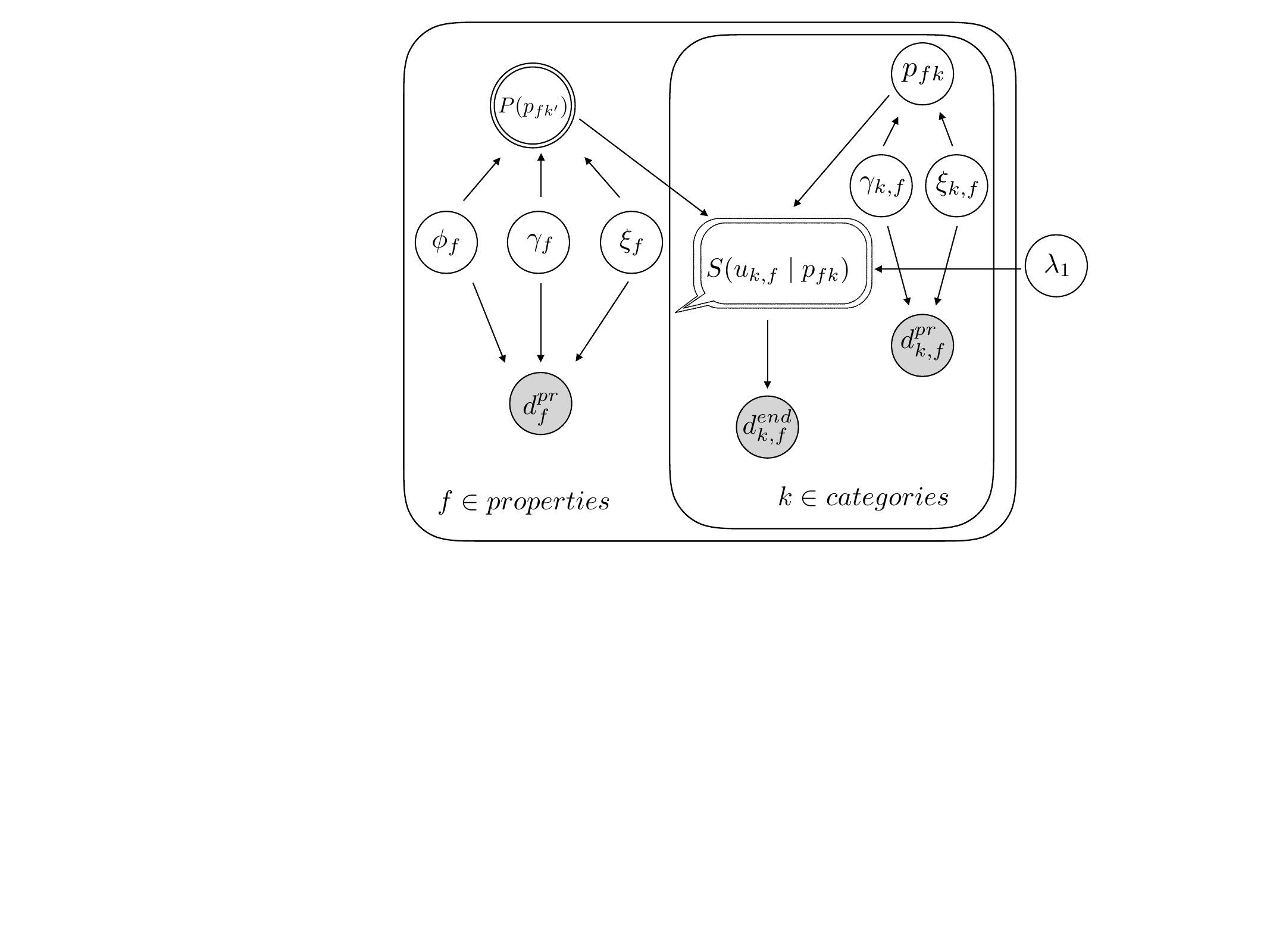} 

}

\caption{Quasi-graphical model corresponding to the fully Bayesian data analysis of the endorsement model for Case Study 1 (generic language). The prevalence prior data $d^{pr}_{f}$ is assumed to be generated from the mixture model validated in Expt. 1a, which has three parameters: mean of the stable-cause distribution $\gamma_{f}$ and concentration (inverse-variance) of the stable-cause distribution $\xi{f}$ and the mixture parameter $\phi_{f}$. The referent-prevalence $d^{pr}_{k,f}$ is generated from a Beta distribution with paramters: mean $\gamma_{k,f}$ and concentration (inverse-variance) $\xi{k,f}$. The posterior predictives of the prevalence prior $P(p_{fk'})$ and the referent prevalence $p_{fk}$ are then fed into the RSA speaker model $S$, which predicts the generic endorsement data $d^{end}_{k, f}$. The speaker model $S$ also takes in the single free parameter $\lambda$ which operates as a soft-max function. This overall structured is repeated (except $\lambda$) for each of the unique properties $f$ and categories $k$ that correspond to the generic sentences in our stimulus set. Note that $S$ corresponds to a probabilistic function and not a random variable that is standard in graphical model notation; $S$ cannot be represented by a graphical model because it has recursion. This entire BDA model is duplicated for the lesioned, fixed-threshold model (which only differs in the definition of $S$). The BDA model for habituals and causals mirrors this one, except they do not infer a referent prevalence $p_{fk}$ (they are assumed to be the same as those experimentally supplied to participants).}\label{fig:genericsModelDiagram}
\end{figure}

\hypertarget{supplementary-model-criticism}{%
\subsection{Supplementary model
criticism}\label{supplementary-model-criticism}}

In addition to examining the posterior predictive distribution on
endorsement judgments (presented in main text), we examined the marginal
posteriors on parameters of the prevalence priors and
referent-prevalence. These marginal distributions are important to
confirm that they have not changed substantially from the parameters
inferred from their primary data sources in isolation. For example, when
modeling the referent-prevalence data in isolation, the model infers
that roughly 65\% of robins lay eggs, as that is what participants on
average produce in the prevalence elicitation task.\footnote{Most
  participants report that 50\% of robins lay eggs, while a minority
  respond 100\%.} If the joint inference model (which models all data
sources---referent prevalence, prevalence prior, and generic
endorsement---simultaneously) infers referent-prevalence values
substantially different from those inferred by a model of referent
prevalence in isolation, that would suggest that the joint-model is
distorting the prevalence parameters to accommodate the endorsement
data. Such a result would call into question the inferences we as
scientists derive from the joint inference model. For example,
incorporating a linear regression model (of the kind presented as
alterantive models in the main text) into this Bayesian joint-inference
analysis model produces posterior predictions that match the generic
endorsement data surprisingly well (e.g., that model predicts
\enquote{Robins lay eggs} is true). Such a model is only able to do
this, however, by distorting the referent-prevalence data, inferring
that 100\% of robins lay eggs; thus, the linear model in this
joint-inference analysis framework sacrifices its goodness-of-fit to the
referent-prevalence data in order to increase its goodness-of-fit to the
endorsement data.\footnote{This distortion effect is why we accout for
  measurement uncertainty in the linear models by bootstrapping the data
  that forms their predictors, rather than performing a Bayesian
  analysis.} Such a distortion manifests as a difference between the
inferred parameters given only the referent-prevalence data and given
the full joint model (all data sources simultaneously).

\begin{figure}

{\centering \includegraphics[width=1\linewidth]{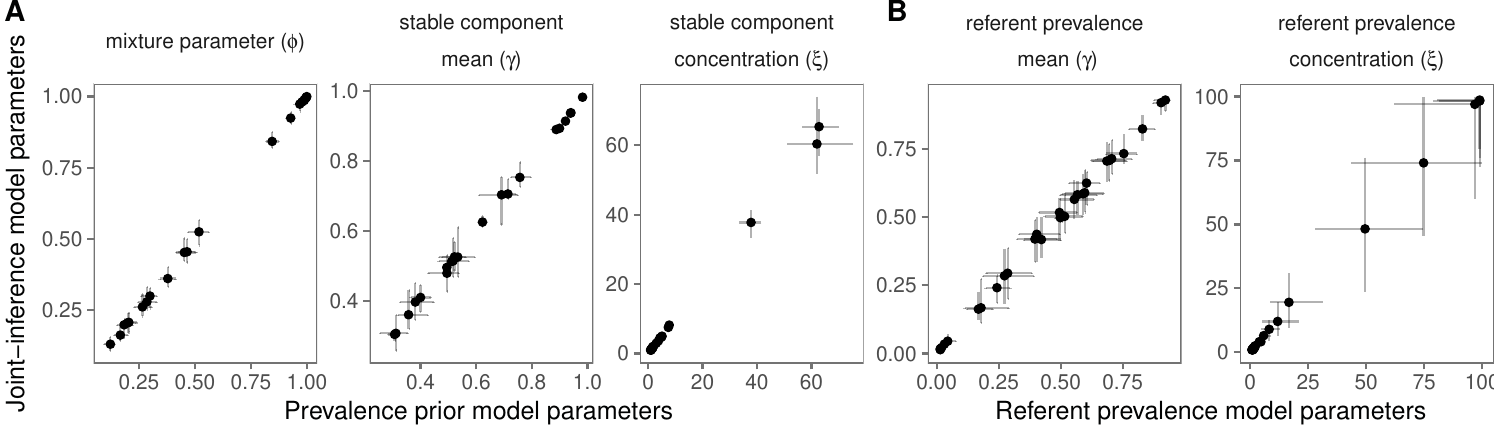} 

}

\caption{No evidence for parameter distortion caused by a joint-inference model. X-axis denotes Maximum A-Posteriori (MAP) estimates inferred using only a single data source (A: Prevalence prior data. B: Referent prevalence data). Y-axis denotes MAP estimates inferred using the joint-inference model which models all data sources simultaneously. Distortions would manifest by systematic deviations from the y=x line.}\label{fig:distortion}
\end{figure}

To investigate this distortion effect in the parameters, we compare the
values inferred for the parameters governing the prevalence priors and
referent prevalence variables before and after the joint-inference model
sees the generic endorsement data. Specifically, we infer the parameters
for prevalence priors and referent prevalence by constructing single
models of these tasks and compare the inferred values to those that
result from the joint-inference model. The inferred parameters for these
two models are shown in Figure~\ref{fig:distortion}. We found that the
referent-prevalences and prevalence priors inferred under the joint
model were almost indistinguishable from those inferred using only the
referent-prevalence and prevalence prior data, respectively (numerical
results reported in main text). This results confirms that modeling all
three data sources simultaneously does not distort some data sources
(e.g., referent prevalence) in order to provide good fits for others
(e.g., generic endorsement).

\newpage

\hypertarget{appendix-d-habituals-items}{%
\section{Appendix D: Habituals Items}\label{appendix-d-habituals-items}}

\begingroup\fontsize{9pt}{10pt}\selectfont
\begin{longtable}{ |p{1in}|p{1.2in}| p{1.75in}|p{1.75in} |}
  \hline
{\bfseries habitual} & {\bfseries referent frequency} & {\bfseries enabling} & {\bfseries preventative} \\ 
  \hline
climbs mountains & 5 years*, 2 years, year & Yesterday, William remembered how much fun that was bought a lot of new mountain climbing gear. & Yesterday, William turned 80 and gave up all strenuous physical exercise because his doctor said it would be deadly. \\ 
   \hline
does cocaine & 5 years, year*, month, week & Yesterday, Tim wanted to get high and bought some cocaine. & Yesterday, William realized he is severely allergic to cocaine and will no longer do it. \\ 
   \hline
drinks beer & year, month*, 2 weeks, week & Yesterday, Tim wanted to get tipsy and bought a six-pack of beer. & Yesterday, William gave up alcohol and entered into Alcoholics Anonymous. \\ 
   \hline
drinks coffee & year, month*, 2 weeks, week & Yesterday, William wanted a morning jolt and bought a pound of fresh roasted coffee. & Yesterday, Veronica developed a caffeine allergy and decided to give up all caffeine. \\ 
   \hline
eats caviar & 5 years*, year, month & Yesterday, William learned about the dietary benefits of eating caviar and bought a jar at the supermaket. & Yesterday, Tina developed a seafood allergy. \\ 
   \hline
eats peanut butter & 5 years, year*, month & Yesterday, Veronica learned about the dietary benefits of eating peanut butter and bought a jar at the supermaket. & Yesterday, Ted developed a peanut allergy. \\ 
   \hline
goes to the ballet & 2 years, year, month &  &  \\ 
   \hline
goes to the movies & 2 years, year, month &  &  \\ 
   \hline
hikes & 2 years, year*, 2 months, week & Yesterday, William remembered how much fun those times were and bought a lot of new hiking gear. & Yesterday, Vince was in a motorcycle accident and will never walk again. \\ 
   \hline
listens to live music & year, month, week &  &  \\ 
   \hline
listens to Pandora & year, month, week &  &  \\ 
   \hline
plays tennis & 5 years, 2 years, year* & Yesterday, Tim remembered how much fun that was and bought a new tennis racket. & Yesterday, William developed crippling arthritis in both elbows and can only move his arms extremely slowly. \\ 
   \hline
plays the banjo & 5 years, 2 years, year* & Yesterday, William remembered how much fun that was and joined his friend\&quotechars band as the banjoist. & Yesterday, William developed crippling arthritis in his hands and no longer can play musical instruments. \\ 
   \hline
runs & 2 years, year, 2 months*, week & Yesterday, Vince remembered how much fun those times were and bought a new pair of running shoes. & Yesterday, Veronica was in a car accident and became permanently paralyzed from the waist down. \\ 
   \hline
sells companies & 5 years, year &  &  \\ 
   \hline
sells things on eBay & 5 years, year &  &  \\ 
   \hline
smokes cigarettes & year, month*, week & Yesterday, Tina wanted a smoke and bought a pack of cigarettes. & Yesterday, Vince quit smoking cigarettes. \\ 
   \hline
smokes marijuana & 5 years, year*, month, week & Yesterday, William wanted to get high and bought some marijuana. & Yesterday, Vince realized he is severely allergic to marijuana and will no longer smoke it. \\ 
   \hline
steals cars & 5 years*, year, month & Yesterday, William learned a new technique for breaking into cars. & Yesterday, Tina got caught and went through a radical transformation, vowing to never break the law again. \\ 
   \hline
steals chewing gum & 5 years*, year, month & Yesterday, Vince learned a new trick to distract shopkeepers. & Yesterday, William got caught and vowed to never break the law again. \\ 
   \hline
volunteers for political campaigns & 5 years*, year & Yesterday, Vince researched a new political candidate in the area and is going to volunteer with them. & Yesterday, Vince grew disillusioned with the political system and wants nothing to do with it anymore. \\ 
   \hline
volunteers for soup kitchens & 5 years*, year & Yesterday, Vince researched a new soup kitchen in the area and is going to volunteer with them. & Yesterday, Tom grew disillusioned with the soup kitchen system and wants nothing to do with it anymore. \\ 
   \hline
watches professional football & 2 years, year*, month & Yesterday, William remembered how much enjoyable that was and upgraded his cable to have access to all professional football games. & Yesterday, Veronica learned about all the corruption in professional sports and no longer can watch it. \\ 
   \hline
watches space launches & 2 years*, year, month & Yesterday, William remembered how much enjoyable that was and researched all of the space launches in the next year within driving distance. & Yesterday, William went through a radical transformation and now it is against his belief to witness anything relating to space travel. \\ 
   \hline
wears a bra & 6 months, month, week &  &  \\ 
   \hline
wears a suit & 6 months, month*, week & Yesterday, Vince got a high paying job on Wall Street. & Yesterday, Veronica got fired from her job on Wall Street and now works in a pizza parlor. \\ 
   \hline
wears a watch & 6 months, month, week &  &  \\ 
   \hline
wears slacks & 6 months, month, week &  &  \\ 
   \hline
wears socks & 6 months, month, week &  &  \\ 
   \hline
writes novels & 5 years*, year & Yesterday, William finished an MFA program and quit his other job to focus on writing novels. & Yesterday, William became fed up with the literary world and decided to never write anything again. \\ 
   \hline
writes poems & 5 years, year* & Yesterday, William finished an MFA program and quit his other job to focus on writing poems. & Yesterday, Veronica became fed up with the poetry world and decided to never write poems again. \\ 
   \hline
\hline
\end{longtable}
\endgroup

Items used in Case Study 2 (Habitual language). Referent frequency
denotes the experimentally supplied time periods during which a person
did an action 3 times (e.g., \enquote{In the past 5 years, John climbed
mountains three times.}; Expt. 2b). Enabling and preventative columns
provide the causal manipulation sentences used in Expt. 2c in order to
enable or prevent future instances of the action (blank entries indicate
that the item was not used in Expt. 2c). Referent frequency with an
asterisk denotes the time interval used in Expt. 2c.

\end{document}